\documentclass{siamonline171218}
\usepackage{amsmath}
\usepackage{amsfonts}

\usepackage{chngcntr}

\usepackage{xr}
\usepackage{braket}
\usepackage{caption,subcaption}% http://ctan.org/pkg/{caption,subcaption}

\usepackage{pgfplots}

\newsiamthm{claim}{Claim}
\newsiamremark{remark}{Remark}
\newsiamremark{hypothesis}{Hypothesis}
\crefname{hypothesis}{Hypothesis}{Hypotheses}

\usepackage{graphicx,epstopdf}

\Crefname{ALC@unique}{Line}{Lines}

\usepackage{graphicx}
\usepackage{color}
\usepackage{algorithm}
\usepackage[nocompress]{cite}
\usepackage{siunitx}
\usepackage{epstopdf}
\usepackage{algpseudocode}

\usepackage[toc, page]{appendix}
\usepackage[normalem]{ulem}
\usepackage{cancel}
\usepackage{multirow}
\usepackage{xspace}
\usepackage{bold-extra}
\usepackage[most]{tcolorbox}

\colorlet{texcscolor}{blue!50!black}
\colorlet{texemcolor}{red!70!black}
\colorlet{texpreamble}{red!70!black}
\colorlet{codebackground}{black!25!white!25}

\lstdefinestyle{siamlatex}{%
  style=tcblatex,
  texcsstyle=*\color{texcscolor},
  texcsstyle=[2]\color{texemcolor},
  keywordstyle=[2]\color{texemcolor},
  moretexcs={cref,Cref,maketitle,mathcal,text,headers,email,url},
}

\tcbset{%
  colframe=black!75!white!75,
  coltitle=white,
  colback=codebackground, % bottom/left side
  colbacklower=white, % top/right side
  fonttitle=\bfseries,
  arc=0pt,outer arc=0pt,
  top=1pt,bottom=1pt,left=1mm,right=1mm,middle=1mm,boxsep=1mm,
  leftrule=0.3mm,rightrule=0.3mm,toprule=0.3mm,bottomrule=0.3mm,
  listing options={style=siamlatex}
}

\DeclareTotalTCBox{\code}{ v O{} }
{ %fontupper=\ttfamily\color{texemcolor},
  fontupper=\ttfamily\color{black},
  nobeforeafter,
  tcbox raise base,
  colback=codebackground,colframe=white,
  top=0pt,bottom=0pt,left=0mm,right=0mm,
  leftrule=0pt,rightrule=0pt,toprule=0mm,bottomrule=0mm,
  boxsep=0.5mm,
  #2}{#1}

\patchcmd\newpage{\vfil}{}{}{}
\newcommand{\vl}{\boldsymbol{l}}

\newcommand{\vu}{\boldsymbol{u}}
\newcommand{\vv}{\boldsymbol{v}}

\newcommand{\vx}{\boldsymbol{x}}

\newcommand{\vzero}{\boldsymbol{0}}
\newcommand{\vone}{\boldsymbol{1}}

\newcommand{\mC}{\boldsymbol{C}}
\newcommand{\mD}{\boldsymbol{D}}
\newcommand{\mG}{\boldsymbol{G}}
\newcommand{\mH}{\boldsymbol{H}}
\newcommand{\mI}{\boldsymbol{I}}

\newcommand{\mP}{\boldsymbol{P}}
\newcommand{\mR}{\boldsymbol{R}}
\newcommand{\mO}{\boldsymbol{O}}
\newcommand{\mS}{\boldsymbol{S}}
\newcommand{\mU}{\boldsymbol{U}}
\newcommand{\mV}{\boldsymbol{V}}

\newcommand{\mW}{\boldsymbol{W}}

\newcommand{\cQ}{\mathcal{Q}}
\newcommand{\cP}{\mathcal{P}}

\newcommand{\mSigma}{\boldsymbol{\Sigma}}

\newcommand{\vmu}{\boldsymbol{\mu}}

\newcommand{\GG}{\mathbb{G}}

\newcommand{\RR}{\mathbb{R}}

\newcommand{\ZZ}{\mathbb{Z}}

\newcommand{\tr}{\mbox{tr}}

\newcommand{\argmax}{\operatornamewithlimits{argmax}}
\newcommand{\argmin}{\operatornamewithlimits{argmin}}

\newcommand{\Perm}{\text{Perm}}
\newcommand{\MGC}{\text{MGC}}
\newcommand{\NAM}{\text{NAM}}
\newcommand{\SO}{\text{SO}}
\newcommand{\med}{\text{med}}

\newcommand{\Jaccard}{\text{Jaccard}}
\newcommand{\diag}{\text{diag}}
\newcommand{\true}{\mathrm{true}}
\newcommand{\orig}{\mathrm{orig}}
\newcommand{\init}{\mathrm{init}}

\newcommand{\est}{\mathrm{est}}

\newcommand{\nb}{\text{nb}}
\newcommand{\all}{\text{all}}
\newcommand{\lleft}{\text{l}}
\newcommand{\ttop}{\text{t}}
\newcommand{\rright}{\text{r}}
\newcommand{\bbottom}{\text{b}}
\newcommand{\Err}{\text{Err}}

\newcommand{\lr}{\text{lr}}
\newcommand{\rl}{\text{rl}}
\newcommand{\tb}{\text{tb}}
\newcommand{\bt}{\text{bt}}

\newcommand{\ltr}{\text{ltr}}
\newcommand{\trb}{\text{trb}}
\newcommand{\blt}{\text{blt}}
\newcommand{\lbr}{\text{lbr}}

\newcommand{\LR}{\text{LR}}
\newcommand{\RL}{\text{RL}}

\begin{tcbverbatimwrite}{tmp_\jobname_header.tex}
\title{Solving Jigsaw Puzzles By The Graph Connection Laplacian
\thanks{%Submitted to the editors \today.
\funding{This research has been supported by NSF awards DMS-14-18386, DMS-18-30418 and CCF–1740858.}}}

\author{Vahan Huroyan\thanks{Department of Mathematics, The University of Arizona, Tucson, AZ 85721 (\email{vahanhuroyan@math.arizona.edu}).}
\and Gilad Lerman\thanks{School of Mathematics, University of Minnesota, Twin Cities, Minneapolis, MN 55455 (\email{lerman@umn.edu}).}
\and Hau-Tieng Wu\thanks{Department of Mathematics and Department of Statistical Science, Duke University, Durham, NC 27708, United States; Mathematics Division, National Center for Theoretical Sciences, Taipei, Taiwan (\email{hauwu@math.duke.edu}).}}

\headers{Solving Jigsaw Puzzles By The Graph Connection Laplacian}{V. Huroyan, G. Lerman, and H-T. Wu}
\end{tcbverbatimwrite}
\title{Solving Jigsaw Puzzles By The Graph Connection Laplacian
\thanks{Submitted to the editors \today.
\funding{This research has been supported by NSF awards DMS-14-18386, DMS-18-30418 and CCF–1740858.}}}

\author{Vahan Huroyan\thanks{Department of Mathematics, The University of Arizona, Tucson, AZ 85721 (\email{vahanhuroyan@math.arizona.edu}).}
\and Gilad Lerman\thanks{School of Mathematics, University of Minnesota, Twin Cities, Minneapolis, MN 55455 (\email{lerman@umn.edu}).}
\and Hau-Tieng Wu\thanks{Department of Mathematics and Department of Statistical Science, Duke University, Durham, NC 27708, United States; Mathematics Division, National Center for Theoretical Sciences, Taipei, Taiwan (\email{hauwu@math.duke.edu}).}}

\headers{Solving Jigsaw Puzzles By The Graph Connection Laplacian}{V. Huroyan, G. Lerman, and H-T. Wu}

\ifpdf
\hypersetup{pdftitle={Solving Jigsaw Puzzles By The Graph Connection Laplacian} }
\fi

\begin{document}

\counterwithout{figure}{section}
\counterwithout{equation}{section}
\counterwithout{table}{section}
\counterwithout{algorithm}{section}

\maketitle

\begin{abstract}
We propose a novel mathematical framework to address the problem of automatically solving large jigsaw puzzles. This problem assumes a large image,
which is cut into equal square pieces that are arbitrarily rotated and shuffled, and asks to recover the original image given the transformed pieces.
The main contribution of this work is a method for recovering the rotations of the pieces when both shuffles and rotations are unknown.
A major challenge of this procedure is estimating the graph connection Laplacian without the knowledge of shuffles.
A careful combination of our proposed method for estimating rotations with any existing method for estimating shuffles results in a practical solution for the jigsaw puzzle problem.
Our theory guarantees, in a clean setting, that our basic idea of recovering rotations is robust to some corruption of the connection graph.
Numerical experiments demonstrate the competitive accuracy of this solution, its robustness to corruption and its computational advantage for large puzzles.
\end{abstract}

\begin{AMS}
{90C20, 90C27, 90C35, 90C90}%, 05C50, 49M25}
\end{AMS}

\section{Introduction}
\label{sec:intro}

Solving jigsaw puzzles is an entertaining task, which is commonly explored by children and adults. It is also a challenging mathematical and engineering problem that occupies researchers in computer science, mathematics and engineering. The solution of this problem is useful for several industrial applications. One example is reassembling archaeological artifacts \cite{brown2008system,koller2006computer,toler2010multi,pintus2016survey,sizikova2017wall}, where one tries to recover the shape of an archaeological object from damaged pieces. Another example is recovering shredded documents or photographs \cite{justino2006reconstructing,marques2009reconstructing,liu2011automated, deever2012semi}, where one tries to recover a document or a picture from small pieces of it. Additional applications appear in biology \cite{marande2007mitochondrial} and speech descrambling \cite{zhao2007puzzle}.

The automatic solution of puzzles, without having any information on the underlying image, is known to be NP hard \cite{altman1989solving, demaine07jigsaw}. The first algorithm that attempted to automatically solve general puzzles was introduced by Freeman and Garder \cite{freeman1964apictorial} in 1964. It was designed to solve puzzles with $9$ pieces by only considering the geometric shapes of the pieces.

In this paper we consider a setting of ``jigsaw'' puzzles, which is common in the imaging sciences \cite{pomeranz2011fully, gallagher2012jigsaw, andalo2012solving, mondal2013robust, son2014solving, paikin2015solving, son2016solving, chen2018new}. In this setting, an image is cut into equal square pieces, and the problem is to recover this image from the given pieces, which are possibly rotated and shifted along the puzzle grid. We refer to these puzzles as square jigsaw puzzles. Some examples are demonstrated in \Cref{fig:1}.
Gallagher \cite{gallagher2012jigsaw} categorized these puzzles into three types. In type 1 puzzles, the pieces are not rotated, but shifted. In type 3 puzzles, the pieces are not shifted, but rotated. In type 2 puzzles, the pieces are both shifted and rotated.
This work aims to solve type 2 and type 3 puzzles.

Many proposals for solving the square jigsaw puzzles are based on greedy methods \cite{pomeranz2011fully, gallagher2012jigsaw, andalo2012solving, mondal2013robust, son2014solving, paikin2015solving, son2016solving, chen2018new}. However, greedy algorithms can easily get trapped in locally optimal solutions, which are not global. Some proposals also involve non-greedy constructive methods \cite{cho2010probabilistic, sholomon2014generalized, sholomon2016automatic}, which are often combined with greedy procedures. This work proposes a constructive framework for recovering orientations of puzzle pieces. The overall procedure for recovering both orientations and locations requires various heuristics. However, it avoids common greedy procedures in solving this problem. Our main purpose is to convey the effective use of the recent mathematical idea of the graph connection Laplacian \cite{singer2012vector} for recovering orientations of type 2 puzzles. Unlike previous methods, it is easy to understand its constructive mechanism and even guarantee some robustness to measurement errors in a clean setting. In practice, we demonstrate robustness to corruption and computational efficiency for large puzzles.

\subsection{Previous Work}

Several algorithms have been recently proposed for the automatic solution of square jigsaw puzzles \cite{cho2010probabilistic, yang2011particle, pomeranz2011fully,  sholomon2016automatic, sholomon2014generalized, sholomon2016dnn, sholomon2014genetic, gallagher2012jigsaw, mondal2013robust, son2014solving, son2016solving, yu2015solving, paikin2015solving, andalo2012solving, chen2018new}. The problem becomes more challenging when the number of puzzle pieces increases and the sizes of puzzle pieces decrease. Some of these algorithms only consider type 1 puzzles (see e.g., \cite{cho2010probabilistic, pomeranz2011fully, sholomon2014generalized, zhao2007puzzle, andalo2012solving}), since recovering orientations increases the possible comparisons between two pieces by four and may also decrease the accuracy of solving the puzzle. The rest of these algorithms focus on type 2 puzzles, where \cite{gallagher2012jigsaw} also separately discusses type 3 puzzles. Other models of jigsaw puzzles and probabilistic results for their solutions are discussed in~\cite{mossel2018shotgun, bordenave2016shotgun, martinsson2016shotgun}.

Cho et al.~\cite{cho2010probabilistic} proposed a probabilistic, graphical model approach to the square jigsaw puzzle problem and discussed different compatibility metrics between puzzle pieces.
Yang et al.~\cite{yang2011particle} proposed another probabilistic solution by using a particle filter and a state permutations framework.
Pomeranz et al.~\cite{pomeranz2011fully} proposed a greedy method, discussed a few compatibility metrics and included some analysis on how to pick the correct compatibility metric for their method.
Gallagher \cite{gallagher2012jigsaw} proposed a tree-based reassembly algorithm, which greedily merges components while respecting the geometric consistence constraints. It runs in three steps: building a constrained tree, trimming and filling. Mondal et al.~\cite{mondal2013robust} used the algorithm of Gallagher~\cite{gallagher2012jigsaw}, but they replaced its proposed metric with a combination of two existing metrics. They claimed to achieve a more robust metric using this technique. Andalo et al. \cite{andalo2012solving} proposed a quadratic assignment approach, which maximizes a constrained quadratic function via constrained gradient ascent.
Jin et al.~\cite{jin2014jigsaw} proposed a scoring approach that, in addition to considering edge similarity, also takes into account content similarity between puzzle pieces.
Paikin and Tal~\cite{paikin2015solving} proposed a greedy algorithm for handling puzzles of unknown size and with missing entries.
Sholomon et al.~\cite{sholomon2014generalized, sholomon2014genetic, sholomon2016automatic} proposed a genetic algorithm.
Sholomon et al.~\cite{sholomon2016dnn} proposed a new Deep Neural Network-Based approach for the prediction of the likelihood of correct matches.

Son et al.~\cite{son2014solving} incorporated the ``geometric structure'' of the square jigsaw puzzle by searching for small loops (4-cycles) of puzzle pieces, which form consistent cycles, and then hierarchically combining these small loops with higher order loops in a bottom-up fashion.
They argued that loop constraints could effectively eliminate pairwise matching outliers. Son et al.~\cite{son2016solving} proposed a growing consensus approach that assembles pieces by multiple modest bonds and uses a new objective function that maximizes consensus configurations.
Yu et al.~\cite{yu2015solving} proposed a linear programming based formulation, which combines global and greedy approaches. Their proposed solver simultaneously exploits all the pairwise matches and globally computes the location of each piece/component at each step of the algorithm.
Chen et al.~\cite{chen2018new} proposed a greedy algorithm and combined several metrics to improve the performance of this algorithm.

The only previous procedure for solving type 3 puzzles is by Gallagher \cite{gallagher2012jigsaw}. It uses a greedy and non-constructive method.
We are unaware of any previous constructive method for finding the orientations of type 2 puzzle pieces.

%Makridis and Papamarkos~\cite{makridis2006new} proposed an algorithm which takes advantage of both geometrical and color features of puzzle pieces.

%Nielsen et al.~\cite{nielsen2008solving} proposed a solution of jigsaw puzzles with considering only image features instead of the shape of the pieces.

%An example of 3D puzzle reconstruction is proposed by Oxholm and NIshino~\cite{oxholm2013flexible}.

%Yao and Shao~\cite{yao2003shape} proposed a shape and image matching method with cyclick ``grouth'' process to solve jigsaw puzzles with various piece shapes.

\subsection{Our Contribution}
\label{sec:our_contribution}

In this paper we propose a novel approach to address type $2$ and type $3$ jigsaw puzzles.
For type 3 puzzles, we suggest a fast, robust, constructive and straightforward solution that uses the graph connection Laplacian (GCL) \cite{singer2012vector}  (discussed in \S\ref{sec:GCL}).
For type $2$ puzzles we propose a novel iterative algorithm, which solves the following two subproblems:  The Rotation Problem (RotP) and the Location Problem (LocP):
\begin{itemize}
\item[\textbf{RotP:}] Finding the orientations of all puzzle pieces.
\item[\textbf{LocP:}] Finding the locations of all puzzle pieces.
\end{itemize}
These two steps are iteratively repeated until the desired result is achieved.
We solve RotP by using the GCL, where the main challenge is to construct the GCL despite the unknown locations.
We solve LocP by applying any state-of-the-art solution of type 1 puzzles to the puzzle obtained from the solution of RotP.
Some information inferred from the solution of LocP is further used to improve the solution of RotP.

All previous algorithms for solving type $2$ puzzles simultaneously address RotP and LocP. On the other hand, this work separately solves the two subproblems, with a constructive and better understood solution of RotP.
Moreover, we aim to present a principled approach and thus avoid greedy steps that are common in previous algorithms and help improve the accuracy.
Empirically, the proposed method is faster than other methods for large puzzles, e.g., with thousands of patches. We can also specify more easily the overall computational complexity of our proposed components (excluding the borrowed type 1 puzzle solver). This complexity is comparable to that of the most common component of any puzzle solver. Our numerical results also demonstrate that our algorithm is more robust to corruption of the sides of puzzle patches than other algorithms.

In theory, we verify robustness under a certain mathematical setting and contribute with perturbation-type result to the general mathematical area of group synchronization. However, we would like to emphasize that the paper addresses a real applied problem, where it is unclear how to make sufficiently good measurements, which are assumed by the theory. We thus have two different components that may seem in tension. One is the clean theory that justifies recovery of orientations given special type of measurements. The other one, is a set of methods, supported by numerical experimentation, with specific choices of measurements of initial relative orientations and graph affinities between puzzle patches. The methods include various heuristics, for example, for improving the latter measurements given better information of patch locations. This is a first attempt to incorporate a successful puzzle solver within a rigorous mathematical setting. We hope that with time the proposed mathematical foundations and algorithms can be further improved and simplified.

\subsection{Structure of This Paper}

This paper is organized as follows:
\S\ref{sec:math_model_sec} mathematically formulates the square jigsaw puzzle problem;
\S\ref{sec:rot_rec} presents a solution for RotP , given some initial measurements of relative orientations and graph affinities between puzzle patches.
It also theoretically guarantees the robustness of the solution to errors in the initial measurements;
\S\ref{sec:conn_graph_const} explains how to measure in practice the initial relative orientations and graph affinities for type 2 and type 3 puzzles. Note that the combination of this construction with the method of \S\ref{sec:rot_rec} provides the desired solution to RotP. While \S\ref{sec:rot_rec} has a clean formulation with a theoretical guarantee, \S\ref{sec:conn_graph_const} relies on various heuristics, which we try to motivate;
\S\ref{sec:loc_est} describes additional heuristics for improving the initial measurements and consequently the solution of RotP  given the solution of LocP. It also summarizes our full algorithm for solving square jigsaw puzzles;
\S\ref{sec:num_exp} presents  numerical experiments that test the accuracy, efficiency and robustness to corruption of the proposed algorithm using digital images; Finally, \S\ref{sec:discussion} concludes with a short discussion that includes possible extensions of this work.

\section{Mathematical Formulation and Notation}
\label{sec:math_model_sec}
We mathematically formulate the square jigsaw puzzle problem and introduce relevant notation in \S\ref{subsec:math_special}. In \S\ref{subsec:math_problem} we emphasize the main challenge in solving this problem. We remark that %\S\ref{subsec:math_general} describes
a more general formulation appears in the supplemental material.

\subsection{Setting and Notation}
\label{subsec:math_special}
%\end{document}
The setting of the square jigsaw puzzle problem assumes a rectangle $M = [a_1,b_1]\times[a_2,b_2]$ in $\RR^2$ and open squares $\{P_i\}_{i=1}^n$ that tile $M$. We refer to $\{P_i\}_{i=1}^n$ as patches
and to the four nearest neighbors of a given patch (left, right, top or bottom) as neighboring patches.
The setting further assumes a function $f\in L^2(M,\RR^k)$ and $k \geq 1$, so that the image to be recovered is the graph $\{(\vx,f(\vx)): \ \vx \in M \subset \RR^2\}$. One may use $k=1$ for gray-scale images, $k=3$ for color images and higher $k$ for multispectral and hyperspectral images. In this paper we use $k=3$. Since a main challenge of the practical problem is dealing with discrete images, we further assume that $f$ is piecewise constant with discrete values in the following way. Each patch is divided by a uniformly spaced grid to $s \times s$ subsquares and the vector-valued $f$ is constant on each subsquare, where each coordinate of the constant vector is discrete; for example, it lies in ${0,\ldots,255}$.

The setting also assumes arbitrary orientations and shuffles of puzzle pieces. It is sufficient to represent all possible orientations with rotations by $\ang{0}$, $\ang{90}$, $\ang{180}$ or $\ang{270}$, which are elements of the cyclic group $\ZZ_4$, and  translations of patches $\{P_i\}_{i=1}^n$ by $\{P_{\sigma(i)}\}_{i=1}^n$, where $\sigma$ is a permutation of degree $n$. Therefore, we can write the set of image patches as
$\cQ = \{(R_{\sigma(i)} (P_{\sigma(i)}), R_{\sigma(i)} \circ f|_{P_{\sigma(i)}})\}_{i=1}^n$, where $R_{\sigma(i)}$ is an element of the cyclic group $\ZZ_4$ and the action $\circ$ is defined
by $R_j \circ f|_{P_j}:=f(R_j^{-1})|_{P_j}$. While we use a very formal notation, an image patch is composed of a shifted and rotated patch together with the appropriately aligned value of $f$. With this notation the square jigsaw puzzle problem can be expressed as follows: Given $M$ and the above set of image patches $\cQ$, recover the function $f|_M$ or equivalently the image $(M, f|_M)$.

\begin{figure}[h]
\begin{center}
\includegraphics[width=0.32\columnwidth]{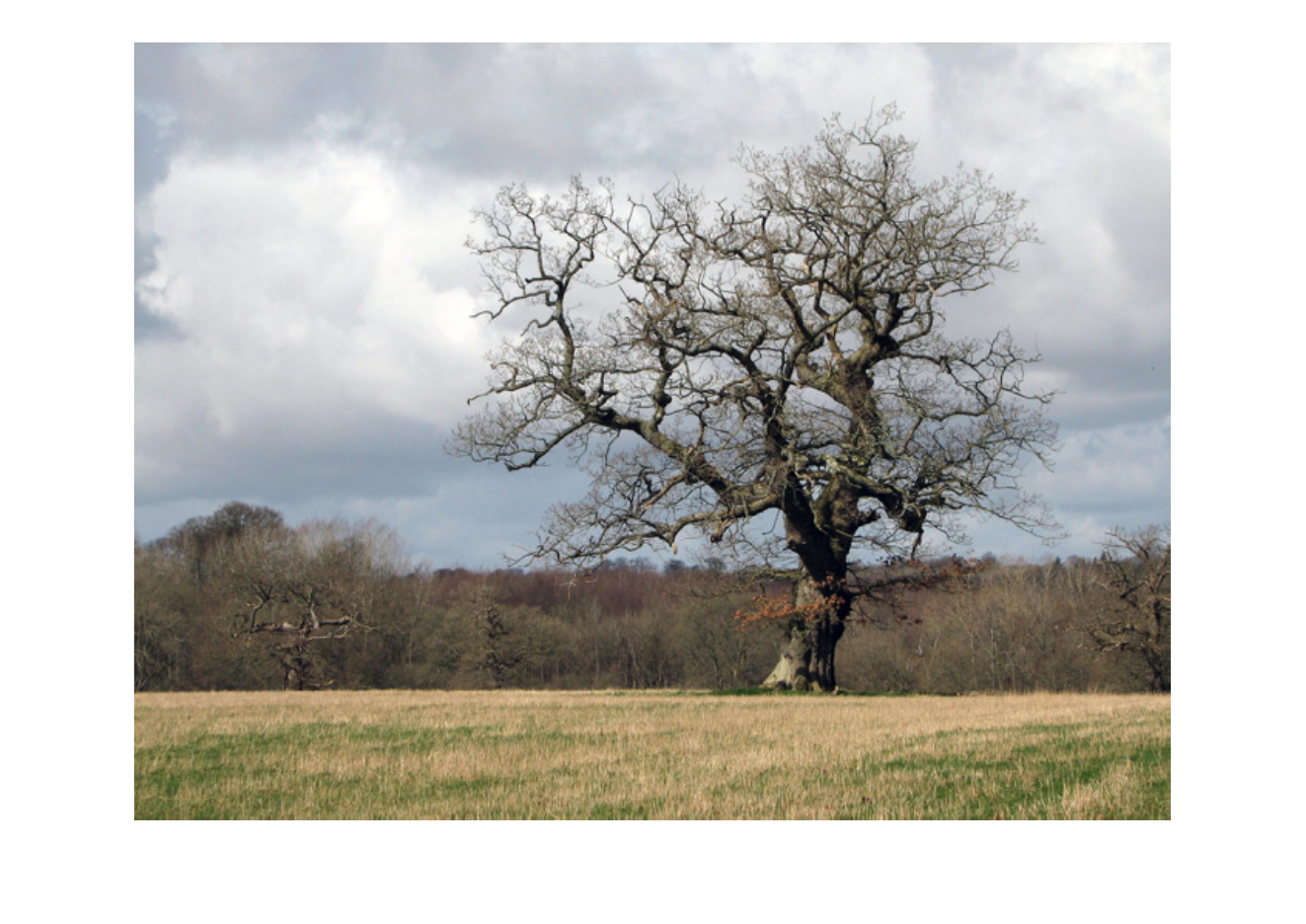}
%\hspace{0.05cm}
\includegraphics[width=0.32\columnwidth]{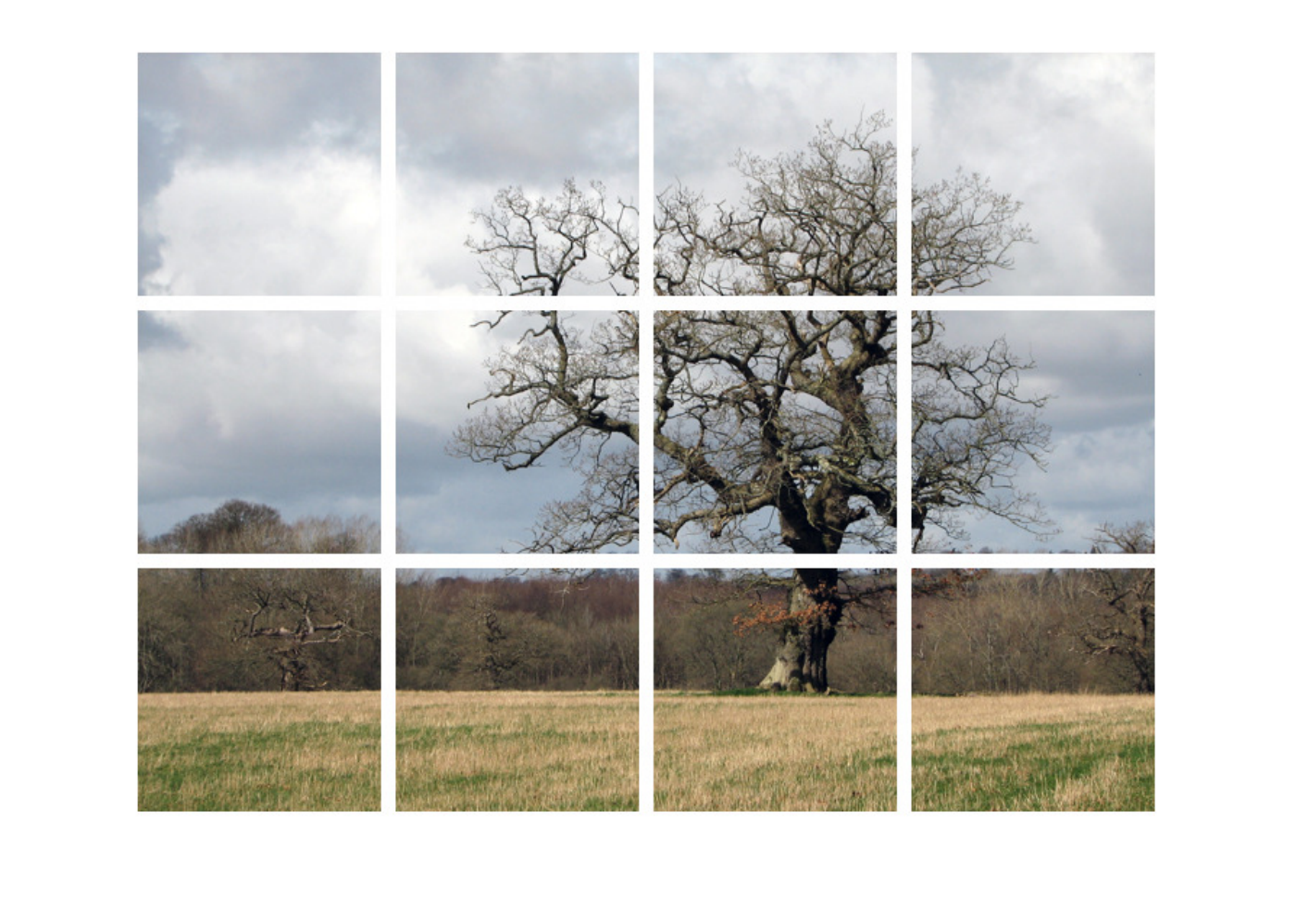}
%\hspace{0.05cm}
\includegraphics[width=0.32\columnwidth]{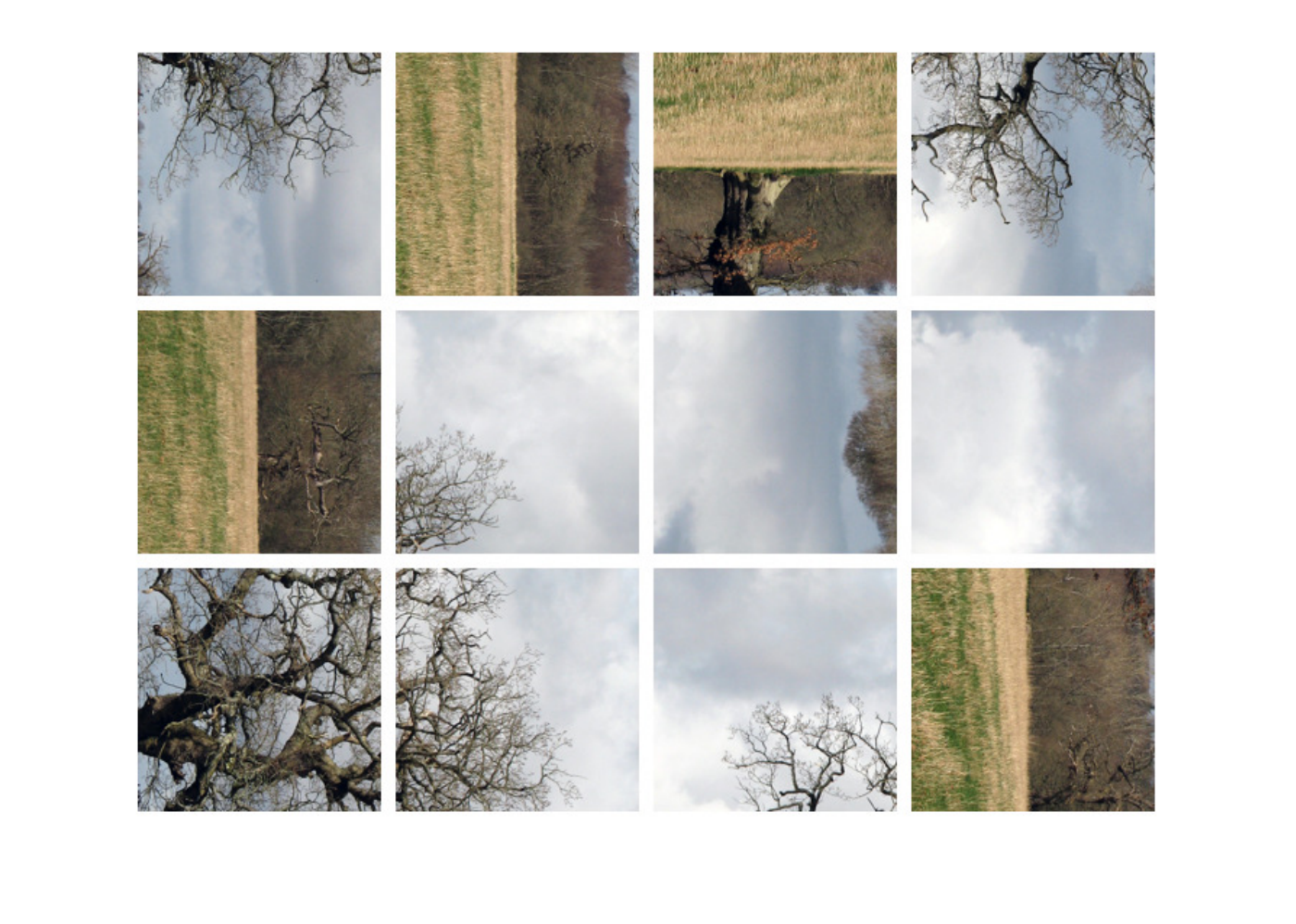}

\includegraphics[width=0.32\columnwidth]{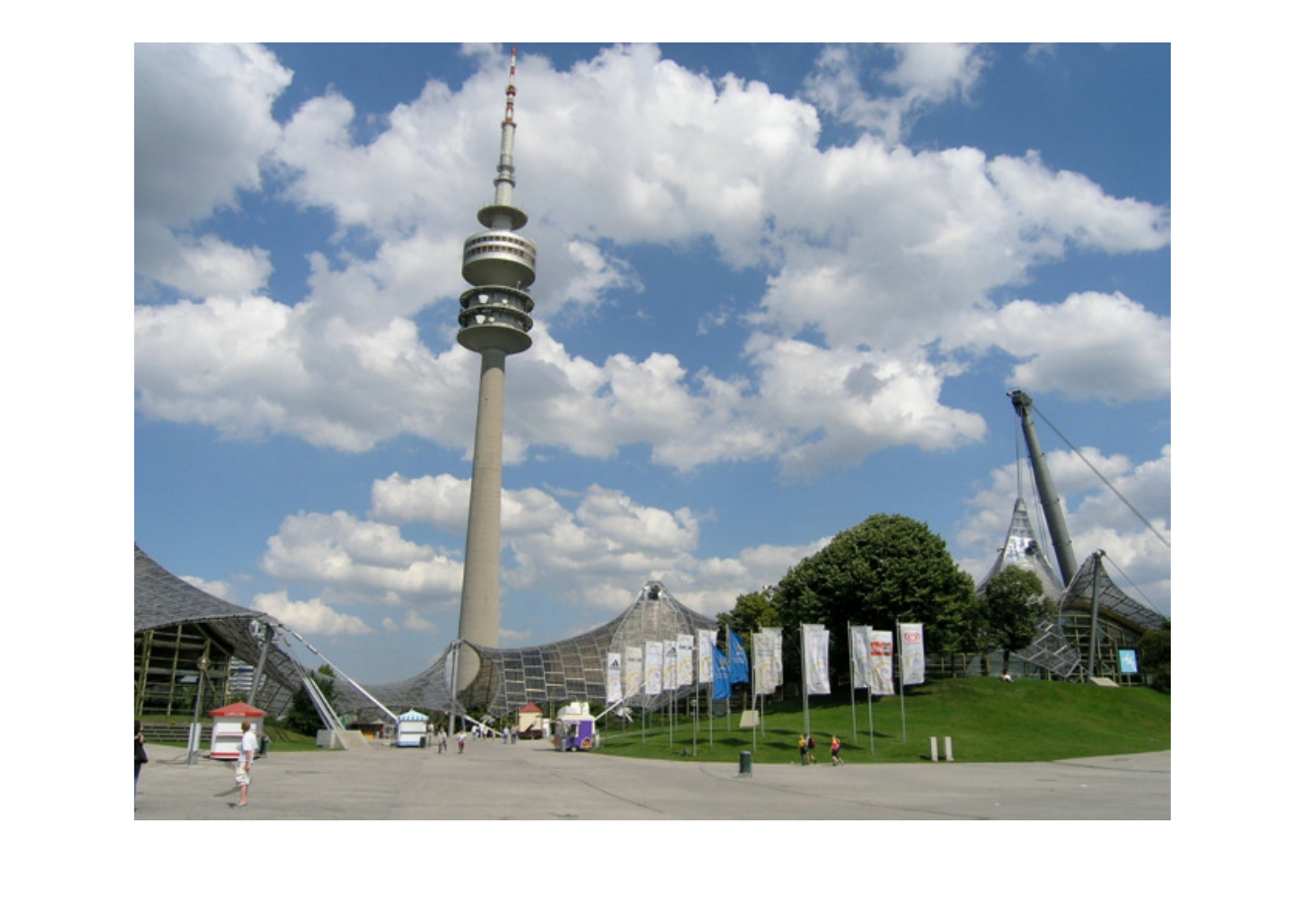}
\includegraphics[width=0.32\columnwidth]{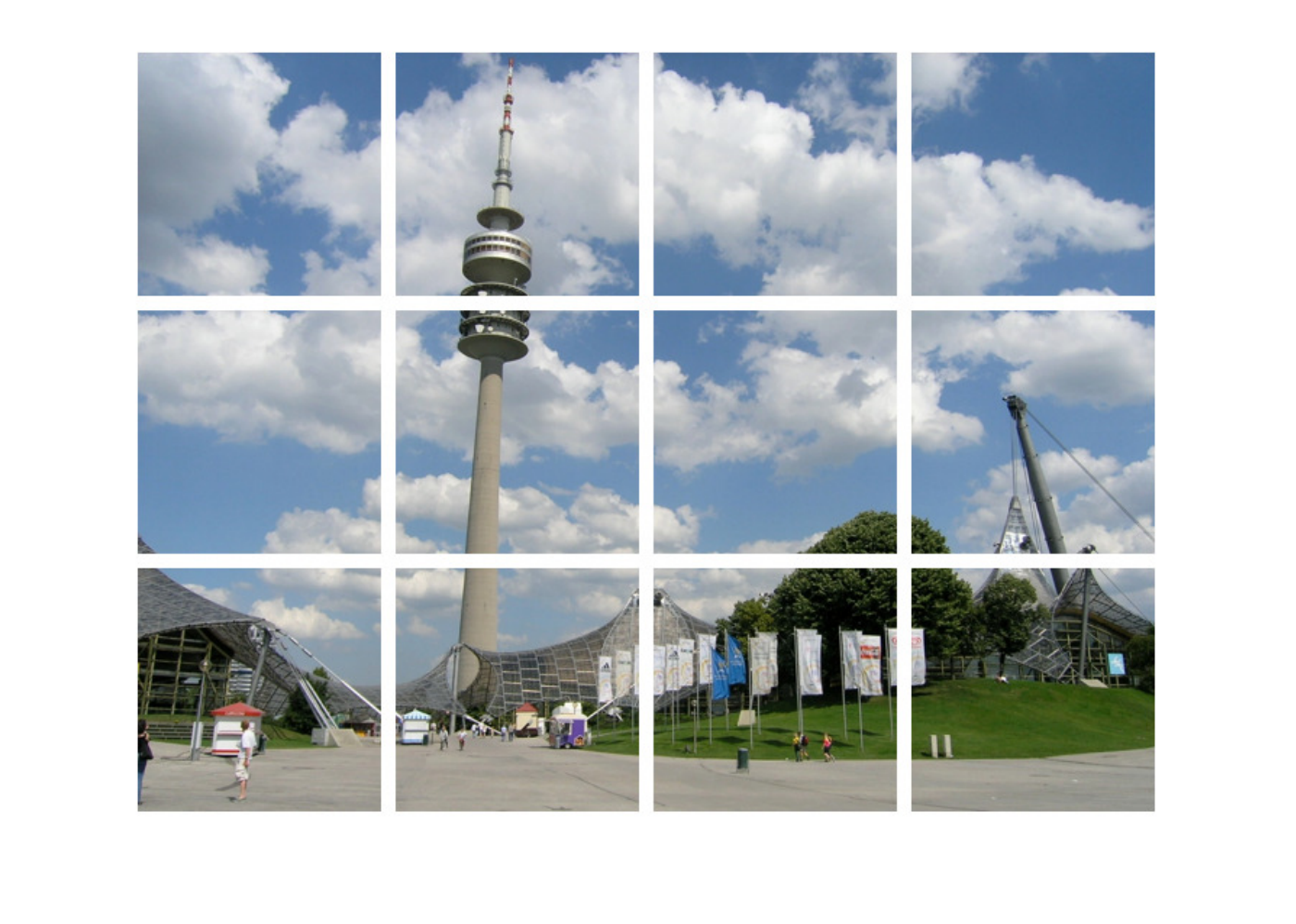}
\includegraphics[width=0.32\columnwidth]{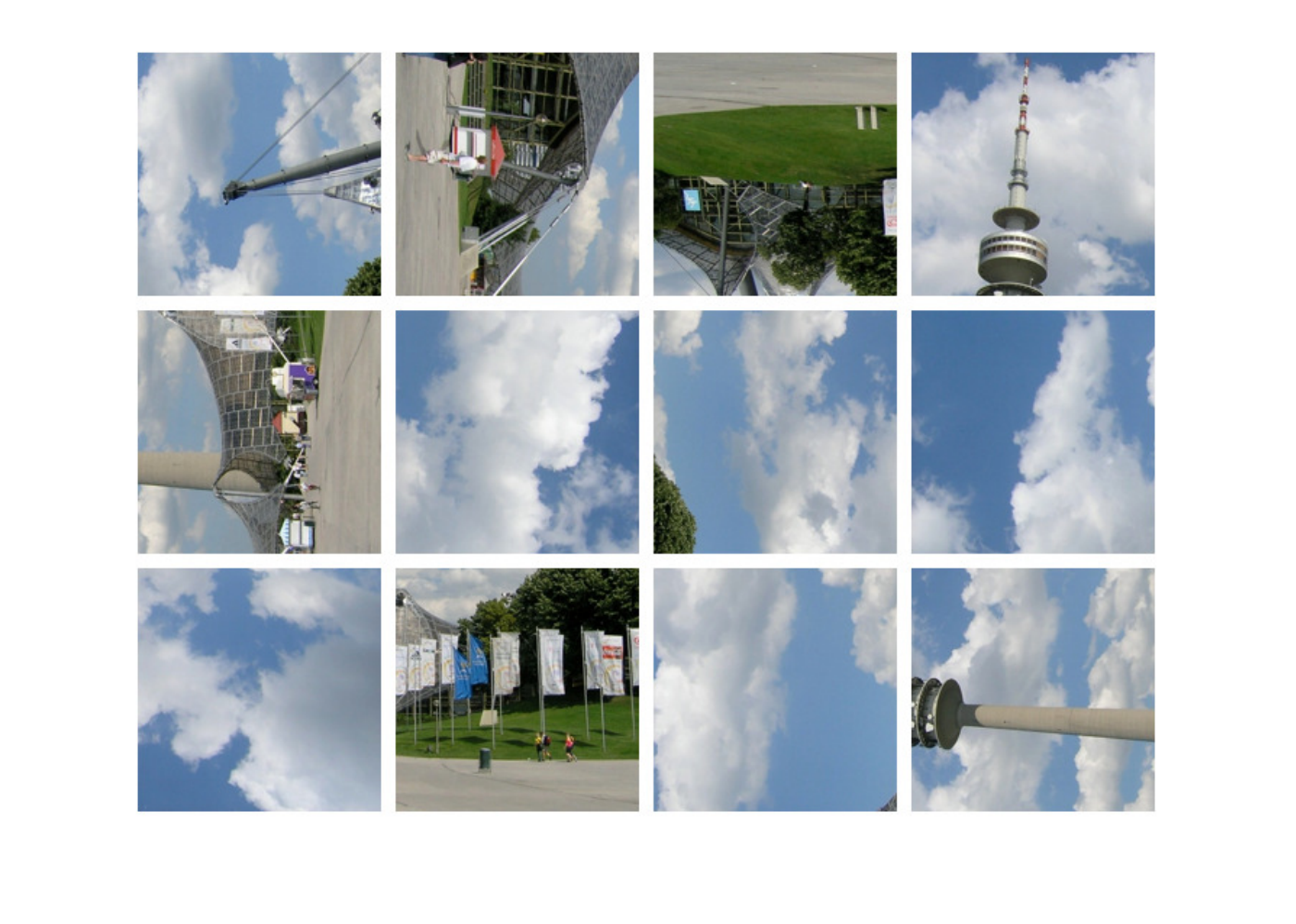}

\end{center}
\caption{Examples of puzzles with $12$ patches. Left column: the original image; Central column: division of the image into $12$ square patches of the same size. Right column: The $12$ patches are randomly reordered and rotated.}
\label{fig:1}
\end{figure}

\Cref{fig:1} demonstrates the particular instance of the square jigsaw puzzle problem we discuss in this paper. We remark that the last column of this figure illustrates the image patches $\cQ = \{(R_{\sigma(i)} (P_{\sigma(i)}), R_{\sigma(i)} \circ f|_{P_{\sigma(i)}})\}_{i=1}^n$ discussed above. We assumed above that $f$ is a piecewise constant function. In this figure, $f$ has constant values on squares corresponding to image pixels. Since the resolution is relatively high, one cannot notice that $f$ is piecewise constant. However, this is noticeable in the low-resolution demonstration of patches of another puzzle at the top right image of \Cref{fig:bad_nb}.

%\begin{figure}[h]
%\begin{center}
%\includegraphics[width=0.4\columnwidth]{puzzleExampleCurvyEdge.jpg}
%\end{center}
%\caption{The patches have curvy edges.}\label{fig:3}
%\end{figure}

%\begin{figure}[h]
%\begin{center}
%\includegraphics[width=0.6\columnwidth]{puzzleExample3D.jpg}
%\end{center}
%\caption{3D puzzle problem.}\label{fig:5}
%\end{figure}

\subsection{A Challenge of Square Jigsaw Puzzles}
\label{subsec:math_problem}

\begin{figure}[h]
\begin{center}
%\subfigure[]{\label{fig:bad_nb_1}
%\includegraphics[width=0.48\columnwidth]{Figures/example_edges.eps}
\includegraphics[width=0.48\columnwidth]{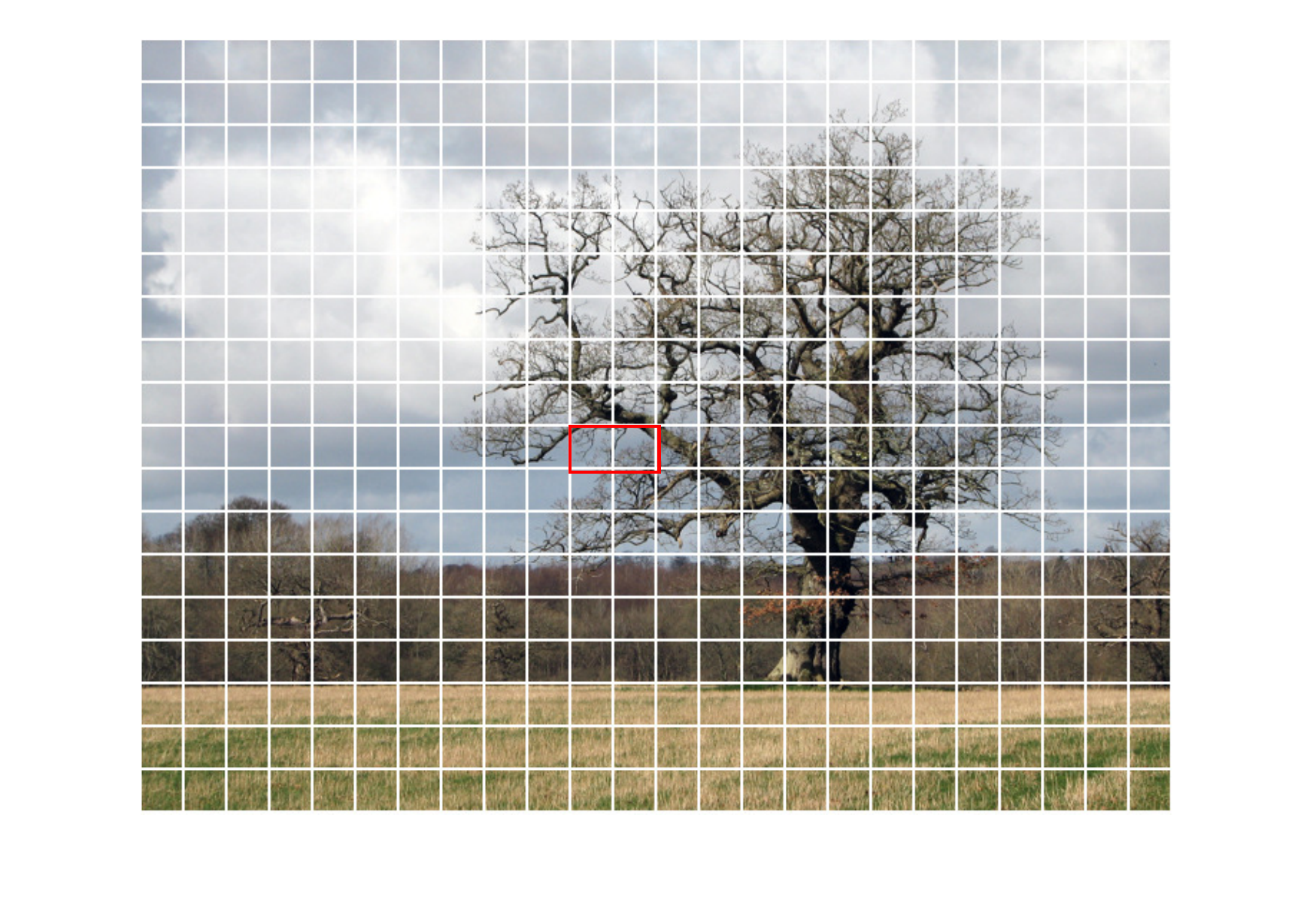}
\includegraphics[width=0.48\columnwidth]{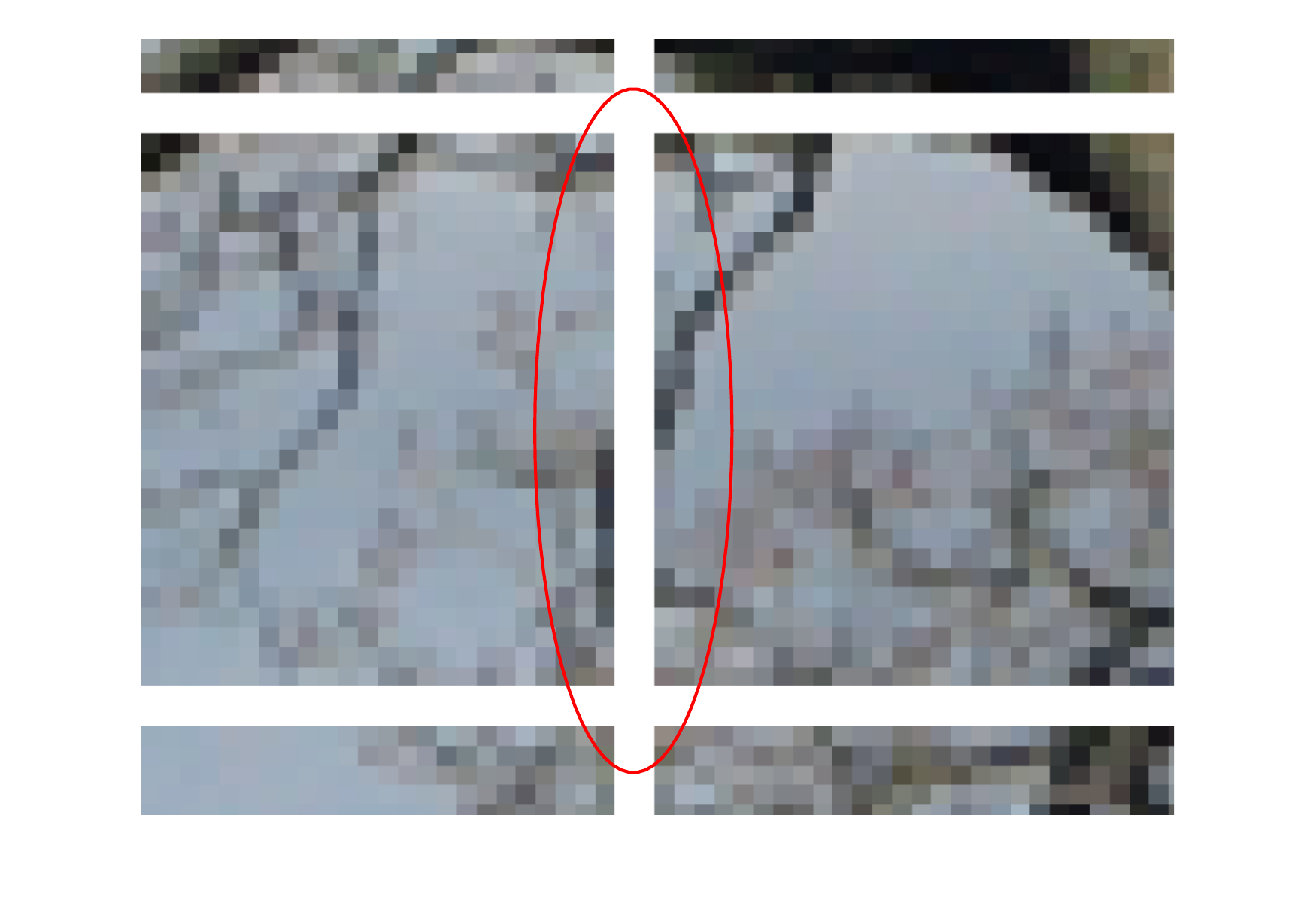}

%\subfigure[]{\label{fig:bad_nb_2}
\includegraphics[width=0.48\columnwidth]{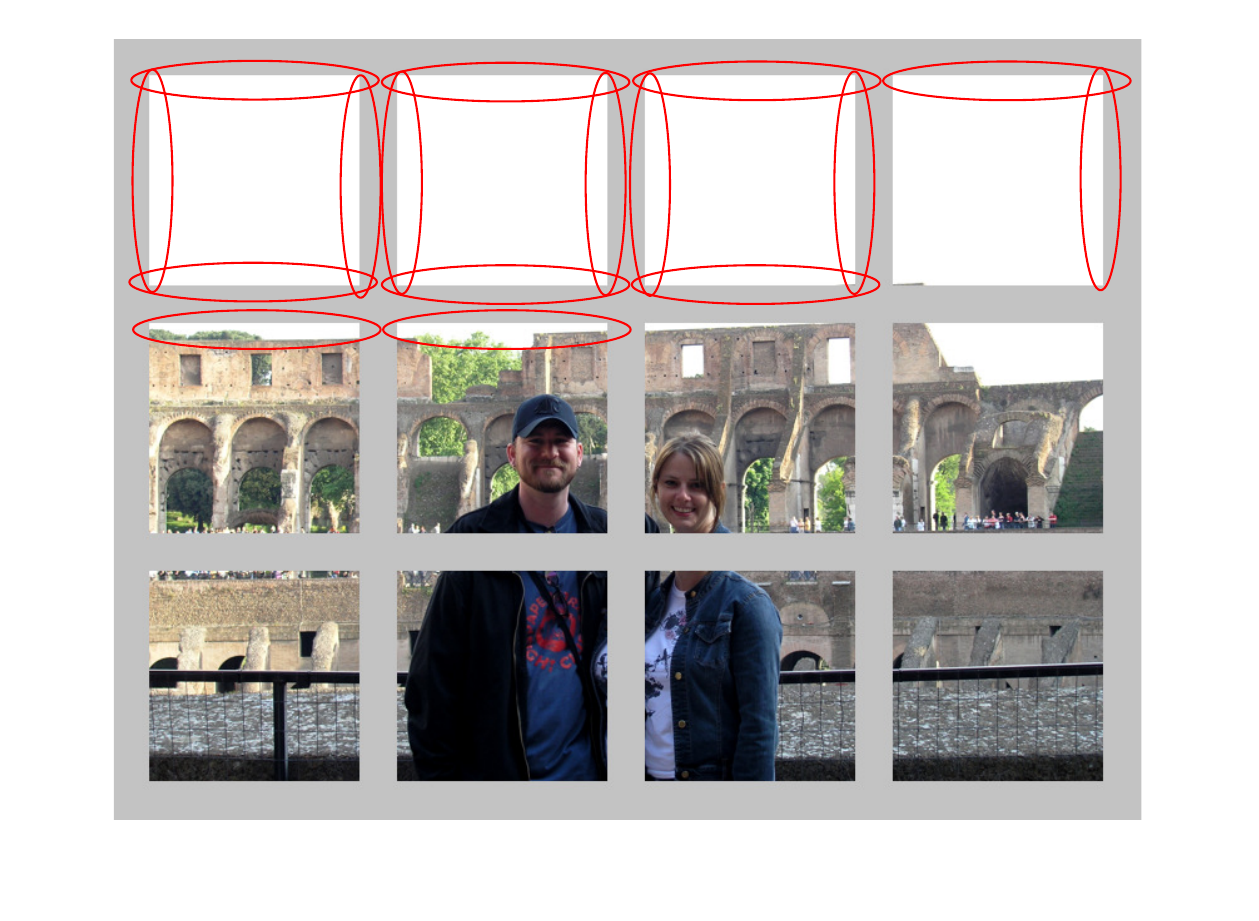}
\includegraphics[width=0.48\columnwidth]{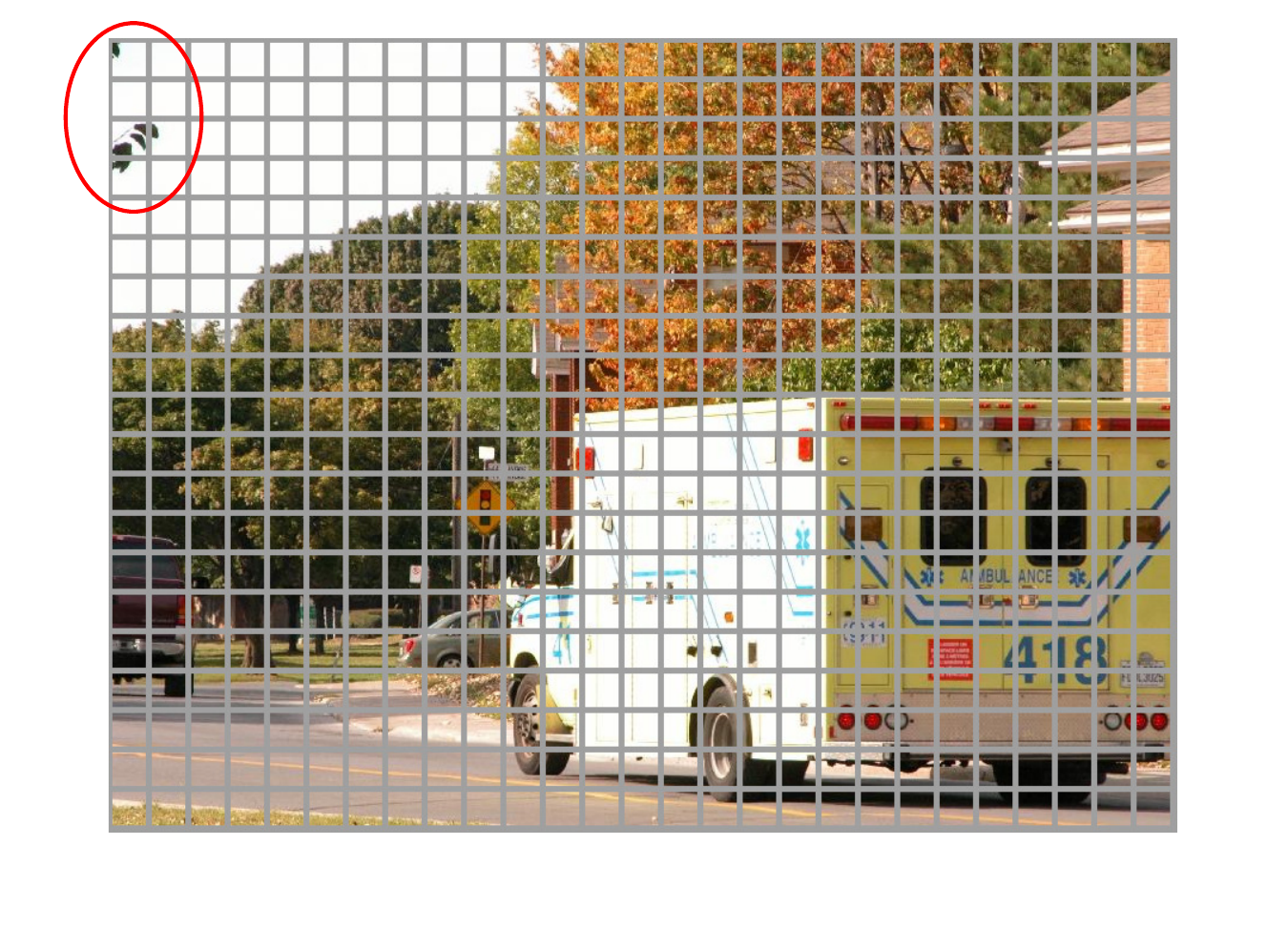}
\end{center}
\caption{Examples of square jigsaw puzzles, where the comparison of two neighboring patches is challenging or impossible. The top left image shows a puzzle with $432$ pieces, each of size $28 \times 28$.
The top right image demonstrates an example of $2$ neighboring patches in the latter puzzle that have different pixel values around the boundaries due to the discrete nature of a digital image.
These patches are circled with red in the original puzzle (top left image) and their nearby sides are circled
with red in the top right image.
The bottom two images demonstrate examples of puzzles that have patches with uniformly white sides (circled with red in the bottom left image) and also have some uniformly white patches.
%The bottom right image has 2 components that are disconnected by uniformly white sky.
Natural solutions of the bottom left puzzle seem to  yield visually correct images that may not coincide with the original assignment.
However, there are natural solutions of the bottom right puzzle that result in different images than the original one. Indeed, the small component of the image circled with red can be placed in different area within the skies.}
\label{fig:bad_nb}
\end{figure}

We recall that the formulation of the square jigsaw puzzle problem requires finding a permutation $\sigma$ and rotations $\{R_i\}_{i=1}^n \subset \ZZ_4$.
Equivalently, one may solve for locations $\{\vx_i\}_{i=1}^n$ on a uniform grid, representing the centers of the patches, and rotations $\{R_i\}_{i=1}^n$.
In order to estimate these from the set of image patches $\cQ$ with a function $f$, one needs to rely on the similar function values on the sides of neighboring patches.
However, in our setting of digital images, $f$ is often discontinuous in the direction from one side of a patch to a side of a neighboring patch.
The top right image of \Cref{fig:bad_nb} demonstrates this phenomenon for two patches selected from the puzzles shown in top left image with lower resolution.
Such discontinuity can result in loss of information for determining neighbors and may lead to ill-posed problems.

There are also special images for which the puzzle problem is ill-posed.
For example, the bottom left image of \Cref{fig:bad_nb} demonstrates a case where several patches look very similar to each other and it is impossible to determine the right permutation.
Nevertheless, the output of common algorithms given this particular puzzle is often visually acceptable.
On the other hand, the bottom right image of \Cref{fig:bad_nb} demonstrates a case where the image consists of two parts that are disconnected by a uniform background. The background is the white sky, one part is the main scene of the image and the other part includes two short branches of another tree at the top left corner of the image.
In this case, it would be impossible to figure out the exact position of the latter part of the image.

The following definition quantifies an ideal type of metric between sides of image patches that, if exists (i.e., if the problem is well-posed), can be used to solve the square jigsaw puzzle problem.
\begin{definition}
\label{defn:perfect}
Fix an image $I$ and a set of image patches $\cQ:=\{P_i, f|_{P_i}\}_{i=1}^n$. A metric defined on all sides of image patches in $\cQ$ is called perfect if there exists $c>0$ so that any two
matching sides (of neighboring patches) have a distance less than $c$ and any two non-matching sides have a distance greater than $c$.
\end{definition}

The main challenge of solving reasonable instances of the square jigsaw puzzle problem is to find a nearly perfect metric.
Empirically, we have found that the Mahalanobis Gradient Compatibility (MGC) metric, defined in \cite{gallagher2012jigsaw} and described in  \S\ref{sec:mgc}, is often near perfect in well-posed cases.

\section{A Framework for Recovering Rotations of Puzzle Pieces}
\label{sec:rot_rec}

This section applies the framework of \cite{singer2012vector, el2016graph} for recovering the global orientations of puzzle patches.
This framework requires the construction of a graph whose vertices correspond to the puzzle patches and whose edges connect neighboring patches.
The rest of the section is organized as follows: \S\ref{sec:GCL} forms the  connection graph and its graph connection Laplacian (GCL) and explains how to estimate the rotations of puzzle patches using this Laplacian; and \S\ref{sec:theoretical_justification} theoretically justifies the method described in \S\ref{sec:GCL}.

\subsection{Estimation of Orientations Using the Connection Graph}
\label{sec:GCL}

The general connection graph~\cite{singer2012vector} $G = (V, E, W, R)$ consists of four components: vertices $V,$ edges $E,$ the {\it affinity function} (or weight function) $W : E \to [0, 1]$ and the {\it connection function} $R : E \to \GG,$ where $\GG$ is a given group.
The first three components are determined by the weighted graph and the fourth depends on the application in which the graph is used.
This formulation is most natural for the problem of group synchronization, which is carefully reviewed in the first two sections of \cite{lerman_shi2019robust}.
In this problem, one is given a graph $G = (V, E)$ with affinity function $W$ and needs to estimate for any vertex $i \in V$ a group element $g_i \in \GG$  from corrupted (or noisy) measurements of group ratios $g_i g_j^{-1} \in \GG$ among all
$\{i,j\} \in E$. It is natural to form the connection function by the given measurements.

In the particular case of the square jigsaw puzzle, $\GG = \ZZ_4$ and the connection function for any edge $\{i, j\}$ assigns the rotation represented by the block $\mR[i, j]$, which was defined in \S\ref{sec:GCL}. For convenience, we represent the affinity and connection functions by their corresponding matrices $\mW \in \RR^{n \times n}$ and $\mR \in \RR^{2n \times 2n}$, and thus write the connection graph as $G = (V, E, \mW, \mR)$.
In this case, the group synchronization problem is referred to as $\ZZ_4$ synchronization. This is the underlying mathematical problem for recovering the orientations of the patches. However, there are several practical considerations that makes the problem of orientation recovery of puzzle patches more complicated than the synthetic $\ZZ_4$ synchronization problem. A primary issue is that one needs to find a way to measure the group ratios, that is, a connection function needs to be estimated from the given puzzle patches. Another issue is that erroneous location assignments of patches may result in poorly estimated connection function.

If one has a perfect metric (recall Definition~\ref{defn:perfect}) for the square jigsaw puzzle, the ideal connection graph is formed as follows. The vertices represent patches in $\cQ$, the edges connect neighboring patches and the weights are $1$ for all edges and $0$ otherwise.
The underlying group $\ZZ_4$ can be represented either by the four complex numbers $\{1, i, -1, -i\}$ with complex multiplication or by the following four $2 \times 2$ matrices:
\begin{equation}
\label{eq:represent_rotations}
  \begin{bmatrix}
    1 & 0 \\
    0 & 1
  \end{bmatrix},
  \begin{bmatrix}
    0 & -1 \\
    1 & 0
  \end{bmatrix},
  \begin{bmatrix}
    -1 & 0 \\
     0 & -1
  \end{bmatrix},
  \begin{bmatrix}
    0 & 1 \\
    -1 & 0
  \end{bmatrix}
\end{equation}
with matrix multiplication.
Note that if $i < j$ and $\{i,j\}$ is an edge connecting patches $P_i$ and $P_j$, then $R[i,j]$ is a rotation in $\GG=\ZZ_4$ whose application to $P_j$, together with an appropriate  plane translation, results in $P_i$. Throughout the paper we use the representation of $\ZZ_4$ in \eqref{eq:represent_rotations}.

For possibly imperfect scenarios of the square jigsaw puzzles, the vertices are formed as above,
but one needs to construct meaningful edges, affinity function and connection function (with $\GG = \ZZ_4$).
A heuristic construction of these is suggested for type 2 and type 3 puzzles in \S\ref{sec:conn_graph_const_Type2} and \S\ref{sec:conn_graph_const_Type3}, respectively.
Here we propose a general heuristic that uses a given connection graph of square jigsaw puzzles to estimate the unknown orientations of the patches.
This heuristic is later justified in \S\ref{sec:theoretical_justification} under special assumptions.
The main idea of this heuristic is to use the GCL for inferring global information (in the form of a certain eigendecomposition) from local information (needed to form the GCL).

Next, we review several matrices associated with a general connection graph.
Recall that the functions $W$ and $R$ are defined on the set $\{1,\dots,n\} \times \{1,\dots,n\}$, where $n$ is the number of puzzle pieces. Thus, from now on, we denote these functions by their corresponding matrices $\mW \in \RR^{n \times n}$ and $\mR \in \RR^{2n \times 2n}$, respectively.
Note that $\mR$ is a block matrix whose $2 \times 2$ blocks represent two-dimensional rotations. For $1 \leq i, j \leq n$, we denote by $\mR[i, j]$ the $[i, j]$-th $2 \times 2$ block of $\mR$.
We index blocks by $[i,j]$ and matrix elements by $(i,j)$.
The connection graph is thus $G = (V, E, \mW, \mR)$.
The {\it connection adjacency matrix} is an $n\times n$ block matrix $\mS$ with $2\times 2$ submatrices, where for $1 \leq i, j \leq n$ the $(i,j)$-th submatrix is
\begin{align}
\label{eq:vdmS}
\mS[i, j]=\left\{
\begin{array}{ll}
\mW(i,j) \mR[i,j], & \mbox{if }\{i,j\}\in E;\\
\vzero,&\mbox{otherwise.}
\end{array}\right.
\end{align}
The {\it degree matrix} is an $n\times n$ block diagonal matrix $\mD,$ where for $1 \leq i \leq n$, its $i$-th diagonal submatrix is
\begin{align}
\label{eq:vdmD}
\mD[i, i]=d(i) \mI_2, \text{ where } d(i) = \sum_{j\neq i} \mW(i,j),
\end{align}
where $\mI_2$ is the $2\times 2$ identity matrix.
We define $\mC:=\mD^{-1}\mS$ as the graph connection weight (GCW) matrix. We define the (normalized) GCL matrix as $\mI-\mC$. In practice, we directly form the GCW matrix and use its eigendecomposition. Clearly, this is equivalent to using the eigendecomposition of the GCL matrix, and we thus refer to or name our method with the term GCL.

The GCW matrix is associated with a random walk, whose transition probability matrices are $\mW(i,j)$, $1 \leq i, j \leq n$. This can be seen by its action on a block vector $\vv\in \RR^{2n \times 2}$, whose $n$-th $2\times 2$ submatrices are
\begin{equation*}
\vv[j] = \begin{bmatrix}
    \vv_{2j-1, 1} & \vv_{2j-1, 2} \\
    \vv_{2j, 1} & \vv_{2j, 2}
  \end{bmatrix} \in \RR^{2\times 2}
, \ 1 \leq j \leq n\,,
\end{equation*}
in the following way
\begin{equation*}
\label{Cmatrix}
(\mC \vv)[i] = \sum_{j: (i,j)\in E} \left[\frac{\mW(i,j)}{\sum_{k: (i,k)\in E} \mW(i,k)}\right] \mR[i,j]\vv[j]\,.
\end{equation*}
That is, a block vector $\vv[j]$ is rotated by $\mR[i,j]$ and assigned to the $i$-th patch with probability ${\mW(i,j)}/{\sum_{k: (i,k)\in E} \mW(i,k)}$.

To recover the global orientations of the puzzle patches, we follow the procedures of \cite{singer2012vector,bandeira2013cheeger}. First, we form the block vector $\mU \in \RR^{2n \times 2}$ whose columns are the top $2$ eigenvectors of $\mC$. Then, we project each of the $2 \times 2$ blocks of $\mU$ onto $\ZZ_4$ (that is, we replace each block with its closest element, with respect to the Frobenius norm, in \eqref{eq:represent_rotations}) and use the resulting blocks
as the global orientations. \Cref{algo:rot_sol} summarizes the above straightforward procedure of recovering the unknown orientations of the image patches for a given square jigsaw puzzle.
\begin{algorithm}
\caption{The GCL Algorithm}
\label{algo:rot_sol}
\begin{algorithmic}
\State{\textbf{Input:} Connection graph: $G = (V, E, \mW, \mR)$}
\begin{itemize}
\State Construct the Connection Adjacency Matrix $\mS$ by \eqref{eq:vdmS}
\State Construct the degree matrix $\mD$ by \eqref{eq:vdmD}
\State Let $\mC = \mD^{-1} \mS$
\State Form $\mU \in \RR^{2n \times 2}$ whose columns are the 2 top eigenvectors of $\mC$
\State For $1 \leq i \leq n$, let $\mR_i \in \ZZ_4$ be the projection of the $i$-th block of $\mU$ onto $\ZZ_4$
\end{itemize}
\State \textbf{Return:} Global rotation matrices  $ \mR_1$,  $\ldots$,  $\mR_n$
\end{algorithmic}
\end{algorithm}

We emphasize that the GCL algorithm for recovering the orientations of patches is non-greedy.
Indeed, it directly constructs the orientation of patches using the information in the connection graph via diffusion.
On the other hand, other methods, such as \cite{pomeranz2011fully, gallagher2012jigsaw, son2014solving, son2016solving}, try to greedily match pieces based on their relative orientations.
We also mention that the GCL algorithm does not use any knowledge of the size of the puzzle image, or equivalently, of the number of puzzle pieces per length or width of the image.

\subsection{Theoretical Justification of the GCL Algorithm}
\label{sec:theoretical_justification}

We show that the proposed estimation of orientations for type 2 puzzles is robust to incorrect measurements, where incorrect measurements are mistakes in estimating the connection graph.
The three puzzles in \Cref{fig:bad_nb} exemplify cases where incorrect measurements are expected
due to indistinguishability or low-resolution of patches. Incorrect measurements can also arise due to mistakes in estimating patches' locations. Indeed, such mistakes result in incorrect estimation of the connection graph.

We distinguish between the ground truth solution (or ``true'' solution) and the estimated one. We denote by $E_{\true}$ the set of ``true edges'', that is, edges connecting neighboring patches of the true solution.
We find it most natural to define $(V,E_{\true})$ as the underlying uniform grid for the patches.
Nevertheless, there is some freedom in defining $(V,E_{\true})$. For example, if one wants to also emphasize diagonal edges, then $(V,E_{\true})$ can be formed by adding these edges to the uniform grid. These two different choices of $(V,E_{\true})$ can make a slight difference in the estimates of our proposed theory described below.
Using the prefixed indexing of patches, we define the true affinity function
\begin{equation}
\label{eq:aff_func}
\mW_{\true} (i, j) =
			\begin{cases}
               1, \text{ if } \{i, j\} \in E_{\true}; \\
               0, \text{ otherwise,}
            \end{cases}
\end{equation}
and the true connection function
\begin{equation}
\label{eq:con_func}
\mR_{\true} [i, j] =
			\begin{cases}
               \mR_i \mR_j^T, \text{ if } \{i, j\} \in E_{\true};\\
               \boldsymbol{0}, \text{ otherwise,}
            \end{cases}
\end{equation}
where $\mR_1$, $\ldots$, $\mR_n$ are the rotation matrices of the rotations $R_1$, $\ldots$, $R_n$ defined in \S\ref{subsec:math_special}.
Note that unlike $E_{\true}$, $\mW_{\true}$ and $\mR_{\true}$ are unknown to the user.
Let $G_{\est} = (V, E_{\est}, \mW_{\est}, \mR_{\est})$ denote the estimated connection graph.
We remark that the graph $(V,E_{\est})$ may be rather different than $(V,E_{\true})$ as the user can have some uncertainties about connecting patches.
Finally, denote by $\mC_{\est}$ the GCW matrix corresponding to $G_{\est}$.

The following perturbation theorem states that if the estimated connection graph is a good approximation of the true connection graph and certain conditions hold, then the estimated rotations are close in some sense to the underlying rotations.
To be more specific, we denote by $\mV_{\est} \in \RR^{2n \times 2}$ the matrix whose 2 columns are the top eigenvectors of $\mC_{\est}$. Recall that the estimated rotations are obtained by projecting the blocks of this matrix onto $\ZZ_4$. We denote the set of underlying rotations by $\mR_1, \dots, \mR_n \in \ZZ_4$. We further denote by $\mR_{\true}^{\mathrm{tot}}$ the block matrix in $\RR^{2n \times 2}$ whose $n$ blocks
are these underlying rotations.
The theorem claims that under some conditions, the principal angles between the column spaces of $\mV_{\est}$ and $\mR_{\true}^{\mathrm{tot}}$ are sufficiently close.
Note that there are only two such angles, $\theta_1$ and $\theta_2$, and recall that they can be computed as follows:  $\sin(\theta_1) = \cos^{-1} (\sigma_1)$ and $\sin(\theta_2) = \cos^{-1}(\sigma_2)$, where $\sigma_1$ and $\sigma_2$ are the singular values of $\mV_{\est}^T \mR_{\true}^{\mathrm{tot}}$.

We use the following notation: $\sin(\Theta(\mR_{\true}^{\mathrm{tot}}, \mV_{\est}))$ denotes the $2 \times 2$ diagonal matrix with diagonal elements $\sin(\theta_1)$ and $\sin(\theta_2)$ (specified above);
$\Vert \cdot \Vert_{F}$ denotes the Frobenius norm; $d_{\max}$ denotes the maximal degree of the estimated graph (using the $\infty$ matrix norm, we can express it as $d_{\max} = \|\mW_{\est}\|_{\infty}$);
for a set $E' \subseteq E$, $\phi_{E'}$ denotes the second smallest eigenvalue of the normalized graph Laplacian of the graph $G' = \left( V, E' \right)$ (a precise formula for $\phi_{E'}$
is given at the end of the proof of the theorem).
Using this notation we express below the closeness of $\mV_{\est}$ and $\mR_{\true}^{\mathrm{tot}}$, which we further discuss and interpret after proving the theorem.
%also implies closeness of the estimated rotations (projections of blocks of $\mV_{\est}$  onto $\ZZ_4$) to the underlying ones (blocks of $\mR_{\true}^{\mathrm{tot}}$).
%, so $\Vert \sin \Theta(\mV'_{\true}, \mV_{\est}) \Vert_{F} = \sqrt{\sin^2(\theta_1)+\sin^2(\theta_2)}$.

\begin{theorem}
\label{theorem:perturbation}
Let $G_{\true} = (V, E_{\true}, \mW_{\true}, \mR_{\true})$ be the ground-truth connection graph, where $\mW_{\true}$ and $\mR_{\true}$ are defined in \eqref{eq:aff_func} and \eqref{eq:con_func}.
Assume that a user estimates this connection graph from possibly corrupted data with the following connection graph $G_{\est} = (V, E_{\est}, \mW_{\est}, \mR_{\est})$, which has maximal degree $d_{\max}$, and that there exists a set $E' \subset E_{\true} \cap E_{\est}$ satisfying the following properties: $(V, E')$ is a connected graph;
\begin{equation}
\mR_{\est} [i, j] = \mR_{\true}[i, j]  \text{ for }  \{i, j\} \in E';
\label{eq:cond_lemma_0}
\end{equation}
there exist $\epsilon > 0$ and $c_1$, $\ldots$, $c_n>0$ such that
\begin{equation}
\label{eq:cond_lemma}
\max_{ 1 \le i \le n} \frac{1}{\sqrt{d_{\max}}} \, \sqrt{ \sum_{j : \{i, j \} \in E'} \left( \mW_{\true}(i, j) - \frac{\mW_{\est}(i, j)}{c_i} \right)^2 +  \sum_{j : \{ i, j\} \notin E'} \left( \frac{\mW_{\est}(i, j)}{c_i} \right)^2} < \epsilon;
\end{equation}
and there exists $\gamma > \epsilon$ such that
\begin{equation}
\label{eq:cond_lemma_2}
\min_{1 \leq i \leq n} \sum_{j : \{i,j\} \in E'} \mW_{\true} (i, j) / d_{\max} > \gamma.
\end{equation}
Then,
\begin{equation}
\label{eq:final_est}
\Vert \sin \Theta(\mR_{\true}^{\mathrm{tot}}, \mV_{\est}) \Vert_{F} \le \frac{2}{\phi_{E'}}\sqrt{\frac{2 n}{d_{\max}}} \, \frac{ \epsilon}{\gamma} \left(1 +  \frac{1}{\gamma - \epsilon} \right).
\end{equation}
\begin{comment}
\begin{equation}
\label{eq:final_est}
\Vert \sin \Theta(\mR_{\true}^{\mathrm{tot}}, \mV_{\est}) \Vert_{F} \le \frac{\sqrt{\frac{2 n}%{d_{\max}}} \, \frac{ \epsilon}{\gamma} \left(1 +  \frac{1}{\gamma - \epsilon} \right)} %{\lambda_1(\mW_{\true}) - \lambda_2(\mW_{\true})) }.
\end{equation}
\end{comment}
\end{theorem}

\begin{proof}
We denote the restriction of the affinity and connection functions,  $\mW_{\true}$ and $\mR_{\true}$, onto the edges in $E'$ by  $\mW'_{\true}$ and $\mR'_{\true}$, respectively. The corresponding connection graph, connection adjacency matrix, diagonal matrix and GCW matrix are denoted by $G'_{\true} = (V, E', \mW'_{\true}, \mR'_{\true})$, $\mS'_{\true}$, $\mD'_{\true}$ and $\mC'_{\true} \in \RR^{2n \times 2n}$, respectively.
In addition of the above notation for the estimated connection graph, we denote its connection adjacency matrix and diagonal matrix by $\mS'_{\true}$ and $\mD'_{\true}$, respectively.
Following \eqref{eq:vdmD}, we denote the diagonal elements of
$\mD'_{\true}$ and $\mD_{\est}$ by $d_{\true}(i)$ and $d_{\est}(i)$, $i = 1$, $\ldots$, $n$. Note that the maximal degree of the estimated graph can be expressed as follows: $d_{\max} = \max_{1 \le i \le n} d_{\est}(i)$.
Let $\mV'_{\true} \in \RR^{2n \times 2}$ denote the matrix whose 2 columns are the top eigenvectors of $\mC'_{\true}$.

We note that division of the $i$-th row of the affinity matrix $\mW_{\est}$ by the constant $c_i$, where $1 \leq i \leq n$, does not change the GCW matrix of $(V, E, \mW_{\est}, \mR_{\est})$. Thus, we can assume, without loss of generality, that $c_1=\ldots=c_n=1$. Using this assumption and the definition of $\mW'_{\true}$ we rewrite \eqref{eq:cond_lemma} as
\begin{equation}
\label{eq:cond_lemma_new}
\max_{ 1 \le i \le n} \sqrt{\frac{1}{d_{\max}} \sum_{j = 1}^n \left( \mW'_{\true}(i, j) - \mW_{\est}(i, j) \right)^2 } < \epsilon.
\end{equation}

We further note that the assumption in \eqref{eq:cond_lemma_2} can be rewritten as
\begin{equation}
\label{eq:cond_lemma_2_revised}
d'_{\true} (i) / d_{\max} \ge \gamma \ \text{ for all } \ 1 \leq i \leq n.
\end{equation}

The proof of \Cref{theorem:perturbation} consists of three steps.

\noindent
{\bf Step I:} We prove that
\begin{equation}
\label{eq:c_true_est}
\Vert \mC'_{\true} - \mC_{\est} \Vert_F \le \sqrt{\frac{2 n}{d_{\max}}} \, \frac{ \epsilon}{\gamma} \left(1 +  \frac{1}{\gamma - \epsilon} \right).
\end{equation}
This result is analogous to Lemmas~3.1 and~3.2 of El Karoui and Wu~\cite{el2016graph}, but uses the Frobenius norm instead of the spectral norm and has weaker conditions in \eqref{eq:cond_lemma} and \eqref{eq:cond_lemma_2}. El Karoui and Wu~\cite{el2016graph} also allow a very small perturbation of the connection function restricted to $E'$, but in the special case of $\ZZ_4$, this assumption is equivalent with \eqref{eq:cond_lemma_0}.
% In fact, its proof is parallel to the proofs of the latter lemmas and is thus omitted here. The result obtained is

Using the definitions of $\mC'_{\true}$ and $\mC_{\est}$ we express and then bound the LHS of \eqref{eq:c_true_est} as follows
\begin{multline}
\label{eq:orig_estimate}
\Vert {\mD'}_{\true} ^{-1} \mS'_{\true} -  \mD_{\est}^{-1} \mS_{\est}  \Vert _F= \Vert {\mD'}_{\true} ^{-1} \left( \mS'_{\true} - \mS_{\est} \right)  + \left(  {\mD'}_{\true} ^{-1} - \mD_{\est}^{-1} \right) \mS_{\est}  \Vert_F \le \\ \Vert {\mD'}_{\true} ^{-1} \left( \mS'_{\true} - \mS_{\est} \right)\Vert_F  + \left\Vert \left(  {\mD'}_{\true} ^{-1} - \mD_{\est}^{-1} \right) \mS_{\est}  \right\Vert_F.
\end{multline}

We follow with bounding the first term of the RHS of \eqref{eq:orig_estimate}:
\begin{multline}
\label{eq:est_first_sum}
\left \Vert {\mD'}_{\true} ^{-1} \left( \mS'_{\true} - \mS_{\est} \right) \right\Vert_F =
\left\Vert \left( {\mD'_{\true}} / d_{\max} \right) ^{-1} \left( \mS'_{\true}/d_{\max} - \mS_{\est}/d_{\max} \right)\right\Vert_F \le \\
\left\Vert\left( {\mD'_{\true}} / d_{\max} \right) ^{-1} \right\Vert_2 \left\Vert \left( \mS'_{\true}/d_{\max} - \mS_{\est}/d_{\max} \right) \right\Vert_F.
\end{multline}
We then control the first multiplicative term in the RHS of \eqref{eq:est_first_sum} using %the fact that $\mD'_{\true}$ is diagonal and
\eqref{eq:cond_lemma_2_revised}:
\begin{equation}
\label{eq:D_norm_est}
\left\Vert\left( {\mD'_{\true}} / d_{\max} \right) ^{-1} \right\Vert_2 \le \max_{1 \leq i \leq n} \frac{ d_{\max} } {d'_{\true}(i) } \le \frac{1}{\gamma}.
\end{equation}
We next control the second multiplicative term in the RHS of \eqref{eq:est_first_sum}, where we follow with justification:
\begin{equation}
\label{eq:first_sum_first_bound}
\begin{split}
& \Vert \mS'_{\true}/d_{\max} - \mS_{\est}/d_{\max} \Vert^2_F = \frac{1}{d_{\max}^2} \sum_{ i =1 }^n \sum_{ j =1 }^n \Vert \mS'_{\true}[i, j]- \mS_{\est}[i, j] \Vert^2_F = \\
& \frac{1}{d_{\max}^2} \sum_{ i =1 }^n \sum_{ j =1 }^n \Vert \mW'_{\true}(i, j) \left( \mR'_{\true} [i, j] - \mR_{\est} [i, j] \right) + \left( \mW'_{\true}(i, j) - \mW_{\est}(i, j) \right) \mR_{\est}[i, j]  \Vert^2_F = \\
& \frac{1}{d_{\max}^2} \sum_{ i =1 }^n \sum_{ j =1 }^n \left( \mW'_{\true}(i, j) - \mW_{\est}(i, j)  \right)^2 \Vert \mR_{\est}[i, j]  \Vert^2_F =
\\
& \frac{2}{d_{\max}^2} \sum_{ i =1 }^n \sum_{ j =1 }^n \left( \mW'_{\true}(i, j) - \mW_{\est}(i, j)  \right)^2 \le
\frac{2n}{d^2_{\max}} \max_{1 \le  i \le n } \sum_{ j =1 }^n \left( \mW'_{\true}(i, j) - \mW_{\est}(i, j)  \right)^2  \le \frac{2 n \epsilon^2}{d_{\max}}.
\end{split}
\end{equation}
The third equality uses \eqref{eq:cond_lemma_0} and the fact that $\mW'_{\true} = 0$ for $\{i,j\} \notin E'$, the
fourth equality uses the fact that $\Vert \mR_{\est} [i, j]\Vert_F^2 = 2$, and the last inequality uses \eqref{eq:cond_lemma_new}.
Combining \eqref{eq:est_first_sum},  \eqref{eq:D_norm_est} and \eqref{eq:first_sum_first_bound} we bound the first term of the RHS of \eqref{eq:orig_estimate}:
\begin{equation}
\label{eq:sum_1_bound}
\left \Vert {\mD'}_{\true} ^{-1} \left( \mS'_{\true} - \mS_{\est} \right) \right\Vert_F \le
\sqrt{\frac{2n}{d_{\max}}} \frac{\epsilon}{\gamma}.
\end{equation}

We control the second term in the RHS of \eqref{eq:orig_estimate} below in \eqref{eq:sum_2_bound}. In order to pursue this, we need to bound several terms. We start with the following bound and then justify it
\begin{equation}
\label{eq:est_d_diff}
\begin{split}
& \left\Vert \mD'_{\true}/d_{\max} - \mD_{\est}/d_{\max} \right\Vert_2 = \max_{1 \leq i \leq n} \left \vert d'_{\true}(i)/d_{\max} - d_{\est}(i)/d_{\max} \right\vert = \\
& { \max_{1 \leq i \leq n} \frac{1}{d_{\max}} \left \vert  \sum_{j = 1}^n  \left( \mW'_{\true}(i, j) - \mW_{\est}(i, j)\right) \right \vert } \leq \\
& \max_{1 \leq i \leq n} \sqrt{\frac{1}{d_{\max}} \sum_{j = 1}^n \left( \mW'_{\true}(i, j) - \mW_{\est}(i, j)\right)^2} \le \epsilon.
\end{split}
\end{equation}
The first inequality is a direct application of Cauchy--Schwarz and the fact that for each $1 \le i \le n $ at most $d_{\max}$ of $\mW'_{\true}(i, j) - \mW_{\est}(i, j)$ are non-zero. In order to realize the last fact, we recall that $E' \subseteq E_{\est}$ so we only need to check this fact when $\{i,j\} \in E_{\est} \setminus E'$. In this case, $\mW'_{\true}(i, j)=0$ and $\mW_{\est}(i, j)$ satisfies the required property due to \eqref{eq:cond_lemma_2_revised}.
The last inequality follows from \eqref{eq:cond_lemma_new}.
Note that \eqref{eq:est_d_diff} implies that $d_{\est} (i) / n \ge d'_{\true} (i) /n- \epsilon$ for $1 \le i \le n$. Applying this inequality and then \eqref{eq:cond_lemma_2_revised},
we obtain that
\begin{multline}
\label{eq:D_est_inv_est}
\left\Vert\left( {\mD_{\est}} / d_{\max} \right) ^{-1} \right\Vert_2 = \max_{1 \le i \le n} \left( d_{\est} (i)/ d_{\max} \right)^{-1} = \frac{1}{\min_{1 \le i \le n} d_{\est} (i) / d_{\max}} \\
\le \frac{1}{\min_{1 \le i \le n} \left(d'_{\true} (i) /d_{\max} - \epsilon\right)} \le \frac{1}{\gamma - \epsilon}.
\end{multline}

Using the facts that for all $1 \leq i$, $j \leq n$,  $0 \le \mW_{\est} (i, j)  \le 1$ and $\left\Vert \mR_{\est}[i, j] \right\Vert^2_F = 2$, we conclude that
\begin{equation}
\label{eq:S_est_final}
\left\Vert \mS_{\est} / d_{\max} \right\Vert^2_F \le \max_{1 \leq i, j \leq n} \frac{n}{d_{\max}}\left\Vert \mS_{\est} [i, j] \right\Vert^2_F =
\frac{n}{d_{\max}} \max_{1 \leq i, j \leq n} \left\vert \mW_{\est} (i, j) \right\vert^2  \left\Vert \mR_{\est}[i, j] \right\Vert^2_F \le \frac{2n}{d_{\max}}.
\end{equation}
The above first inequality is due to the fact that each column and row of $\mS_{\est}$ contains at most $d_{\max}$ non-zero elements, so the LHS is bounded by $n d_{\max}$ times the maximal squared Frobenius norm of a block. The latter fact follows from the argument described below \eqref{eq:est_d_diff}. Combining \eqref{eq:D_norm_est}, \eqref{eq:est_d_diff}, \eqref{eq:D_est_inv_est} and \eqref{eq:S_est_final}, we bound the second term in the RHS of \eqref{eq:orig_estimate} as follows
\begin{equation}
\label{eq:sum_2_bound}
\begin{split}
& \left\Vert \left(  {\mD'}_{\true} ^{-1} - \mD_{\est}^{-1} \right) \mS_{\est}  \right\Vert_F \le
\left\Vert \mS_{\est}/d_{\max} \right\Vert_F \left\Vert \left(  \left({\mD'}_{\true} /d_{\max} \right)^{-1} - \left(\mD_{\est}/d_{\max} \right)^{-1} \right)  \right\Vert_2 \le \\
& \left\Vert \mS_{\est}/d_{\max} \right\Vert_F \left\Vert \left({\mD'}_{\true} / d_{\max} \right)^{-1} \left({\mD'}_{\true} / d_{\max} - \mD_{\est}/d_{\max} \right) \left(\mD_{\est}/d_{\max}\right)^{-1}  \right\Vert_2 \le  \\
& \left\Vert \mS_{\est}/d_{\max} \right\Vert_F \left\Vert \left({\mD'}_{\true} / d_{\max} \right)^{-1} \right\Vert_2 \left\Vert \left({\mD'}_{\true} /d_{\max} - \mD_{\est}/d_{\max} \right) \right \Vert _2\left\Vert \left(\mD_{\est}/d_{\max}\right)^{-1}  \right\Vert_2 \le \\
& \sqrt{\frac{2 n}{d_{\max}}} \frac{\epsilon}{\gamma (\gamma - \epsilon)}.
\end{split}
\end{equation}
Clearly, \eqref{eq:orig_estimate}, \eqref{eq:sum_1_bound} and \eqref{eq:sum_2_bound} imply \eqref{eq:c_true_est}.

\noindent
{\bf Step II:} This step uses the Davis-Kahan $\sin \Theta$ theorem \cite{davis1970rotation} to prove the following inequality, where $\lambda_2(\mC'_{\true})$ and $\lambda_3(\mC'_{\true})$ denote the second and third largest eigenvalues of $C'_{\true}$, respectively:
\begin{equation}
\label{eq:final_est_modify}
\Vert \sin \Theta(\mV'_{\true}, \mV_{\est}) \Vert_{F} \le \frac{2 \sqrt{\frac{2 n}{d_{\max}}} \, \frac{ \epsilon}{\gamma} \left(1 +  \frac{1}{\gamma - \epsilon} \right)} {\lambda_2(\mC'_{\true}) - \lambda_3(\mC'_{\true}) }.
\end{equation}
We use a specific variant of Davis-Kahan according to \cite{yu2015useful}. This variant implies that
\begin{multline}
\label{eq:eig_bound_final}
\Vert \sin \Theta(\mV'_{\true}, \mV_{\est}) \Vert_{F} \le
\\
\frac{2 \min \left( \sqrt{2} \Vert\mC'_{\true} - \mC_{\est}\Vert_2, \Vert\mC'_{\true} - \mC_{\est}\Vert_F\right) }{\lambda_2(\mC'_{\true}) - \lambda_3(\mC'_{\true})}
\le \frac{2 \Vert\mC'_{\true} - \mC_{\est}\Vert_F}{\lambda_2(\mC'_{\true}) - \lambda_3(\mC'_{\true})}.
\end{multline}
The combination of \eqref{eq:c_true_est} and \eqref{eq:eig_bound_final} implies \eqref{eq:final_est_modify}.

\noindent
{\bf Step III:} This step shows that $\mV'_{\true} = \mR_{\true}^{\mathrm{tot}}$ and $\lambda_2(\mC'_{\true}) - \lambda_3(\mC'_{\true}) = \phi_{E'}$ and  thus in view of \eqref{eq:final_est_modify} it concludes the proof of \eqref{eq:final_est}.
We denote by ${\hat{\mD}_{\true}}' \in \RR^{n \times n}$ the reduction of the matrix ${{\mD'}_{\true}} \in \RR^{2n \times 2n}$ obtained  by replacing each of its $2 \times 2$ scalar blocks,  $\{d'_{\true}(i) \mI_2\}_{i=1}^n$, with diagonal elements, $\{d'_{\true}(i)\}_{i=1}^n$.
The corresponding normalized weight matrix is ${\widetilde{\mW}_{\true}}' = {\hat{\mD}_{\true}}'^{-1} \mW'_{\true}$.
Clearly, the largest eigenvalue of ${\widetilde{\mW}_{\true}}'$ is 1, it has multiplicity 1 (since $E'$ is connected) and its eigenspace is spanned by a column vector of ones in $\RR^n$, which we denote by $\vone_{n \times 1}$.

We next relate the eigendecomposition of ${\widetilde{\mW}_{\true}}'$ to that of $\mC'_{\true}$.
Let $\lambda_1, \dots \lambda_n$ denote the eigenvalues of ${\widetilde{\mW}_{\true}}'$ in
decreasing order and let $\vl_1, \dots \vl_n \in \RR^n$ denote the corresponding orthogonal eigenvectors, each with norm $\sqrt{n}$.
%Recall that $\lambda_1=1$ and $\vl_1=\vone_{n \times 1}$.
Thus, for all $1 \le i$, $k \le n$, $\sum_{j = 1}^n {\widetilde{\mW}_{\true}}'(i, j) \vl_k(j) = \lambda_k \vl_k(i)$. This equation is equivalent to the following one: $\sum_{j = 1}^n {\widetilde{\mW}_{\true}}'(i, j) \mR_i \mR_j^T \mR_j \vl_k(j) = \lambda_k \vl_k(i) \mR_i$. Since $\mC'_{\true}[i, j] = {\widetilde{\mW}_{\true}}'(i, j) \mR_i \mR_j^T$ for all $1 \le i, j \le n$, the last equation can be written as
$$\sum_{j = 1}^n \mC'_{\true}[i, j] \vl_k(j) \mR_j  = \lambda_k \vl_k(i) \mR_i.$$ This equation can be further written as $\mC'_{\true}[i, j]  \mU_k = \lambda \mU_k$, where
for $1 \le k \le n$, $\mU_k \in \RR^{2n \times 2}$ satisfies $\mU_k[i] = \vl_k(i) \mR_i$ for $1 \le i \le n$. Note that the $2n$ columns of $\mU_1$, $\ldots$, $\mU_n$ form an orthogonal system and that we
obtained a one-to-one correspondence between the eigenvalues and eigenvectors of ${\widetilde{\mW}_{\true}}'$ and $\mC'_{\true}$.

A first implication of the above property is that 1 is also the largest eigenvalue of $\mC'_{\true}$ with multiplicity 2. Furthermore, since $\vl_1=\vone_{n \times 1}$, $\mU_1 = \mR_{\true}^{\mathrm{tot}}$. By definition,
$\mV'_{\true} = \mU_1$ and thus, as claimed, $\mV'_{\true} =  \mR_{\true}^{\mathrm{tot}}$. Another implication of this property is that $\lambda_2(\mC'_{\true}) - \lambda_3(\mC'_{\true}) = \lambda_1({\widetilde{\mW}_{\true}}') - \lambda_2({\widetilde{\mW}_{\true}}')
= 1 - \lambda_2({\widetilde{\mW}_{\true}}')$. Recall that $\phi_{E'}$ is the second smallest eigenvalue
of the normalized graph Laplacian, which can be written as $\mI - {\hat{\mD}_{\true}}'^{-1} \mW'_{\true} = \mI - {\widetilde{\mW}_{\true}}'$.
%(or equivalently, of the other normalized graph Laplacian $\mI - {\hat{\mD}_{\true}}'^{-0.5} \mW'_{\true}{\hat{\mD}_{\true}}'^{-0.5}$).
Therefore, $\phi_{E'} = 1 - \lambda_2({\widetilde{\mW}_{\true}}') = \lambda_2(\mC'_{\true}) - \lambda_3(\mC'_{\true})$, as claimed.
\end{proof}

For square jigsaw puzzles, \Cref{theorem:perturbation} implies that if one can construct a connection graph and find a connected subgraph of it that satisfies \eqref{eq:cond_lemma_0}-\eqref{eq:cond_lemma_2} with $\epsilon/\phi_{E'} = O(1/\sqrt{n})$, then \Cref{algo:rot_sol} can nearly recover the correct orientations. More precisely, the estimated rotations of \Cref{algo:rot_sol} are obtained by projection onto $\ZZ_4$ of a block matrix whose column space is sufficiently close to the column space of the block matrix of the underlying rotations.

There are several conditions that need to hold in order to imply the conclusion of the theorem. We review them and discuss whether they are reasonable for our proposed method.
The first and simple condition is actually the most restrictive one for our proposed method. It requires $(V, E')$ to be connected.
In the case of puzzles with uniform regions (such as the ones demonstrated in the bottom of Figure \ref{fig:bad_nb}), the edges in $E_{\est}$ obtained by our proposed method, and possibly most methods, may arbitrarily connect patches in these regions, while maintaining a small degree for each patch. These edges can be very different than the ones of $E_{\true}$ and thus the intersection of $(V,E_{\est})$ with $(V,E_{\true})$ may result in a disconnected graph. Since $E' \subset E_{\true} \cap E_{\est}$, $(V, E')$ will also be disconnected in this case of uniform regions.
Nevertheless, under this setting of uniform regions, the mathematical problem does not have a unique solution and is generally ill-conditioned. All tested algorithms did not perform well in this setting when using standard reconstruction metrics; however, most solved puzzles with uniform regions looked similar to their ground truth solutions. For non-uniform regions of natural images, this condition is more reasonable to ask from our proposed method.
The other requirement is that the connection function is correctly estimated on $(V, E')$ (see \eqref{eq:cond_lemma_0}) and we also find it reasonable for non-uniform regions of natural images.
We note that we can obtain the requirement in \eqref{eq:cond_lemma_2} with $\gamma = \Theta(1)$ if we assume that $d_{\max}$ is sufficiently small and that $(V,E')$ is connected.
Indeed, the latter assumption implies that the LHS of \eqref{eq:cond_lemma_2} is positive, so that $\gamma \geq 1/d_{\max}$. % and thus if $d_{\max}$ is sufficiently small, \eqref{eq:cond_lemma_2} holds with $\gamma = \Theta(1)$. 
Our construction trims many unnecessary edges (see \Cref{sec:conn_graph_const}) so that $d_{\max}$ is either at most 8 or slightly larger than 8, where the typical 8 neighboring edges include the four nearest ones and 4 diagonal ones. Therefore, \eqref{eq:cond_lemma_2} with $\gamma = \Theta(1)$ is a reasonable requirement for our proposed method.
At last, we clarify the condition in \eqref{eq:cond_lemma}, where we further discuss the typical size of $\epsilon$ below. First of all, note that the constants $c_1$, $\ldots$, $c_n$ used in this condition
are needed because $\mW_{\true}(i, j)$ obtains values in $\{0,1\}$, whereas $\mW_{\est}(i,j)$ can obtain various nonnegative values.
%As the proof indicates the scaling by $c_1,\ldots,c_n$ is not important,
The underlying assumption of \eqref{eq:cond_lemma} is that the needed proportion $c_i$ for patch $i$ is similar for ``all neighbors''. That is,
when $\{i, j \} \in E'$, the scaled value of $\mW_{\est}(i, j)$ by $c_i$ is close to the binary weight $\mW_{\true}(i, j)$. Moreover, if $\{i, j \} \notin E'$, this scaled value is close to zero.
One may expect such an assumption in some practical instances. For example,  if the image is continuous at a patch, then the affinities with the nearby patches are expected to be all large and comparable.
Similarly, if the image is discontinuous with respect to all neighbors of a patch, then all neighboring affinities are expected to be very small.
Therefore, such a constant $c_i$ may be chosen for each patch. We remark that our proposed algorithm for solving type 2 puzzles (Algorithm \ref{algo:type2_sol}) aims to assign very small affinities whenever there is any possible inconsistency in the construction of the graph connection Laplacian. Such assignment aims to guarantee that the second sum in \eqref{eq:cond_lemma} is sufficiently small. Nevertheless, we cannot really guarantee that our heuristic choices work in practice. We also comment that given the order of $\epsilon$ established below, condition \eqref{eq:cond_lemma} is rather sensitive. Indeed, if for $\{i, j \} \notin E'$, $\mW_{\true}(i, j) = \Theta(1)$, then this condition is violated.

As we mentioned above, in our construction $d_{\max} = \Theta(1)$, where often $d_{\max} \leq 8$ or slightly larger than 8, and thus we can choose $\gamma = \Theta(1)$. Therefore, as long as $\epsilon/\phi_{E'} = O(1/\sqrt{n})$ the bound in  \eqref{eq:final_est} is sufficiently small. That is, the perturbation bound in \eqref{eq:cond_lemma} is more restrictive as the size of the puzzle increases and  $\phi_{E'}$ decreases. To get a better idea of $\phi_{E'}$, we can assume a puzzle grid with lengths and widths of order $\sqrt{n}$. If the graph $(V,E')$ is a lattice, then by direct application of Cheeger's inequality, $\phi_{E'}=\Theta(1/\sqrt{n})$. Similarly, in the worst case of a path, $\phi_{E'}=\Theta(1/n)$.

Few additional remarks are in order.
First of all, the conditions of the theorem are sufficient but not necessary.
Second, one may use in the proof above another variant of Davis-Kahan $\sin \Theta$ theorem according to \cite{fan2017linfinity} and consequently obtain a bound similar to \eqref{eq:final_est}, but controlling the infinity norm, and not the Frobenius norm. Third, while we think of $(V,E_{\true})$ as having a grid-type structure, one can assume in theory a general graph $(V,E_{\true})$.
At last, the theorem can be easily generalized to any group synchronization and not just to $\ZZ_4$ synchronization. This generalization can be possibly applied to other puzzles, such as 3D puzzles or non-squared jigsaw puzzles.

\section{Connection Graph Construction for Type 2 and Type 3 Puzzles}
\label{sec:conn_graph_const}

As we have discussed in \S\ref{sec:GCL}, if we are given a perfect metric, we can easily construct the connection graph.
However, there is no perfect metric that would work for all images.
For example, if part of the image contains a region with a uniform color, such as sky or ocean (see the images on the second row of \Cref{fig:bad_nb}),
the metric between the sides of the image patches from this region will be close to zero. Thus, all these patches should be wrongly identified by a perfect metric as neighbors.
Therefore, the idea of finding a perfect metric and using a threshold to identify neighbors may not lead to a correct affinity graph.
Instead, we suggest to iteratively update the graph construction, while identifying possibly incorrect edges and reassigning zero or small affinities to them.
Nevertheless, our initial estimate of the graph affinities is based on the Mahalanobis Gradient Compatibility (MGC) of \cite{gallagher2012jigsaw}, since empirically it seems to be nearly perfect in well-posed cases. We review its construction in \S\ref{sec:mgc}.
The constructions of the graph affinities and GCW for type 3 and type 2 puzzles are described in \S\ref{sec:conn_graph_const_Type3} and \S\ref{sec:conn_graph_const_Type2}, respectively.

\subsection{Gallagher's MGC Metric}
\label{sec:mgc}

We review the construction of the MGC metric \cite{gallagher2012jigsaw}.
This metric between sides of patches quantifies the proximity of the gradients computed at each side. It outputs a symmetrized version of the Mahalanobis distance between the vector of gradients of one side and the estimated distribution of gradients at the other side, where the distribution is represented by its estimated covariance.

We assume two neighboring image patches $P_i$ and $P_j$ of size $s \times s$. There are four different relative positions of $P_i$ and $P_j$.
We assume without loss of generality the left-right relative position ($P_i$ on left and $P_j$ on right), and compute the corresponding MGC, which we denote by
$\MGC_{\lr} (P_i, P_j)$, as follows.
For each color channel $c$ (red, green and blue) and each row $r$, $1 \le r \le s$, of the $s \times s$ patch $P_i$, we find the derivatives near the right side of the image patch $P_i$ in the direction left-right as follows:
$$
G_{iL}(r, c) = P_i(r, s, c) - P_i(r, s-1, c).
$$
The subscript $L$ in the above equation indicates that patch $P_i$ is on the left side of the patch $P_j$.
To avoid some numerical problems, Gallagher\cite{gallagher2012jigsaw} suggests adding the following 9 ``dummy gradients'' $(0, 0, 0)$, $(1, 1, 1)$, $(-1, -1, -1)$, $(0, 0, 1)$, $(0, 1, 0)$, $(1, 0, 0)$, $(-1, 0, 0)$, $(0, -1, 0)$ and $(0, 0, -1)$ as additional rows of the matrix $\boldsymbol{G}_{iL}$. The extended matrix in $\RR^{(s + 9) \times 3}$ is denoted by $\tilde{\boldsymbol{G}}_{iL}$.

Next, for each color channel $c$ we define
\begin{equation*}
\vmu_{iL}(c)= \frac{1}{s} \sum \limits_{r=1}^s \mG_{iL}(r, c).
\end{equation*}
The regularized covariance matrix $\mSigma_{iL} \in \RR^{3 \times 3}$ between color channels is
\begin{equation*}
\mSigma_{iL} = \frac{1}{s+8} (\tilde{\mG}_{iL} - \text{mean}(\tilde{\mG}_{iL}))^T (\tilde{\mG}_{iL} - \text{mean}(\tilde{\mG}_{iL})),
\end{equation*}
where
\begin{equation*}
\text{mean}(\tilde{\mG}_{iL})= \frac{1}{s+9}\sum \limits_{r=1}^{s+9} \tilde{\mG}_{iL}(r, c) =  \frac{1}{s+9}\sum \limits_{r=1}^s \mG_{iL}(r, c).
\end{equation*}
We also define $\mG_{ij \LR}(r)$, the derivative from the left $s \times s$ image patch $P_i$ to the right $s \times s$ image patch $P_j$ at row $r$ and color $c$, by
\begin{equation*}
\mG_{ij \LR} (r, c) = P_j(r, 1, c) - P_i(r, s, c).
\end{equation*}
The left-to-right compatibility measure from $P_i$ to $P_j$ is defined by
\begin{equation*}
D_{\LR}(P_i, P_j) = \sum \limits_{r=1}^s (\mG_{ij \LR} (r) - \vmu_{iL}) \mSigma_{iL}^{-1} (\mG_{ij \LR}(r) - \vmu_{iL})^T.
\end{equation*}
Similarly, one can define the right-to-left compatibility measure from $P_j$ to $P_i$ in the same left-right setting, where $P_i$ is to the left of $P_j$.
The left-right MGC metric then has the symmetrized form
\begin{equation}
\label{eq:mgc_val_def}
\MGC_{\lr} (P_i, P_j) = D_{\LR}(P_i, P_j) + D_{\RL}(P_j, P_i).
\end{equation}
The right-left, top-bottom and bottom-top MGC's, denoted by $\MGC_{\rl} (P_i, P_j)$, $\MGC_{\tb} (P_i, P_j)$ and $\MGC_{\bt} (P_i, P_j)$, respectively, are similarly computed.

\subsection{Connection Graph Construction for Type 3 Puzzles}
\label{sec:conn_graph_const_Type3}

For type 3 puzzles, the locations of patches are given. Furthermore, edges are drawn between neighboring patches.
The affinity function is set by $\mW(i, j) = 1$ for all $\{i, j\} \in E$.
One need only find the unknown orientations, that is, the unknown connection matrix $\mR$.

%We remark that this is a much simpler problem type 2 puzzles, where the weighted graph is unknown,
%since the translations of the edges are unknown and thus one has no knowledge how to connect them.

To construct the connection function we propose to use the MGC metric, described in \S\ref{sec:mgc}.
For all neighboring patches $P_i$ and $P_j$, we calculate the possible $16$ values of the MGC metric (corresponding to the $16$ relative positions of $P_i$ and $P_j$) and
select the smallest of these numbers and its corresponding rotation $\mR[i, j]$.
% such that $P_i$ and the application of $\mR(i, j)$ to  $P_j$ match.
If there is no unique minimum among these $16$ values we suggest assigning $\mW(i, j) = 1/2$ (or another value smaller than 1) and letting $\mR[i, j]$ be the mean of the candidate rotations that obtain the minimal value.

\subsection{Connection Graph Construction for Type 2 Puzzles}
\label{sec:conn_graph_const_Type2}

We propose the following step-by-step procedure for constructing the affinity graph, the affinity function and the connection function for type 2 puzzles and then summarize this procedure in \Cref{algo:aff_gr}.
The rest of this section is organized as follows:
\S\ref{sec:init_step} discusses the initial step of constructing the connection graph;
\S\ref{sec:jaccard_step} discusses the Jaccard index and explains how to use it to update the affinity function;
lastly, \S\ref{sec:making_conn} describes how to deal with the cases when the connection graph is disconnected;
\S\ref{sec:diag_nb} describes how to find and use diagonal neighbors in order to construct a more reliable connection graph.

\subsubsection{Initial Step}
\label{sec:init_step}

We start with an initial construction of the directed graph $G = (V, E_{\est})$. The vertex set $V$ contains the patches in $\cQ$.
The edge set $E_{\est}$ is updated by the following procedure. In order to describe it, we denote by $\mR \cdot P$ the action of the rotation $\mR \in \ZZ_4$ on the patch $P$.
For a patch $P_i$, we find the patches $P_{i_{\ttop}}, P_{i_{\lleft}}, P_{i_{\bbottom}}, P_{i_{\rright}}$ and the corresponding rotations $\mR[i, i_{\ttop}], \mR[i, i_{\lleft}], \mR[i, i_{\rright}], \mR[i, i_{\bbottom}] \in \ZZ_4$ such that
\begin{equation}
\label{eq:split_4_P_R}
\begin{split}
\{ P_{i_{\ttop}}, \mR[i, i_{t}] \} \in \argmin_{P \in \cQ, \mR \in \ZZ_4} \MGC_{\bt}( P_i, \mR \cdot P), \\
\{ P_{i_{\lleft}}, \mR[i, i_{\lleft}] \} \in \argmin_{P \in \cQ, \mR \in \ZZ_4} \MGC_{\rl}( P_i, \mR \cdot P), \\
\{ P_{i_\bbottom}, \mR[i, i_{\bbottom}] \} \in \argmin_{P \in \cQ, \mR \in \ZZ_4} \MGC_{\tb}( P_i, \mR \cdot P), \\
\{ P_{i_{\rright}}, \mR[i, i_{\rright}] \} \in \argmin_{P \in \cQ, \mR \in \ZZ_4} \MGC_{\lr}( P_i, \mR \cdot P).
\end{split}
\end{equation}
The set $E_{\est}$ of directed edges contains the edges that connect each vertex with index $i$, $1 \leq i \leq n$, to the vertices with indices $i_{\ttop}$, $i_{\lleft}$, $i_\bbottom$ and $i_{\rright}$.
Note that these indices solving \eqref{eq:split_4_P_R} may not be unique and we consider all solutions of \eqref{eq:split_4_P_R} when forming $E_{\est}$.

\begin{figure}
\centering     %%% not \center
%\subfigure[Original Puzzle]{\label{fig:init_step_1}
\includegraphics[width=0.49\columnwidth]{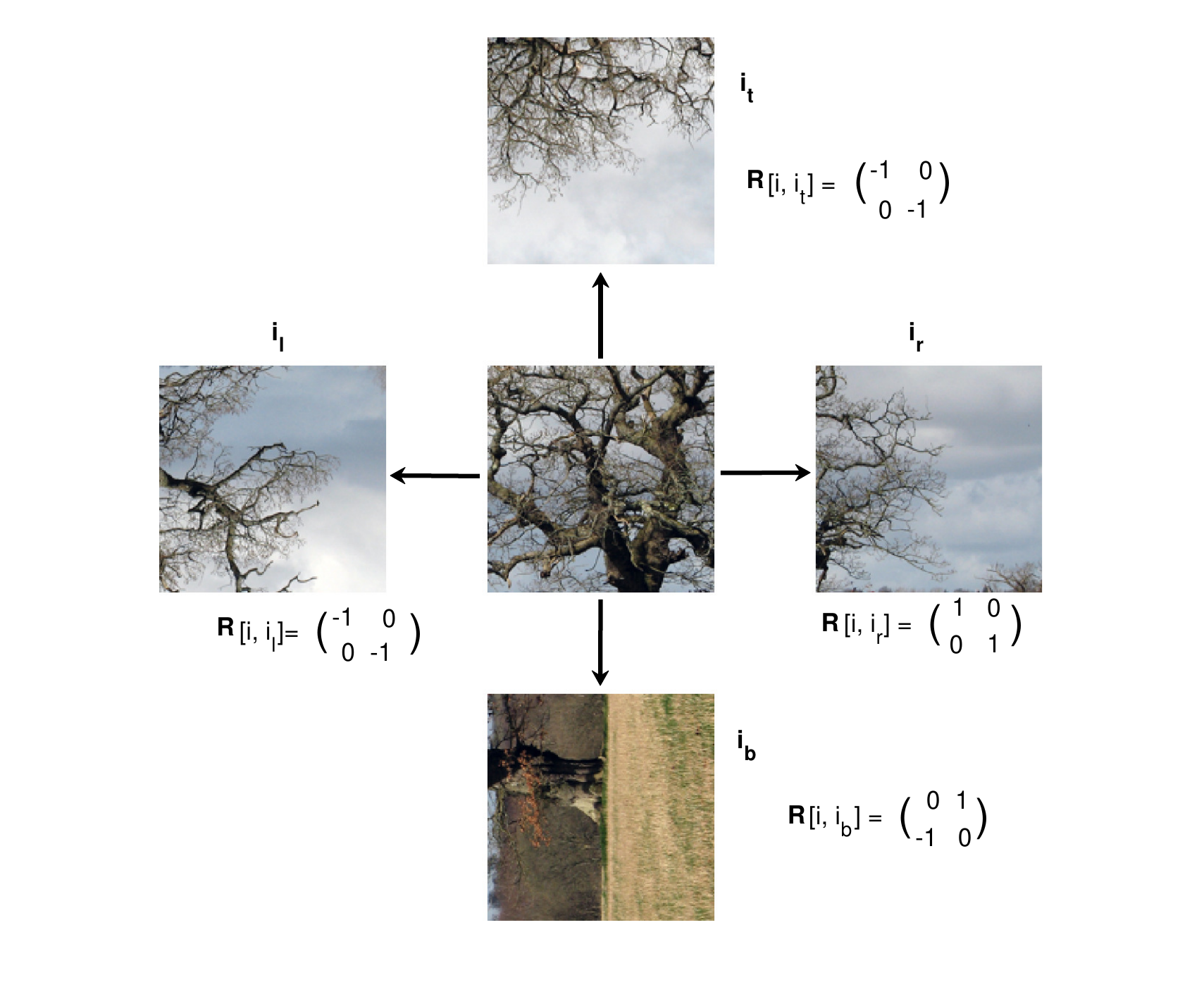}
%\subfigure[After Finding Rotations]{\label{fig:init_step_2}
%\includegraphics[width=0.32\columnwidth]{Figures/first_step_example_rot.png}
%\subfigure[Making the Graph Undirected]{\label{fig:init_step_3}
\includegraphics[width=0.49\columnwidth]{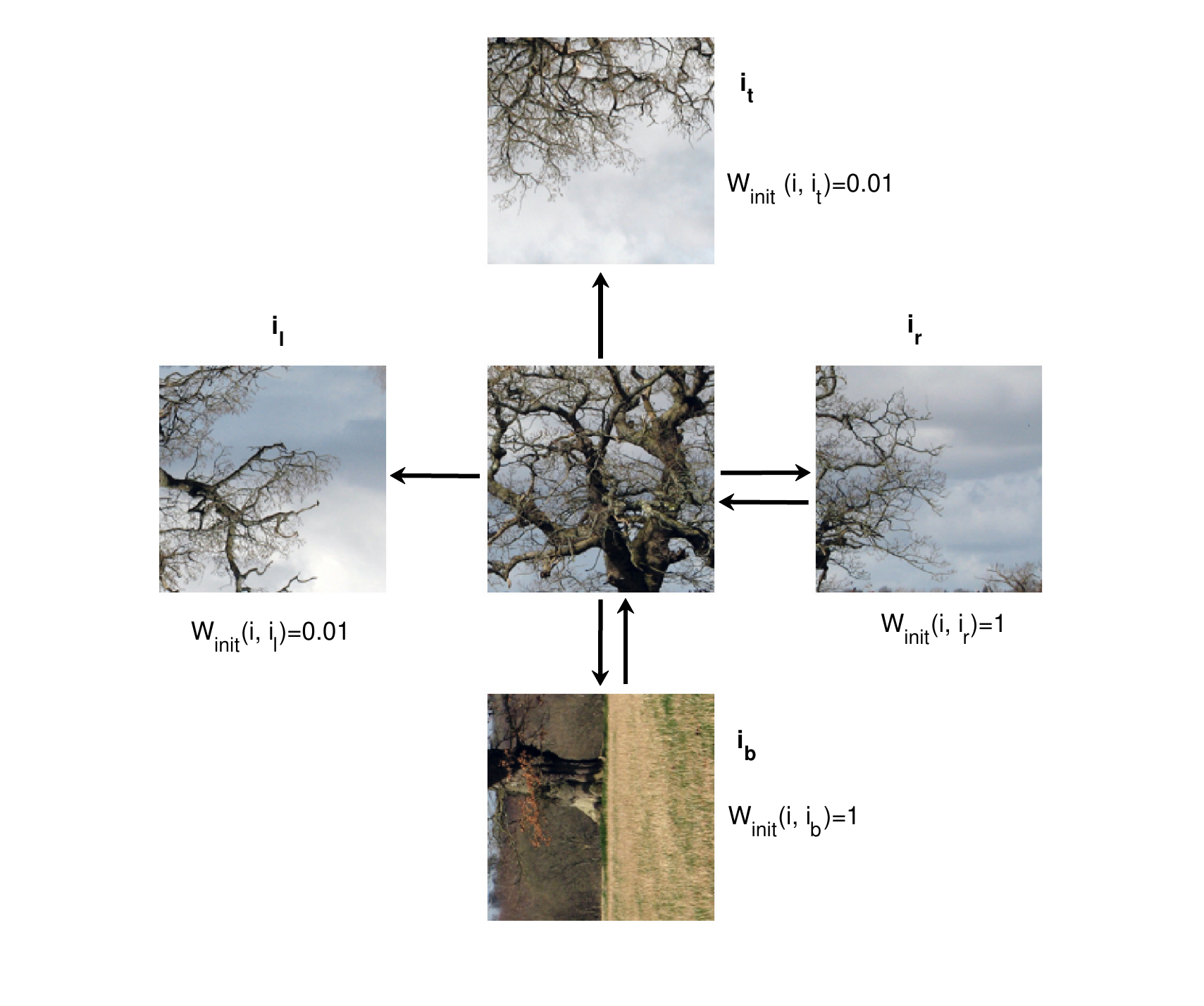}
\caption{Demonstration of the initial step for the construction of the connection graph. The left figure demonstrates the best matches for a given patch from the four directions: top, left, bottom and right. For each matching patch it records the rotation whose application to this patch results in correct matching with the central patch.
The  right figure shows the application of these rotations to the matching patches and demonstrates how to assign the weights to the undirected graph.
In this example, the matching patches from top and left were originally connected by a single direction; their weights in the undirected graph are thus $0.01$. On the other hand,
the patches from  right and bottom are connected in both directions and thus their weights in the undirected graph are $1$.}
\label{fig:init_step}
\end{figure}

We next modify the directed graph $G = (V, E_{\est})$ into an undirected graph. We use the default parameter 0.01 and form an initial affinity matrix $\mW_{\init} \in \RR^{n \times n}$  whose elements for $1 \leq i, j \leq n$ are
\begin{equation}
\label{eq:w_init_def}
\mW_{\init}(i, j) = \mW_{\init}(j, i) =
			\begin{cases}
                1, 	& \text{ if both } (i, j) \text{ and } (j, i) \in E_{\est}; \\
               0.01, & \text{ if only one of } (i, j) \text{ or } (j, i) \text{ is in } E_{\est}; \\
               0, 		& \text{ otherwise.}
            \end{cases}
\end{equation}
%$\mW_{\init}(i, j) = \mW_{\init}(j, i) = w_1$ if both $(i, j)$ and $(j, i) \in E_{\est}$
%and $\mW_{\init}(i, j) = \mW_{\init}(j, i) = w_2$ if only one of $(i, j)$ or $(j, i)$ is in $E_{\est}$.
\Cref{fig:init_step} demonstrates the above construction for a particular patch.

Next, we enforce the constraint that, for square jigsaw puzzles, each patch can have at most one neighbor for each direction by trimming some edges
that are likely not neighbors. This is done as follows. Assume without loss of generality that patch $P_i$ has more than one neighbor in the top direction and denote these neighbors by
$P_{i_1}, \dots, P_{i_k}$ where $k>1$. Then we solve the minimization problem
\begin{equation}
\label{eq:best_match}
j \in \argmin_{1 \le j \le k} \MGC_{\bt}(P_i, \mR[i, i_j] \cdot P_{i_j}).
\end{equation}
We remark that since \eqref{eq:w_init_def} makes the graph undirected, $P_{i_1}, \dots, P_{i_k}$ might contain more patches than the ones obtained from \eqref{eq:split_4_P_R}.
Therefore, the optimization problems \eqref{eq:split_4_P_R} and \eqref{eq:best_match} can be different. In particular, a unique solution of \eqref{eq:split_4_P_R} might become non-unique for \eqref{eq:best_match}.
If \eqref{eq:best_match} has a unique solution, we keep the edge $\{i, i_j\}$ and remove the rest of the edges. Otherwise, we remove all edges $\{i, i_j\}_{j=1}^k$ from $E.$
The procedure is analogous if $P_i$ has more than one neighbor from left, bottom or right. If edges were eliminated from $E_{\est}$,
then the matrix $\mW_{\init}$ is updated so it is zero on the corresponding indices. This process results in the following initial connection graph $G = (V, E_{\est}, \mW_{\init}, \mR).$

%Due to the special topology of the true affinity graph for two-dimensional square jigsaw puzzles, which is, each vertex has at most $4$ neighbors;
%one for each direction (top, bottom, left and right), we suggest to use the nearest-neighbors algorithm to start the construction of the affinity graph.
The construction of this graph uses a nearest-neighbor construction. For a high-noise regime, El Karoui and Wu~\cite{el2016graph} recommend avoiding a nearest-neighbor construction.
However, due to the special lattice structure of the true graph, the nearest-neighbor initial construction is natural for the square jigsaw puzzle problem.

\subsubsection{Use of Jaccard Index to refine the graph}
\label{sec:jaccard_step}

Next, we refine the connection graph by trying to assess the validity of the edges and decrease the weights of edges that do not seem valid; that is, they may not appear in the true connection graph.
The idea is to check after removing an edge whether its neighbors are still connected in some weak sense to each other. If so, then the edge seems to be valid, and otherwise, it may not be valid.
For this purpose, we use the Jaccard index \cite{jaccard1901etude}.

\begin{figure}
\centering     %%% not \center
%\subfigure{\label{fig:first_second}
\includegraphics[width=0.4\columnwidth]{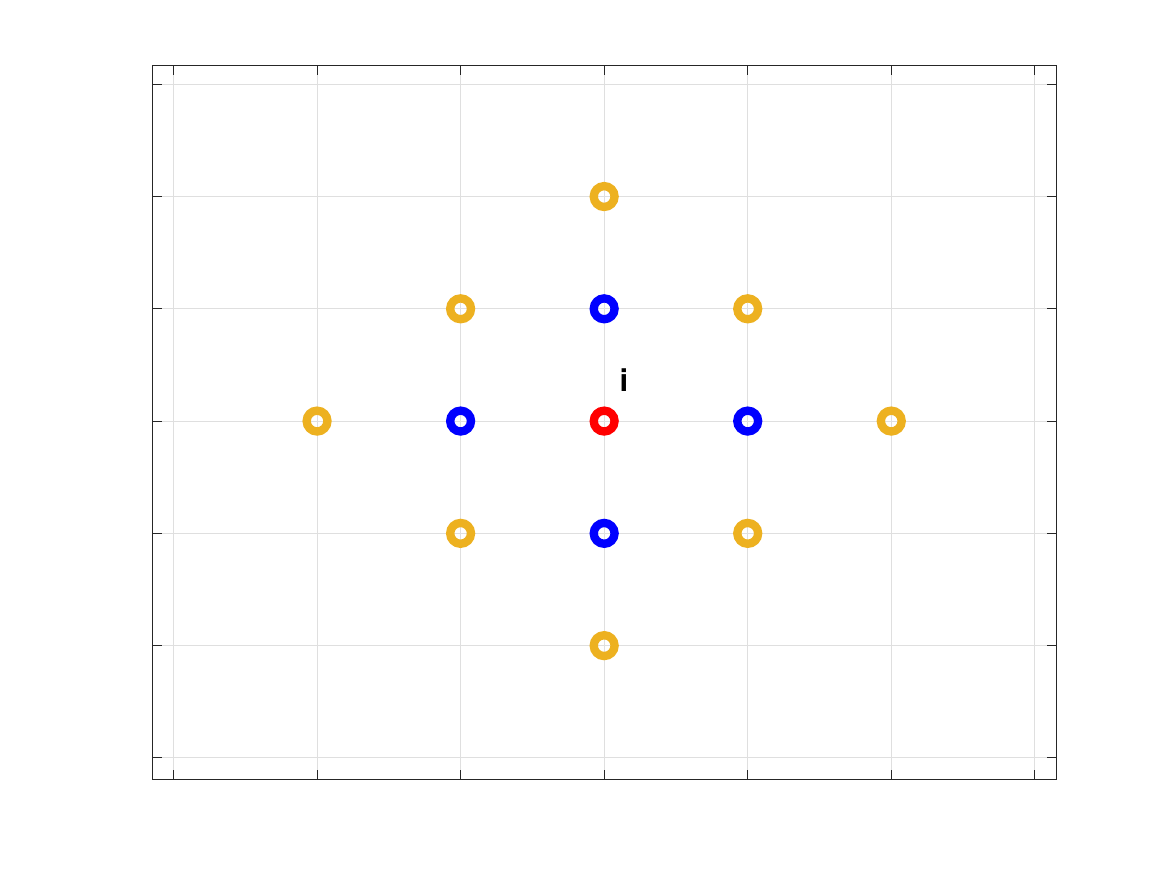}
\caption{Demonstration of the sets $N_{G, i}^1$ and $ N_{G, i}^2$. A given vertex $i$ is colored in red,
the elements of the set $N_{G, i}^1$ are colored in blue and the elements of the set $N_{G, i}^2$ are colored in blue and orange.}
\label{fig:nb_description}
\end{figure}

The description of this index uses the following notation in a graph $G = (V, E)$. Given a vertex $i$, $1 \le i \le |V|$,
let $N_{G, i}^1$ denote the set of  vertices in $V$ which are connected to vertex $i$, that is, $N_{G, i}^1 = \{j \in V | \{i, j\} \in E\}$.
Using our terminology, $N_{G, i}^1$ contains the neighbors of $i$.
The set $N_{G, i}^2$
contains all vertices that are at most $2$ steps away from vertex $i$, except vertex $i$.
That is, $N_{G, i}^2 = \bigcup_{j \in N_{G, i}^1} N_{G, j}^1 \setminus \{i\}$.
Finally, let $G^{\setminus (i, j)} = (V, E \setminus \{(i, j)\} )$ denote the graph with the edge $(i,j)$ removed. The sets $N_{G, i}^1 $ and $N_{G, i}^2$ are demonstrated in \Cref{fig:nb_description}.

\begin{figure}
\centering     %%% not \center
%\subfigure{\label{fig:first_second}
\includegraphics[width=0.4\columnwidth]{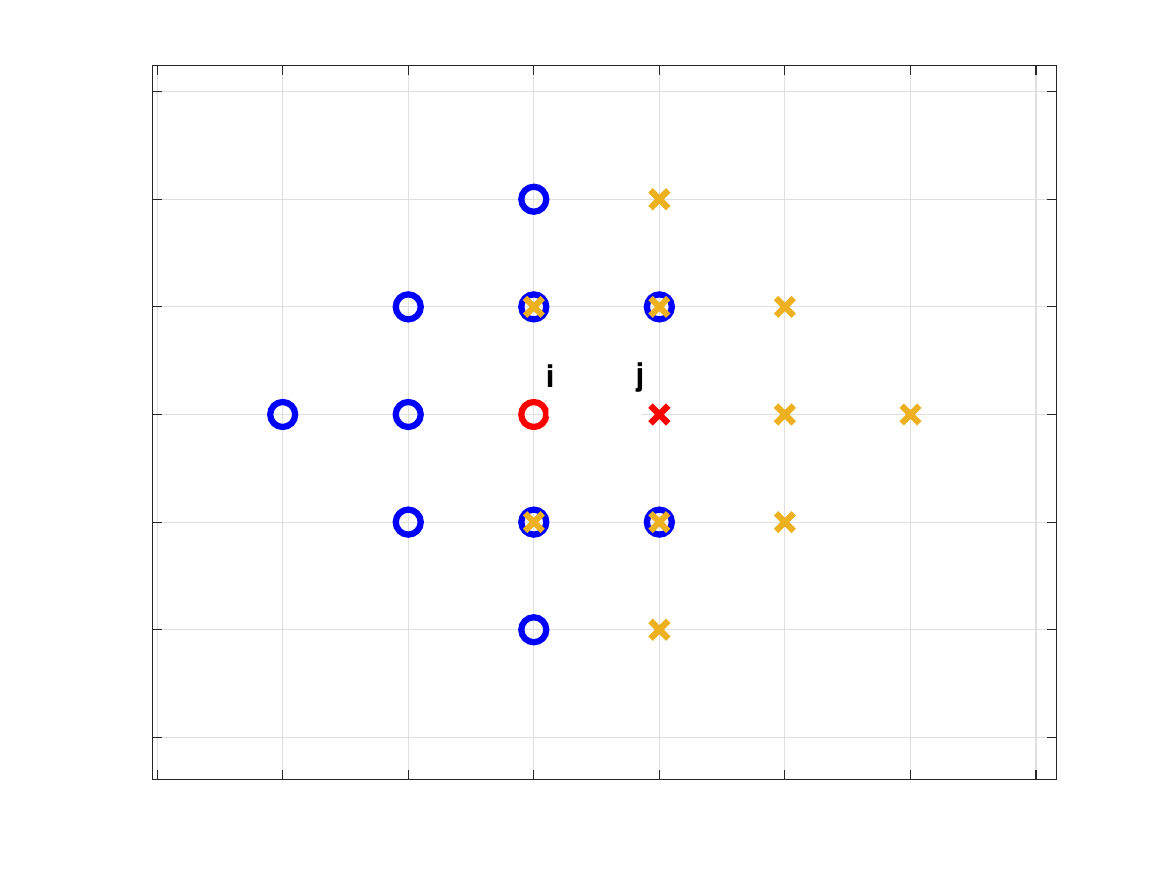}
\caption{Demonstration of Jaccard index. Vertex $i$ is denoted by a red circle and vertex $j$ is denoted by a red cross. The edge between these vertices was removed from the grid.
The elements of $N_{G^{\setminus (i, j)}, i}^2$ are denoted by blue circles and the elements of $N_{G^{\setminus (i, j)}, j}^2$
by orange crosses. The Jaccard index is four since there are four elements in $N_{G^{\setminus (i, j)}, i}^2 \cap N_{G^{\setminus (i, j)}, j}^2$ (denoted by blue circles filled with orange crosses).}
\label{fig:nb_jac}
\end{figure}

By using this notation, we define the Jaccard index between vertices $i$ and $j$ as
\begin{equation}
\label{eq:jac_eq}
\mu_{\Jaccard}(i, j) = |N_{G^{\setminus (i, j)}, i}^2 \cap N_{G^{\setminus (i, j)}, j}^2|\,,
\end{equation}
where $|\cdot|$ denotes the cardinality of a set.
This definition is similar to the one in \cite{jaccard1901etude}, but there are two differences. The first one is that we consider the graph $G^{\setminus (i, j)}$ instead of $G$ to emphasize the common neighbors, while excluding the obvious pair $(i,j)$. The second one is that we do not divide by
$|N_{G^{\setminus (i, j)}, i}^2 \cup N_{G^{\setminus (i, j)}, j}^2|$. The latter division does not matter to us as we only care about the positivity of this index.
\Cref{fig:nb_jac} demonstrates calculation of the Jaccard index for a special example.
Note that the chance of two vertices $i$ and $j$ to be neighbors in the graph $(V, E_{\est})$ is higher if $\mu_{\Jaccard} (i, j) > 0$ than if $\mu_{\Jaccard} (i, j) = 0$.
Thus, we propose to use the Jaccard indices to refine the connection graph. We use another weight matrix $\mW_{\Jaccard} \in \RR^{n \times n}$, defined as
\begin{align}
\label{eq:w_jaccard}
\mW_{\Jaccard} (i, j) = \mW_{\Jaccard} (j, i) =\left\{
\begin{array}{ll}
 0,		& \{i, j\} \in E_{\est}\mbox{ and }\mu_{\Jaccard}(i, j) = 0;\\
 \mW_{\init}(i, j), & \mbox{otherwise.}
\end{array}
\right.\,
\end{align}
%If $\{i, j\} \in E_{\est}$ and $\mu_{\Jaccard}(i, j) = 0$, then $\mW_{\Jaccard} (i, j) = \mW_{\Jaccard} (j, i) = 0$. If on the other hand $\mu_{\Jaccard}(i, j) > 0$, then $\mW_{\Jaccard} (i, j) = \mW_{\Jaccard} (j, i) = \mW_{\init} (i, j)$.

Since this procedure might also remove many correct edges by zeroing out the corresponding values of the affinity function,
we propose a linear combination of $\mW_{\init}$ and $\mW_{\Jaccard}$ with larger coefficient given to $\mW_{\Jaccard}$.
In our experiments we use the default parameter $0.2$ (and $0.8=1-0.2$) to set
\begin{equation}
\label{eq:w_nb}
\mW_{\nb} = 0.2 \times \mW_{\init} + 0.8 \times \mW_{\Jaccard}
\end{equation}
and form the affinity graph $G = (V, E, \mW_{\nb}, R)$.
\subsubsection{Making the Affinity Graph Connected}
\label{sec:making_conn}

\begin{figure}
\centering     %%% not \center
%\subfigure{\label{fig:grid_example_disconected}
\includegraphics[width=0.49\columnwidth]{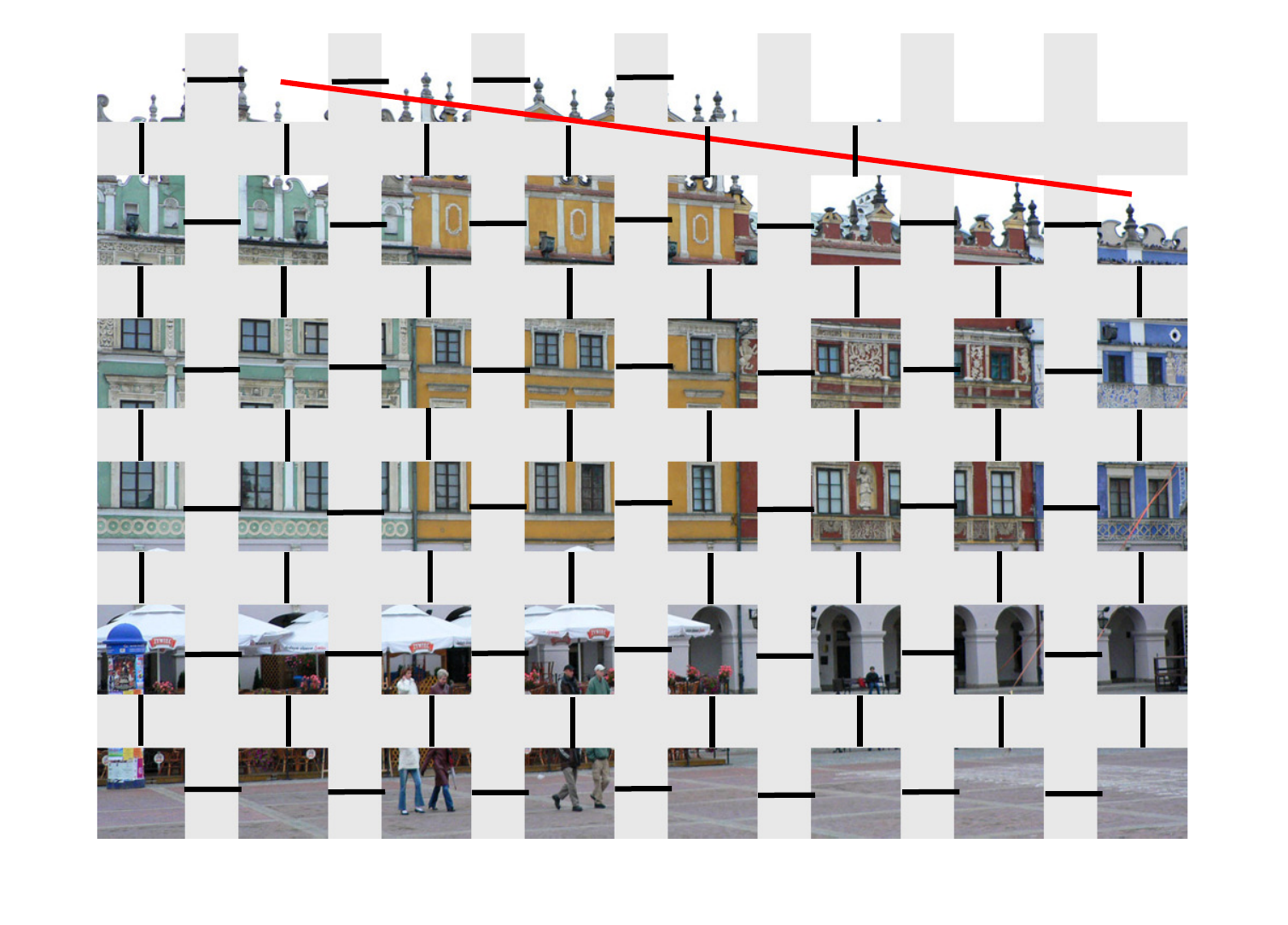}
%\subfigure{
\includegraphics[width=0.49\columnwidth]{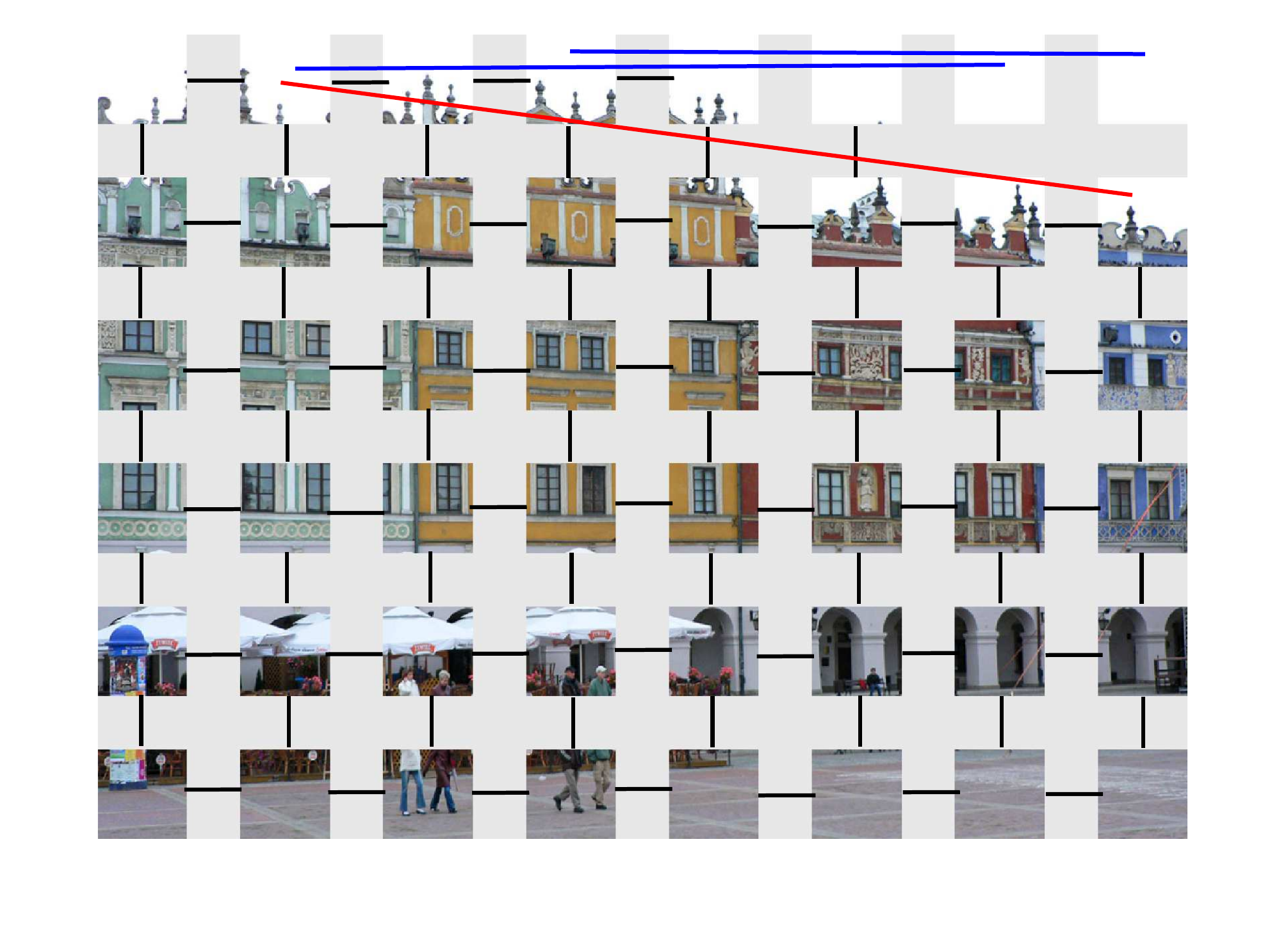}
\caption{Demonstration of a disconnected affinity graph and the way it got connected. The left figure shows an example where the resulting affinity graph using our method is disconnected.
Indeed, the two top right patches are not connected to any of the other patches. The black edges connect between true neighbors and the only red edge is a wrongly determined edge.
The right figure demonstrates the result of the simple procedure described in \S\ref{sec:making_conn}. The connected graph has two new blue edges. While these blue edges connect between non-neighboring patches,
the originally disconnected patches are uniformly white and thus their rotations do not matter for the reconstruction of the image.}
\label{fig:grid_example_disc_conn}
\end{figure}

The procedures described in \S\ref{sec:init_step} and \S\ref{sec:jaccard_step} might result in a disconnected affinity graph $G$
as demonstrated in the left image of \Cref{fig:grid_example_disc_conn}.
To complete $G$ so it is connected, we first find all connected components of $G$. Assume that there are $k$ connected components with corresponding vertices $V_1$, $\ldots$, $V_k$ that partition the set of vertices $V$.
Assume further that they are labeled by descending size order, i.e., $|V_1| \ge |V_2| \dots \ge |V_k|.$ Next, we find vertices $i \in V_1$ and $j \in V \setminus V_1$ that minimize the MGC metric
between the patches $P_i$ and $P_j$. Mathematically, we find
\begin{equation}
\label{eq:find_conn}
\begin{split}
\{i, j, \mO \} \in \argmin \limits_{i \in V \setminus V_1, j \in V_1, \mO \in \ZZ_4} \min\{ \MGC_{\lr}(P_i, \mO \cdot P_j)), \MGC_{\tb}(P_i, \mO \cdot P_j)), \\ \MGC_{\rl}(P_i, \mO \cdot P_j)), \MGC_{\bt}(P_i, \mO \cdot P_j)) \}.
\end{split}
\end{equation}
If the solution of \eqref{eq:find_conn}  is not unique, we randomly choose one solution.
We then add the edge $\{i, j\}$ of the chosen solution to $E_{\est}$ and update the weight as follows:
\begin{equation}
\label{eq:w_nb_005}
\mW_{\nb}(i, j) = \mW_{\nb}(j, i) = 0.005, \ \  \mR[i, j] = \mO \ \text{ and } \ \mR[j, i] = \mO^T.
\end{equation}
We remark that 0.005 is a free parameter chosen to be half the size of the parameter 0.01 in \eqref{eq:w_init_def}.
We iterate the procedure described above for $V_2,\ldots, V_k$ until the graph becomes connected.
The number of iterations needed is $k-1$ since there are $k$ connected components and at each iteration we connect the largest component with a remaining component.

\subsubsection{Taking Advantage of 4-Loops}
\label{sec:diag_nb}

\begin{figure}
\centering     %%% not \center
\includegraphics[width=0.4\columnwidth]{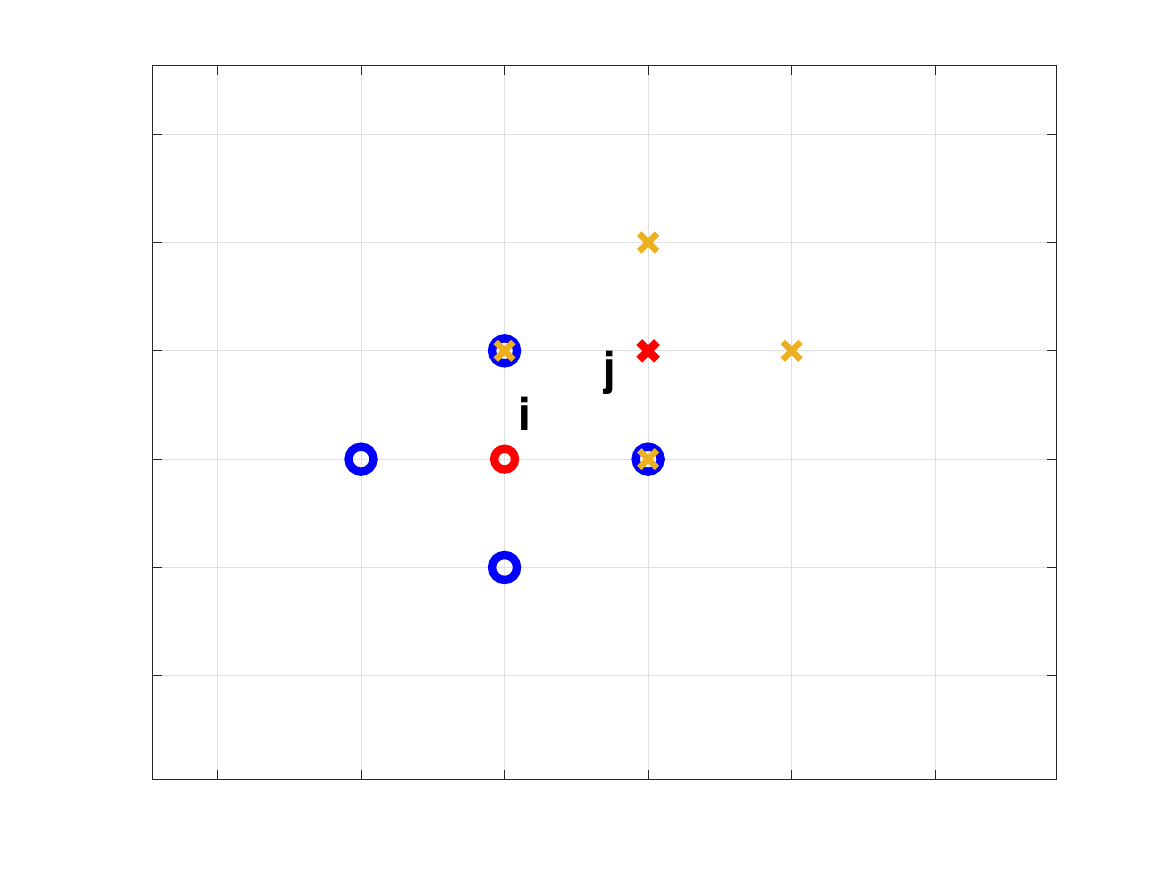}
\caption{Demonstration of finding diagonally neighboring vertices in a uniform grid. Two vertices $i$ and $j$ are denoted by a red circle and a red cross, respectively. The elements of the sets $N_{G, i}^1$ and $N_{G, j}^1$ are
colored by blue and orange, respectively. The intersection of these sets yields the two diagonally neighboring vertices to $i$ and $j$. Together with $i$ and $j$ they form a 4-loop.}
\label{fig:diag_nb_1}
\end{figure}

We refine the constructed connection graph by using the following property of
the square jigsaw puzzle:
If two patches $P_i$ and $P_j$ are diagonal neighbors, then there exist exactly two other patches $P_{n_1}$ and $P_{n_2}$ and a cycle containing the vertices $i$, $n_1$, $j$ and $n_2$. This idea is demonstrated in \Cref{fig:diag_nb_1}. Such a cycle of 4 vertices is referred to as a 4-loop by
\cite{son2014solving}. In this latter work, $4$-loops were used to solve the puzzle problem. We use them to define a better connection graph.
As we have already discussed, for a uniform grid, each patch can have at most $4$ direct neighbors (right, top, left or bottom). Furthermore, each patch has at most $4$ diagonal neighbors.
Exactly four diagonal neighbors are obtained for a patch in the interior of the puzzle, a single diagonal neighbor occurs for a corner patch and there are 2 diagonal neighbors for a patch that lies on the boundary of the uniform grid but not on a corner.

\begin{figure}
\centering     %%% not \center
\includegraphics[width=0.6\columnwidth]{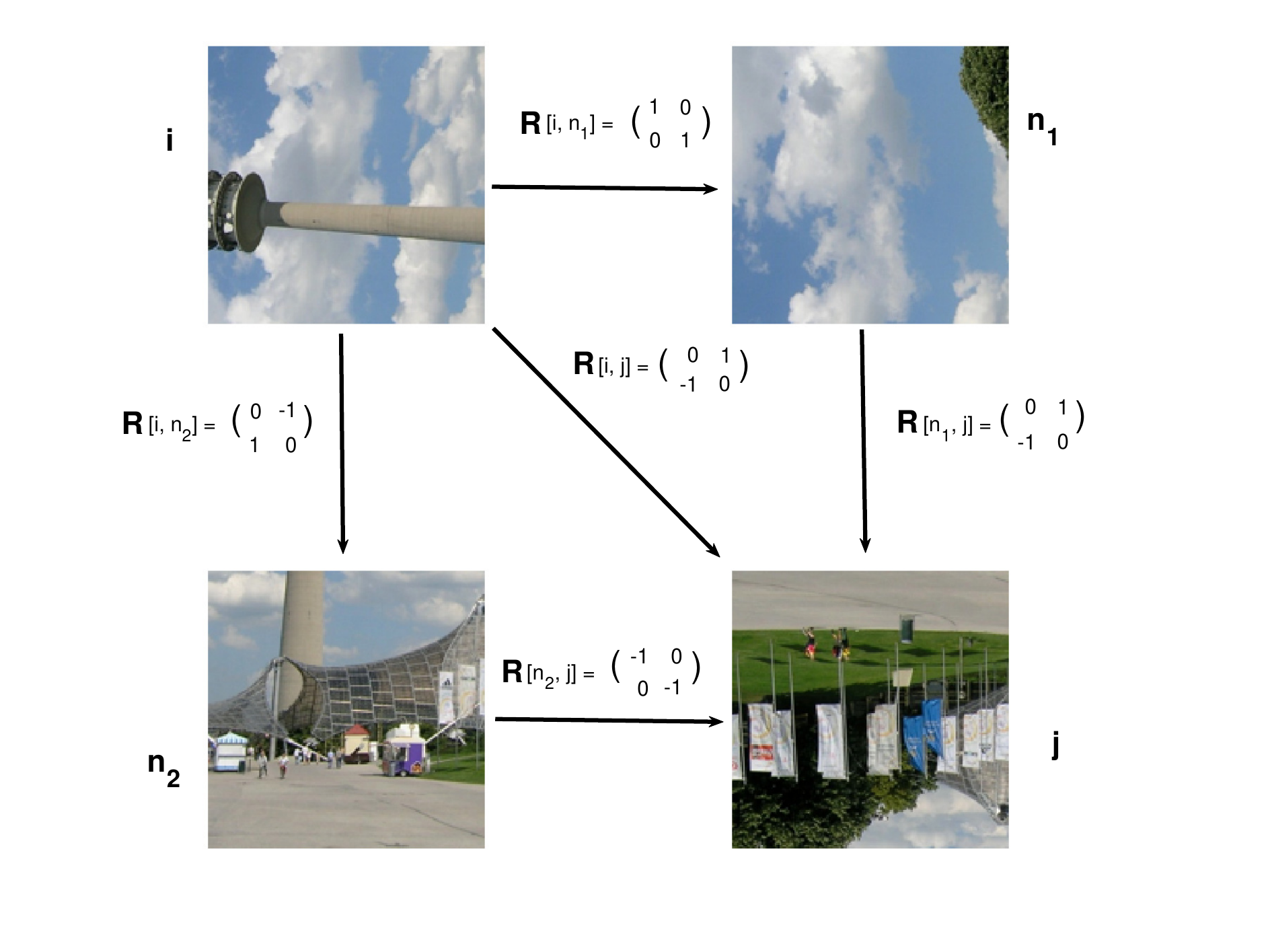}
\caption{
Intuition for the requirement in \eqref{eq:diag_nb_update} and \eqref{eq:diag_nb_dir_nb_update}. The two vertices $i$ and $j$ are diagonal neighbors
and the vertices $n_1$ and $n_2$ satisfy \eqref{eq:NG1}. Thus, $i$, $j$, $n_1$ and $n_2$ form a cycle of size 4, that is, a 4-loop. The relative rotations between vertices are indicated on the corresponding edges.
We note that both $\mR[i, n_1] \mR[n_1, j]$ and $\mR[i, n_2] \mR[n_2, j]$ are equal to the relative rotation $\mR[i,j]$ shown on edge $(i,j)$. In particular,  $\mR[i, n_1] \mR[n_1, j] = \mR[i, n_2] \mR[n_2, j]$.
The assigned weights thus try to encourage this constraint and penalize cases where it is not satisfied.
\label{fig:diag_nb_2}}
\end{figure}

For patches $P_i$ and $P_j$ we define
\begin{equation}
\label{eq:diag_deg}
\delta_{\diag}(i, j) = |N_{G, i}^1 \cap N_{G, j}^1|.
\end{equation}
We observe that patches $P_i$ and $P_j$ are diagonal neighbors in the uniform grid if and only if $\delta_{\diag}(i, j) = 2.$
To find the diagonal neighbors for graph $G = (V, E)$ we propose a two step procedure.
First, we find the set of all pairs of vertices $\{i, j\} \in V \times V$ for which $\delta_{\diag}(i, j) = 2.$
For each such pair $\{i, j\}$ there exists another pair $\{n_1,n_2\}$ such that
\begin{equation}
\label{eq:NG1}
N_{G, i}^1 \cap N_{G, j}^1 = \{n_1, n_2\},
\end{equation}
or equivalently, $i$, $n_1$, $j$ and $n_2$ are contained in a 4-loop.
%If rotations in this 4-loop match in the following way: $\mR(i, n_1) \mR(n_1, j) = \mR(i, n_2) \mR(n_2, j)$, then we set $\mW_{\diag}(i, j) = \mW_{\diag}(j, i) = 1$, $\mR(i, j) = \mR(n_1, j) \mR(i, n_1)$ and $\mR(j, i) = \mR(i, j)^T$.
%Otherwise we reduce the weights between the pairs $\{i,n_1\}$, $\{i,n_2\}$, $\{j,n_1\}$, $\{j,n_2\}$ by a factor of 2.
We set
\begin{align}
\label{eq:diag_nb_update}
\mW_{\diag}(i, j) = \left\{
\begin{array}{ll}
1, & \mbox{when } \delta_{\diag}(i, j) = 2 \text{ and } \mR[i, n_1] \mR[n_1, j] = \mR[i, n_2] \mR[n_2, j];\\
0,&\mbox{otherwise.}
\end{array}\right.
\end{align}
If $\mW_{\diag}(i, j) =1$, we add a diagonal edge between vertices $i$ and $j$ and assign the following value to the connection function:
\begin{equation}
\label{eq:diag_rot_update}
\mR[i, j] = \mR[i, n_1] \mR[n_1, j].
\end{equation}
We remark that the condition $ \mR[i, n_1] \mR[n_1, j] = \mR[i, n_2] \mR[n_2, j]$ in \eqref{eq:diag_nb_update} is naturally satisfied in the ground-truth graph as demonstrated in \Cref{fig:diag_nb_2}.
Therefore, when it is satisfied and also $\delta_{\diag}(i, j) = 2$, the maximal weight of 1 is assigned to the corresponding diagonal edge.
We note that $\delta_{\diag}(i, j) = 2$ if and only if $\delta_{\diag}(n_1, n_2) = 2$,  $\mR[i, n_1] \mR[n_1, j] = \mR[i, n_2] \mR[n_2, j]$ if and only if
 $\mR[n_1,i] \mR[i, n_2] = \mR[n_1,j] \mR[j, n_2]$ and thus
$\mW_{\diag}(i, j) =1$ if only if $\mW_{\diag}(n_1, n_2) =1$.

We further update blocks of the matrix $\mW_{\nb}$ so that their weights align better with the diagonal weights.  We denote by $\mW_{\nb}([i, j], [n_1 ,n_2])$ the $2 \times 2$ submatrix of $\mW_{\nb}$ indexed by $(i,n_1)$, $(i,n_2)$, $(j,n_1)$ and $(j,n_2)$. We fix the default parameters $1/3$ and $2/3$, and describe the assignment of $\mW_{\nb}([i, j], [n_1 ,n_2])$ in the following formula (though algorithmically one does not need to implement it per each block):
\begin{align}
\label{eq:diag_nb_dir_nb_update}
\mW_{\nb}([i, j], [n_1, n_2]) = \left\{
\begin{array}{ll}
\frac{\mW_{\nb}([i, j], [n_1, n_2])}{3}, &
\text{ if } \delta_{\diag}(i, j) = 2 \text{ and } \\
~ & \mR[i, n_1] \mR[n_1, j] \neq \mR[i, n_2] \mR[n_2, j]; \\
~ & ~ \\
\vone_{2\times2}, &
\text{ if } \delta_{\diag}(i, j) = 2 \text{ and } \\
~ & \mR[i, n_1] \mR[n_1, j] = \mR[i, n_2] \mR[n_2, j]; \\
~ & ~ \\
\frac{2 \mW_{\nb}([i, j], [n_1, n_2])}{3}, & \mbox{otherwise.}
\end{array}\right.
\end{align}
We note that if $\{i,j\}$ and $\{n_1,n_2\}$ are diagonal edges satisfying the rotation condition, then we enforce weight 1 for the nearest neighboring edges in the corresponding 4-loop: $\{i,n_1\}$, $\{i,n_2\}$, $\{j,n_1\}$ and $\{j,n_2\}$.
If they are not diagonal and don't satisfy the rotation condition, then we decrease the existing weights of these neighboring edges by a factor of 2/3. If the diagonal condition, $\delta_{\diag}(i, j) = 2$, holds, but neither does the rotation condition, which is somewhat contradicting, we reduce the weights of the former neighboring edges by a factor of 1/3.

We note that the support sets of $\mW_{\nb}$ and $\mW_{\diag}$ are disjoint. We set
\begin{equation}
\mW = \mW_{\nb} + \mW_{\diag},
\end{equation}
and this is the final step of constructing
the connection graph $G = (V, E_{\est}, \mW, \mR)$ for square jigsaw puzzles. The full algorithm of this construction is summarized in \Cref{algo:aff_gr}.

\begin{algorithm}
\caption{Connection Graph Construction for Type 2 Puzzles}
\label{algo:aff_gr}
\begin{algorithmic}
\State{\textbf{Input:} Puzzle Patches: $\{P_i\}_{i=1}^n  \subset \RR^{s \times s \times 3}$}
\begin{itemize}
\State For all $1 \le i < j \le n$ calculate the $16$ $\MGC$ metric values between patches $P_i$ and $P_j$ as explained in \S\ref{sec:mgc}
\State Construct $G = (V, E_{\est}, \mW_{\init}, \mR)$ according to the procedure described in \S\ref{sec:init_step} with the following three stages:
nearest-neighbors construction based on \eqref{eq:split_4_P_R}, symmetrization of $\mW_{\init}$ and pruning extra neighbors with the use of \eqref{eq:best_match}
\State For all $\{i,j\} \in E_{\est}$, calculate $\mu_{\Jaccard}(i,j)$ according to \eqref{eq:jac_eq}
\State For all $\{i,j\} \in E_{\est}$, if $\mu_{\Jaccard}(i,j) = 0$, set $\mW_{\Jaccard}(i,j) = 0$; otherwise, $\mW_{\Jaccard}(i,j) = 1$
\State Set $\mW_{\nb} = 0.8 \times \mW_{\Jaccard}  + 0.2 \times \mW_{\init}$
\State If the graph $G$ is disconnected, iteratively connect the largest connected component to smaller connected components as explained in \S\ref{sec:making_conn}
\State For all $i$, $j \in V$, calculate $\delta_{\diag}(i,j)$ according to \eqref{eq:diag_deg} and if $\delta_{\diag}(i,j)=2$, calculate $n_1$ and $n_2$ according to
\eqref{eq:NG1}
\State Form $\mW_{\diag}$ according to \eqref{eq:diag_nb_update}
and update $\mR$ and $\mW_{\nb}$ according to \eqref{eq:diag_rot_update} and \eqref{eq:diag_nb_dir_nb_update}
\State Set $\mW = \mW_{\nb} + \mW_{\diag}$
\end{itemize}
\State{\textbf{Return:} $G = (V, E_{\est}, \mW, \mR)$, MGC values for all pairs of patches}
\end{algorithmic}
\end{algorithm}

\section{Solution for Type 2 Puzzles via GCL and Location Solver}
\label{sec:loc_est}

This section completes the solution of type 2 jigsaw puzzles.
It assumes the formation of the connection graph according to \Cref{algo:aff_gr}, solution of the correct orientations by \Cref{algo:rot_sol} and then application of an existing location solver.
The new component is a procedure for updating the affinity function and the connection function based on the estimated rotations and locations. One can then estimate again
the orientations and locations and repeat this procedure several times.
This procedure and the complete solution of type 2 puzzles that uses this procedure are summarized below in \S\ref{sec:w_upd}.
At last, \S\ref{sec:time_complex} summarizes its time complexity.

\subsection{Updating the Affinity and Connection Functions and the Resulting Solution}
\label{sec:w_upd}

By now there are many successful solutions to type 1 puzzles.
According to our numerical tests, the stand-alone algorithms for solving type 1 puzzles of both Gallagher~\cite{gallagher2012jigsaw} and Yu et al.~\cite{yu2015solving} are highly competitive.
We have often noticed a slight advantage of the latter algorithm, which applies a linear programming procedure.
Therefore, we use the algorithm of Yu et al.~\cite{yu2015solving} (under the ``fixed-rotation mode'') as a default solver for type 1 puzzles in our algorithm.
One could use instead any algorithm that solves type 1 puzzles.

The basic idea for updating the values of the affinity and connection functions is that a given estimated solution for the orientations and locations can be used to infer possible mismatches. These identified mismatches could be used to reassign values for the affinity and connection functions that may lead to a more accurate solution.

First, we figure out which patches are wrongly placed in the assembled puzzle and remove them from the grid.
%(see demonstration in the first row of \Cref{fig:update_res}, where the boundaries of such patches are emphasized with dashed blue lines).
For this purpose we use the following kinds of metrics, which we refer to as $\NAM$ (Neighbor-Averaged Metric). For each patch $i$, $1 \leq i \leq n$, with neighbors $i_{\ttop}$, $i_{\lleft}$, $i_\bbottom$ and $i_{\rright}$,
from top, left, bottom and right, respectively, we utilize the MGC metric (see \eqref{eq:mgc_val_def} and text below it) to
define
\begin{equation}
\label{eq:nb_met}
\NAM_{\all}(i) = (\MGC_{\bt}(i, i_{\ttop}) + \MGC_{\rl}(i, i_{\lleft}) + \MGC_{\tb}(i, i_{\bbottom}) + \MGC_{\lr}(i, i_{\rright})) / 4.
\end{equation}
If a patch $i$ is at the edge or corner of the puzzle grid, then it has 3 or 2 neighbors, respectively. In this case, we only sum up the respective MGC values and divide the sum by the number of neighbors.
Similarly we define the following four metrics:
\begin{equation}
\label{eq:nb_met_3}
\begin{split}
& \NAM_{\ltr}(i) = (\MGC_{\rl}(i, i_{\lleft}) + \MGC_{\bt}(i, i_{\ttop}) + \MGC_{\lr}(i, i_{\rright}))/3, \\
& \NAM_{\trb}(i) = (\MGC_{\bt}(i, i_{\ttop}) + \MGC_{\lr}(i, i_{\rright}) + \MGC_{\tb}(i, i_\bbottom))/3, \\
& \NAM_{\blt}(i) = (\MGC_{\tb}(i, i_\bbottom) + \MGC_{\rl}(i, i_{\lleft}) + \MGC_{\bt}(i, i_{\ttop}))/3, \\
& \NAM_{\lbr}(i) = (\MGC_{\rl}(i, i_{\lleft}) + \MGC_{\tb}(i, i_{\bbottom}) + \MGC_{\lr}(i, i_{\rright}))/3.
\end{split}
\end{equation}
Again, if the patch is at the edge or corner of the puzzle grid, then we sum the appropriate MGC values and divide by the corresponding number of neighbors.

\begin{figure}
\centering
\includegraphics[width=0.40\columnwidth]{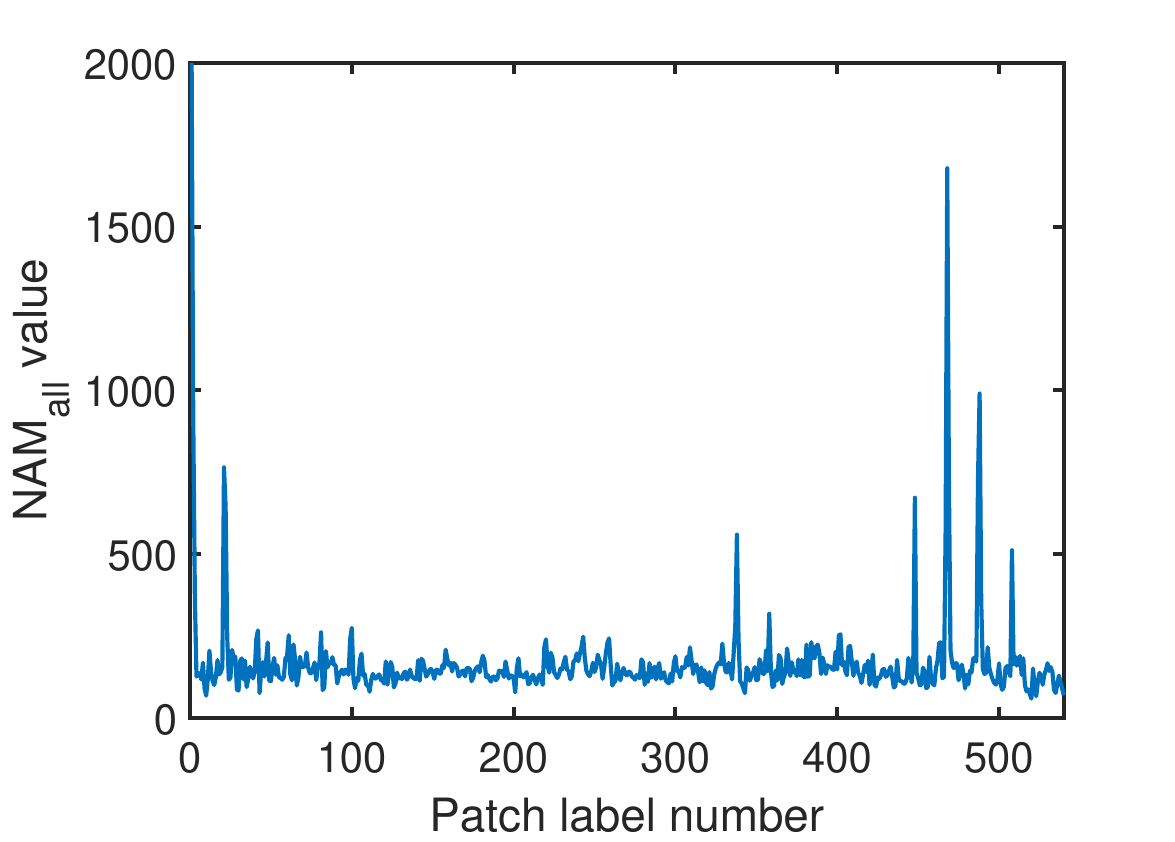}
\includegraphics[width=0.40\columnwidth]{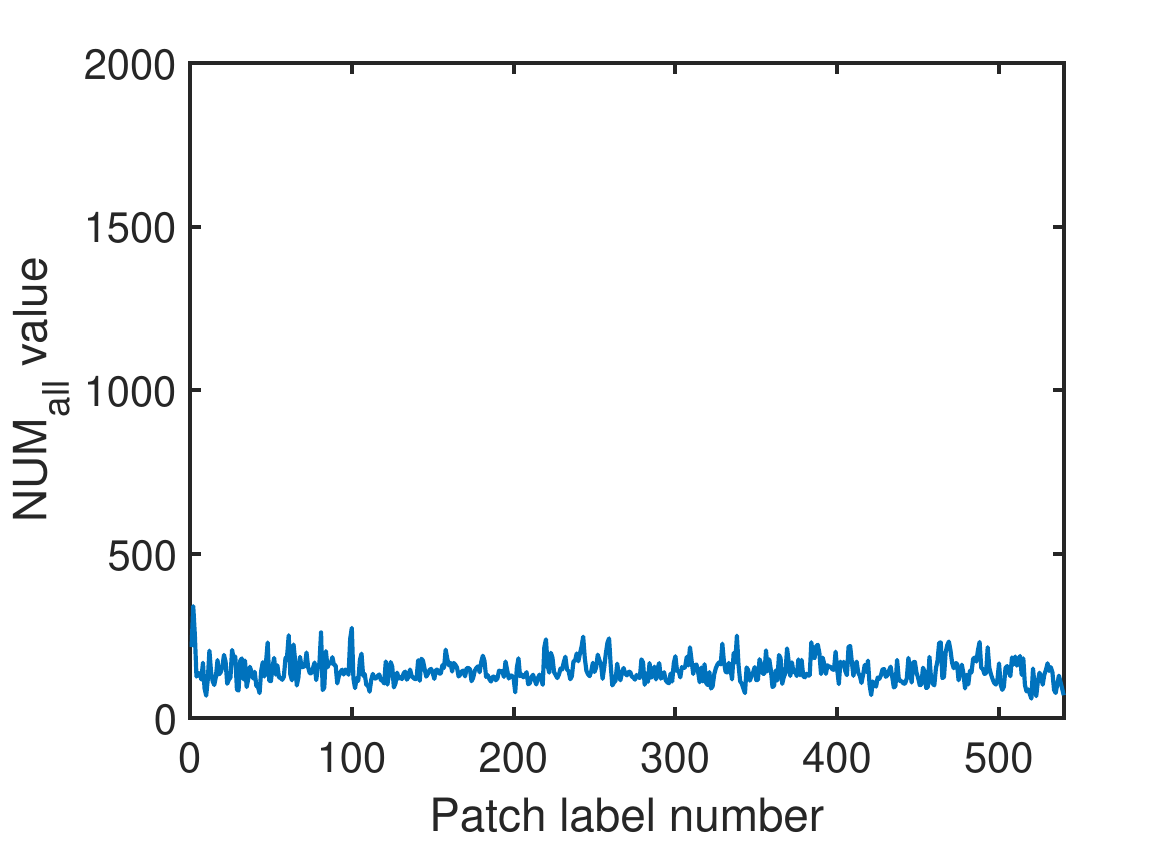}

\includegraphics[width=0.40\columnwidth]{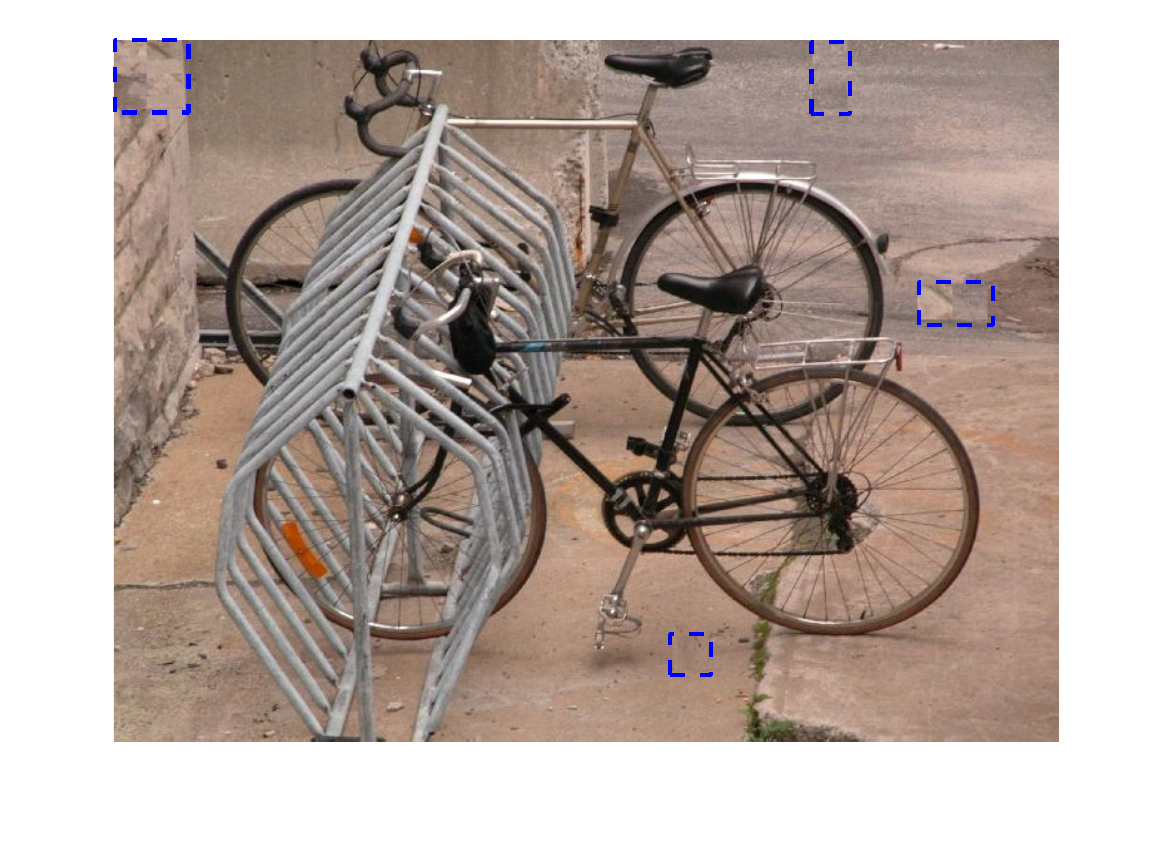}
\includegraphics[width=0.40\columnwidth]{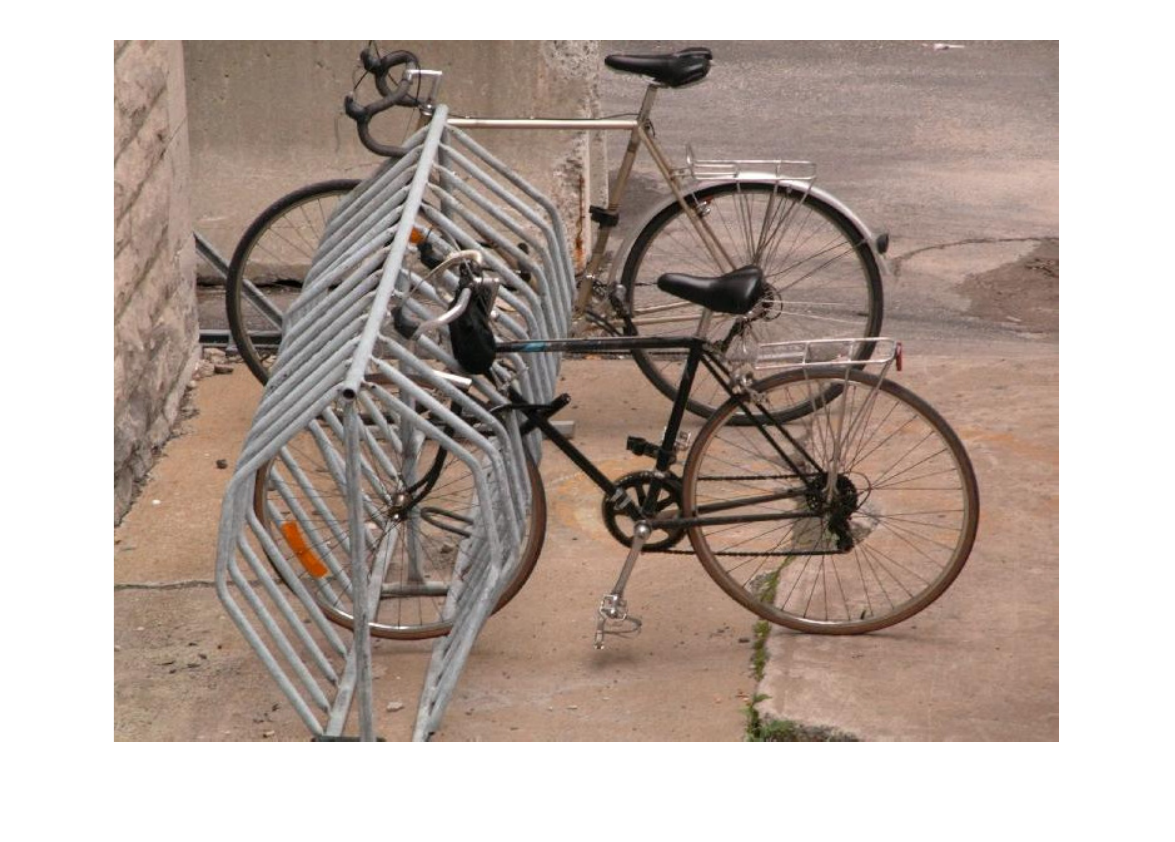}

\includegraphics[width=0.40\columnwidth]{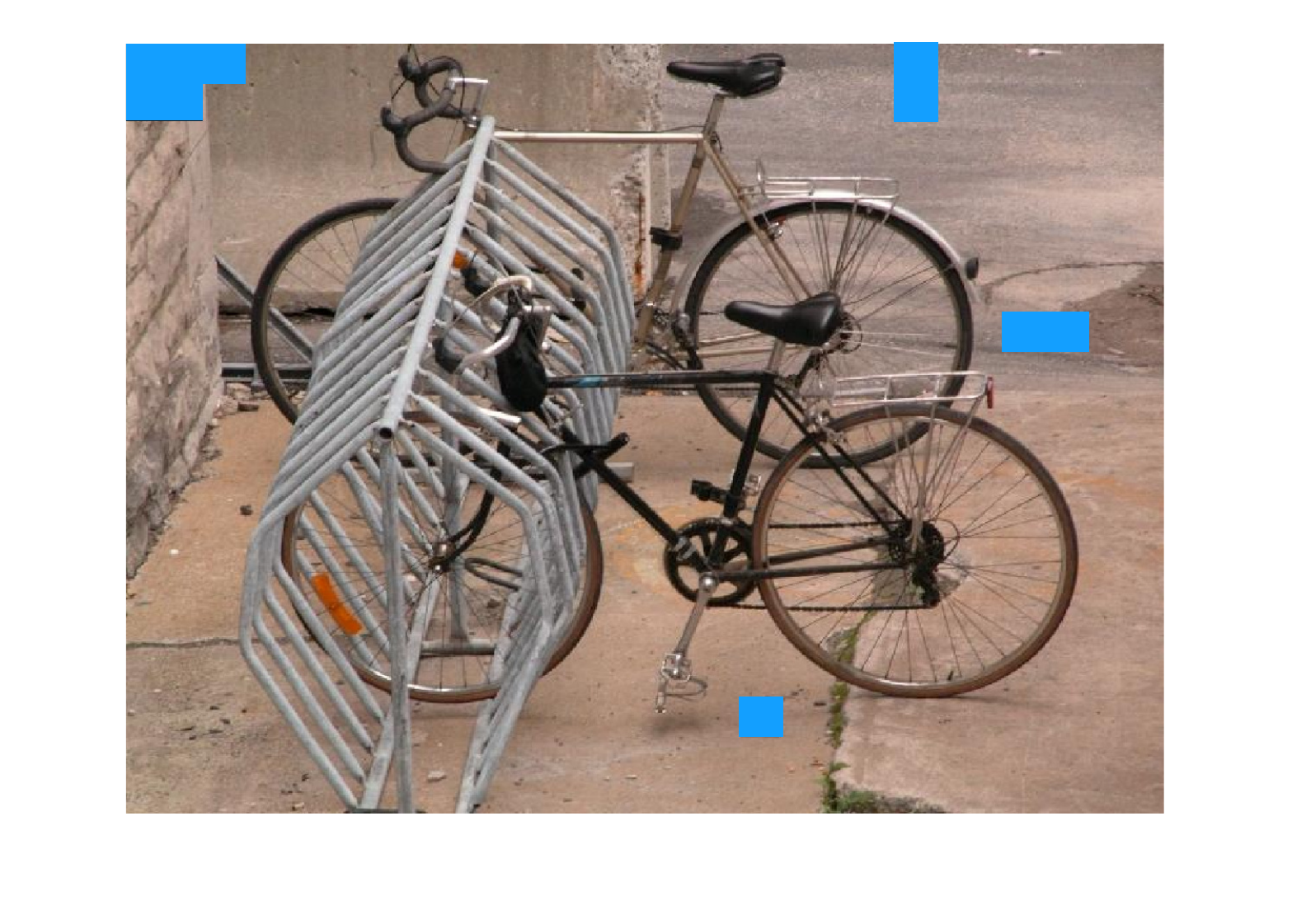}
\includegraphics[width=0.40\columnwidth]{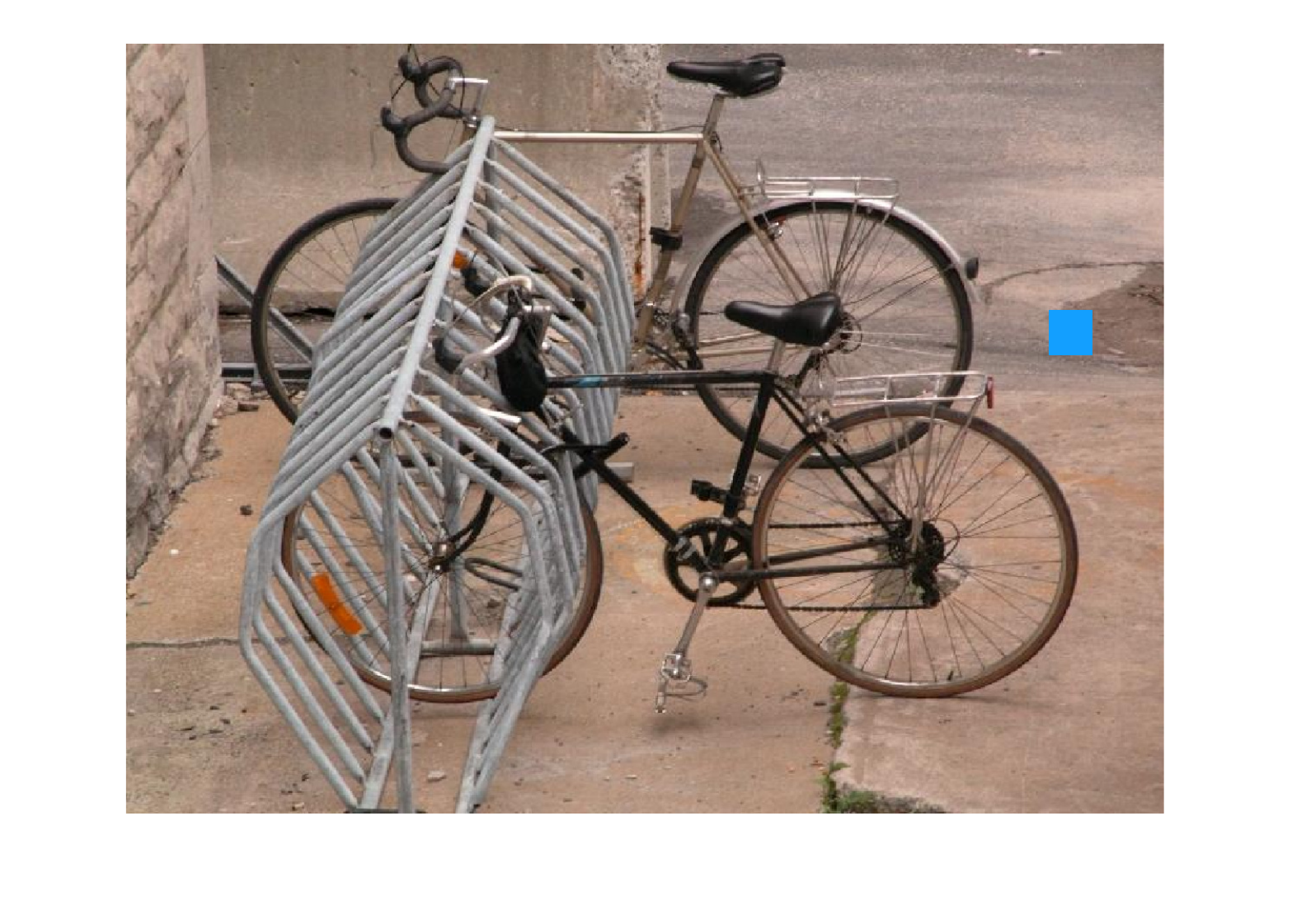}
\caption{Demonstration of the update step for $2$ iterations, described in \S\ref{sec:w_upd}, of a type 2 puzzle with 540 pieces each with sizes of $28 \times 28$. The first row shows the histograms of the $\NAM_{\all}$ metric values for all patches, defined in \eqref{eq:nb_met}. The second row shows the solution of the puzzle after each iteration of assembling the puzzle, and the third row shows the remaining patches of an assembled puzzle after removing the patches that are wrongly placed or oriented.}
\label{fig:update_res}
\end{figure}

If the puzzle is correctly assembled, the $\NAM$ values of all patches are relatively small as demonstrated for $\NAM_{\all}$ values in last figure of the first row of \Cref{fig:update_res}.
Otherwise, if there are some wrongly placed patches, their corresponding $\NAM$ values should be relatively higher.
This is demonstrated in the first row of \Cref{fig:update_res}, where the spikes of $\NAM_{\all}$ values correspond to wrongly orientated or placed patches.
Based on this observation, we suggest to find all patches
for which the corresponding $\NAM_{\all}$ value and at least one of the $\NAM_{\ltr}$, $\NAM_{\trb}$, $\NAM_{\blt}$ and $\NAM_{\lbr}$ values exceeds $1.5$ times the median of all corresponding $\NAM$ values.
We remove the corresponding edges from the grid. For example, for  $\NAM_{\ltr}$ we remove the edges connecting vertex $i$ with its left, top and right neighbors.
We refer to a location as empty if all edges connecting the patch in this location to its  top, bottom, left and right neighbors were removed.
Patches at empty locations at each iteration of this procedure are demonstrated in the second row of \Cref{fig:update_res}. Their removal, which literally creates empty locations, is demonstrated in the last row of this figure.

Next, we select all patches at empty locations for which at least 2 of 4 neighboring locations in the puzzle grid (including edges removed in the current process) are not empty. For each selected patch, we denote the set of neighboring patches  by $S_{\nb}$.
Note that by our selection criterion, $S_{\nb}$ contains either $2$, $3$ or $4$ indices.
For each selected patch, we find an oriented patch in $S_{\nb}$
that minimizes the averaged MGC metric.
We denote this minimal value by $\NAM_{S_{\nb}}$and also denote by $\med(\NAM_{\all})$ the median of all $\NAM_{\all}$ values.
For a selected patch $i$ and a patch $j \in S_{\nb}$, we fix the default parameters 0.3 and 0.6 and update the affinity function as follows:
\begin{align}
\label{eq:update_nb_aff_final}
\mW_{\est}(i, j) = \mW_{\est}(j, i) = \left\{
\begin{array}{ll}
0.6, & \mbox{ if } \NAM_{S_{\nb}} < \med(\NAM_{\all});\\
0.3, & \mbox{ if } \med(\NAM_{\all}) < \NAM_{S_{\nb}} < 2 \med(\NAM_{\all})
\end{array}\right. .
\end{align}
We also update the connection function as follows, where we denote by $\mR_{i, j}$ the current solution of the pairwise orientation of patch $i$ with respect to patch $j$,
\begin{equation}
\label{eq:update_nb_conn_final}
\mR_{\est}[i, j] = \mR_{i, j} \text{ and } \mR_{\est}[j, i] = \mR_{i, j}^T \text{ if } \NAM_{S_{\nb}} < 2 \med(\NAM_{\all}).
\end{equation}
\Cref{algo:updating} summarizes this update procedure.

\begin{algorithm}
\caption{Updating the affinity and connection functions at a given iteration}
\label{algo:updating}
\begin{algorithmic}
\State{\textbf{Input:} $\MGC$ metric values between all patches, current solution to the puzzle problem}
\begin{itemize}
\State Calculate the $\NAM$ values for all patches according to \eqref{eq:nb_met} and \eqref{eq:nb_met_3}
\State Remove all edges from the solution grid
for which the corresponding $\NAM_{\all}$ value and at least one of the $\NAM_{\ltr}$, $\NAM_{\trb}$, $\NAM_{\blt}$ and $\NAM_{\lbr}$ values exceeds $1.5$ times the median of all corresponding $\NAM$ values
\State For any patch at an empty location (that is, location whose all edges were removed), which has at least two non-empty neighboring locations (according to the natural grid of locations),
find the patch with the correct rotation which best fits in that position and update the affinity function and the connection function according to \eqref{eq:update_nb_aff_final} and \eqref{eq:update_nb_conn_final}
\end{itemize}
\State{\textbf{Return:} $G = \{V, E_{\est}, \mW_{\est}, \mR_{\est}\}$}
\end{algorithmic}
\end{algorithm}

Finally, our proposed algorithm for reassembling square jigsaw puzzles is summarized in \Cref{algo:type2_sol}.
It iteratively solves the puzzle by repeating the following 3 steps: finding the orientations of all patches, finding the locations of all patches and updating the connection function and the affinity function. To measure how good the solution is at each iteration, we recommend using the following metric
\begin{equation}
\label{eq:check_sol}
\begin{split}
\Err (\{R_i\}_{i = 1}^n, \sigma ) = \sum_{i = 1}^n ( \MGC_{\lr}(R_i \cdot P_i, R_{i_{\sigma, r}} \cdot P_{i_{\sigma,r}}) + \MGC_{\tb}(R_i \cdot P_i, R_{i_{\sigma, b}} \cdot P_{i_{\sigma, b}}) \\ + \MGC_{\rl}(R_i \cdot P_i, R_{i_{\sigma, l}} \cdot P_{i_{\sigma, l}}) + \MGC_{\bt}(R_i \cdot P_i, R_{i_{\sigma, b}} \cdot P_{i_{\sigma, b}}) ),
\end{split}
\end{equation}
where $i_{\sigma, t}, i_{\sigma, l}, i_{\sigma, b}$ and $i_{\sigma, r}$ are the indices of the neighbors of patch $i$ from top, left, bottom and right, respectively, according to the solution $\sigma$.
If patch $i$ is at the edge or corner of the puzzle grid, we only sum the respective MGC values.

\begin{algorithm}
\caption{Solution of type 2 puzzles}
\label{algo:type2_sol}
\begin{algorithmic}
\State{\textbf{Input:} Puzzle Patches: $\{P_i\}_{i=1}^n \subset \RR^{s \times s \times 3}$}
\begin{itemize}
\State Apply \Cref{algo:aff_gr} with $\{P_i\}_{i=1}^n$ to construct the Affinity Graph $G = (V, E, \mW, \mR)$ and obtain $\MGC$ values between all patches
\State Run \Cref{algo:rot_sol} with $G = (V, E, \mW, \mR)$ to find the orientations $\{ R_i \}_{i=1}^n$
%\State \textbf{Optional:} {Update the MGC metric according to the procedure explained in \S\ref{sec:VDD_est}}
\State Apply the type 1 jigsaw puzzle solver of \cite{yu2015solving} to solve the type 1 puzzle with patches $\{R_i \cdot P_i\}_{i=1}^n$ and obtain their estimated permutation vector $\sigma$
\State Compute and record $\Err (\{R_i\}_{i = 1}^n, \sigma )$ by \eqref{eq:check_sol}
\For{iterations 1:5}
\begin{itemize}
\item[$\bullet$] Apply \Cref{algo:updating} with $\sigma$, $\{ R_i \}_{i=1}^n$ and the MGC values to obtain the updated connection graph $G = (V, E, \mW, \mR)$
\item[$\bullet$] Apply \Cref{algo:rot_sol} with $G = (V, E, \mW, \mR)$ to recover the orientations $\{ R_i \}_{i=1}^n$
%\State \textbf{Optional:} {Update the MGC metric according to the procedure explained in \S\ref{sec:VDD_est}}
\item[$\bullet$] Apply the type 1 jigsaw puzzle solver of \cite{yu2015solving} to solve the type 1 puzzle with patches $\{R_i \cdot P_i\}_{i=1}^n$ and obtain their estimated permutation vector $\sigma$
\item[$\bullet$] Compute and record $\Err (\{R_i\}_{i = 1}^n, \sigma )$ by \eqref{eq:check_sol}
\end{itemize}
\EndFor
\end{itemize}
\State{\textbf{Return:} $\{R_i\}_{i = 1}^n$ and $\sigma$, which minimize $\Err (\{R_i\}_{i = 1}^n, \sigma )$ among all the above choices}

\end{algorithmic}
\end{algorithm}

We remark that most state-of-the-art methods use a greedy step to make final corrections to the solved puzzle.
On the other hand, the step discussed here only updates the connection graph and is thus non-greedy.
It is possible to incorporate greedy procedures that may improve the performance of our algorithm, however, we would like
to show that a more principled method can be competitive.

The above description of our algorithm mentions various parameters and we further review them in Section \ref{sec:parameter_discussion} of the supplemental material. All parameters, but the number of iterations (which can be rather small), are of graph affinities. That is, throughout the algorithm, various tests are performed, and based on these, the affinities of the estimated graph are changed. While the relative order of these parameters was often clear to us, we performed some experiments to get the actual effective range of these parameters, and noticed some stability to changes of the chosen parameters. We remark that the use of unknown parameters are common in other puzzle solvers and other graph-based algorithm.

\subsection{Time Complexity of \Cref{algo:type2_sol}}
\label{sec:time_complex}

Empirically, the most time consuming step is finding the MGC metric between all puzzle pieces.
This step is vital for all jigsaw puzzle solvers. Its order of operations is
$O(n^2 d),$ where $n$ is the number of image patches and $d$ is the number of pixels in the side of each square image patch.
One may parallelize this procedure and achieve faster computation.

After finding the MGC metric between all puzzle pieces, our proposed procedure
constructs the connection graph by following \Cref{algo:aff_gr}.
The main computation of this step is in finding the nearest neighbors of each point in the lattice, which is of order $O(n^2)$.
Next, our procedure finds the orientations of all patches. Here, the main computation is in finding the top two eigenvectors of a sparse symmetric matrix with 4 nonzero elements in each column and row.
Using Lanczos algorithm, the time complexity of this step is of order $O(n)$.
Once the orientations are computed, the type 2 puzzle problem becomes type 1.
The complexity of solving the type 1 puzzle depends on the state-of-the-algorithm being used.

The last step of the proposed procedure is the update step, which is summarized in \Cref{algo:updating}. It first requires finding a median of an array of size $n$, whose computation is of order $O(n \log(n))$.
%(some randomized algorithms find the median in linear time).
It then requires filling up the empty locations, whose computation is of order $O(\text{\#(of empty locations)} \times n)$, where the \# of empty locations is at most $n/2$. This step recurs at most 5 times.

In summary, the complexity of the whole procedure, excluding the chosen type 1 solver, is of order $O(n^2)$. A main computation is that of the MGC metric, where the constant multiplying $n^2$ seems to be large in practice. As we mention below, in theory, the worst-case complexity of the type 1 solver of Yu et al.~\cite{yu2015solving} can be higher than $O(n^2)$, but we did not see any evidence for this in our numerical experiments in \S\ref{sec:num_exp}.

At last, we discuss the complexity of the type 2 solvers of Gallagher~\cite{gallagher2012jigsaw} and Yu et al.~\cite{yu2015solving}. We first mention that they both implement greedy procedures, so it is hard to bound their complexity.
Moreover, Yu et al.~\cite{yu2015solving} use multiple iterations (till convergence) with no guarantees of convergence.
Nevertheless, we discuss two of their main algorithms. Gallagher~\cite{gallagher2012jigsaw} uses Kruskal's algorithm to find the minimum spanning tree, whose complexity is of order $O\left(n^2 \log(n)\right)$ \cite{cormen2009introduction}, though another implementation for the minimum spanning tree has complexity of order $O\left(n^2 \log^{*}(n)\right)$ \cite{fredman1987fibonacci}. Yu et al.~\cite{yu2015solving} solve a linear program at each iteration, where the worst-case complexity of linear programs is $O\left(n^{2.5}\right)$ \cite{vaidya1989speeding}.
We remark that the same bound on the order of complexity holds for the type 1 solver, which we use in our algorithm. However, our numerical experiments in \S\ref{sec:num_exp} may indicate different orders of complexity for the type 1 and type 2 puzzle solvers of Yu et al.~\cite{yu2015solving}. One component which makes their type 2 puzzle solver slower is the artificial enlargement of the number of puzzle patches by 4 in the former one, but this only increases the complexity in a constant factor.

\section{Numerical Experiments}
\label{sec:num_exp}

%In this section we show n

%\subsection{Results for Type 2 Puzzles}

\begin{figure}
\centering
\includegraphics[width=0.47\columnwidth]{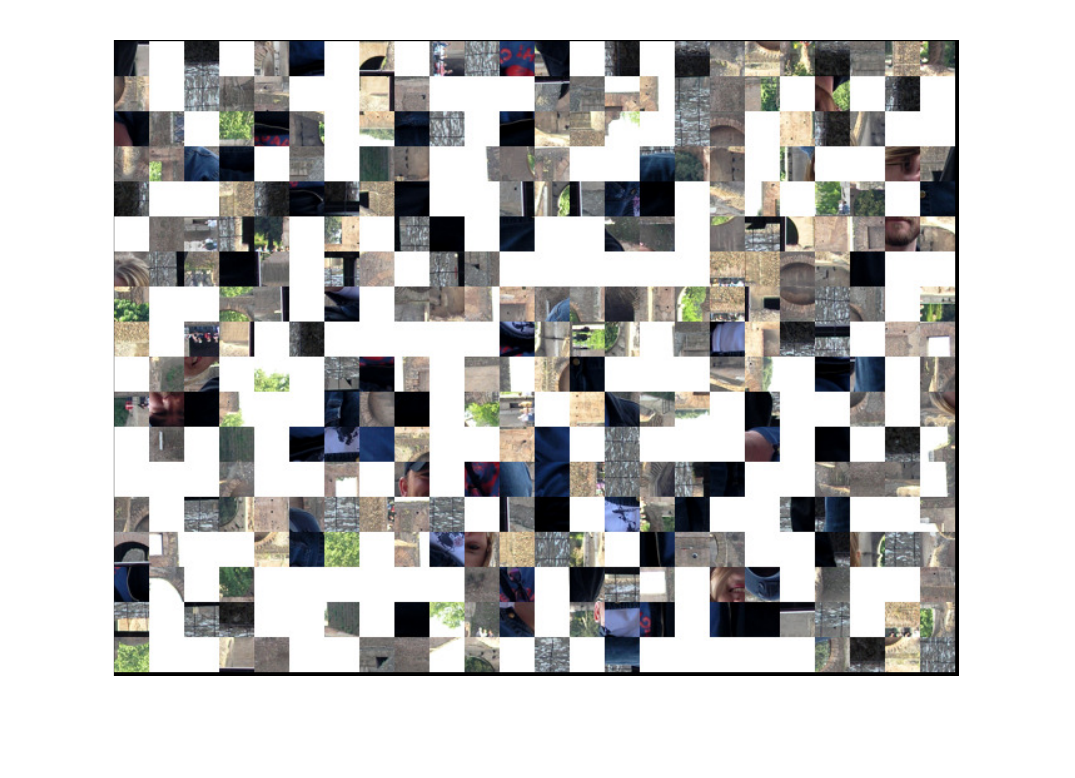}
\includegraphics[width=0.47\columnwidth]{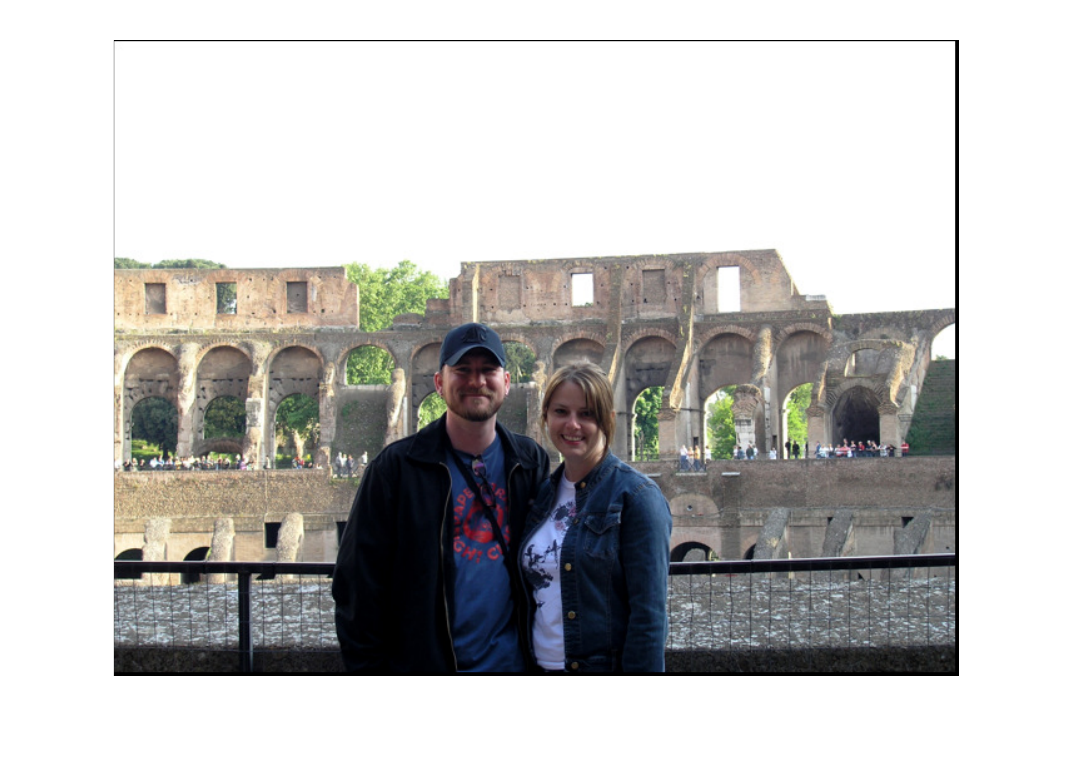}

\includegraphics[width=0.47\columnwidth]{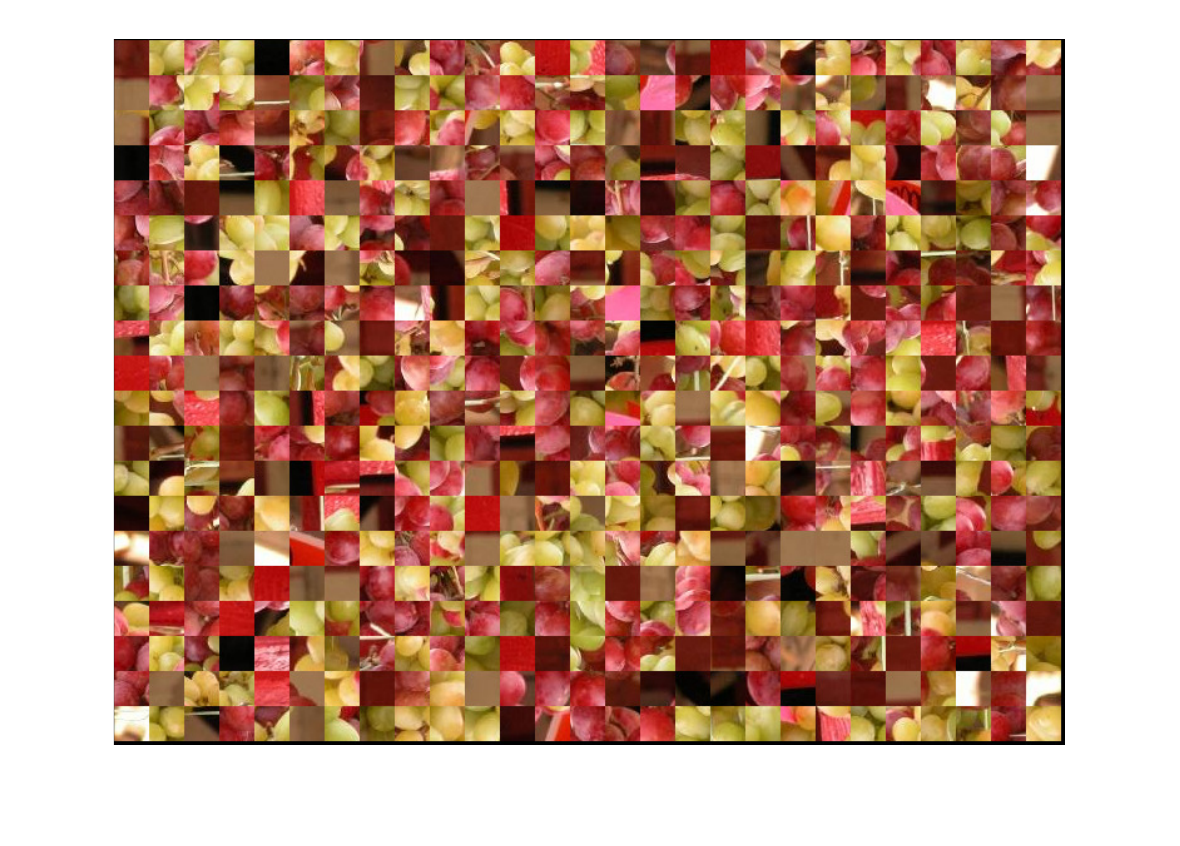}
\includegraphics[width=0.47\columnwidth]{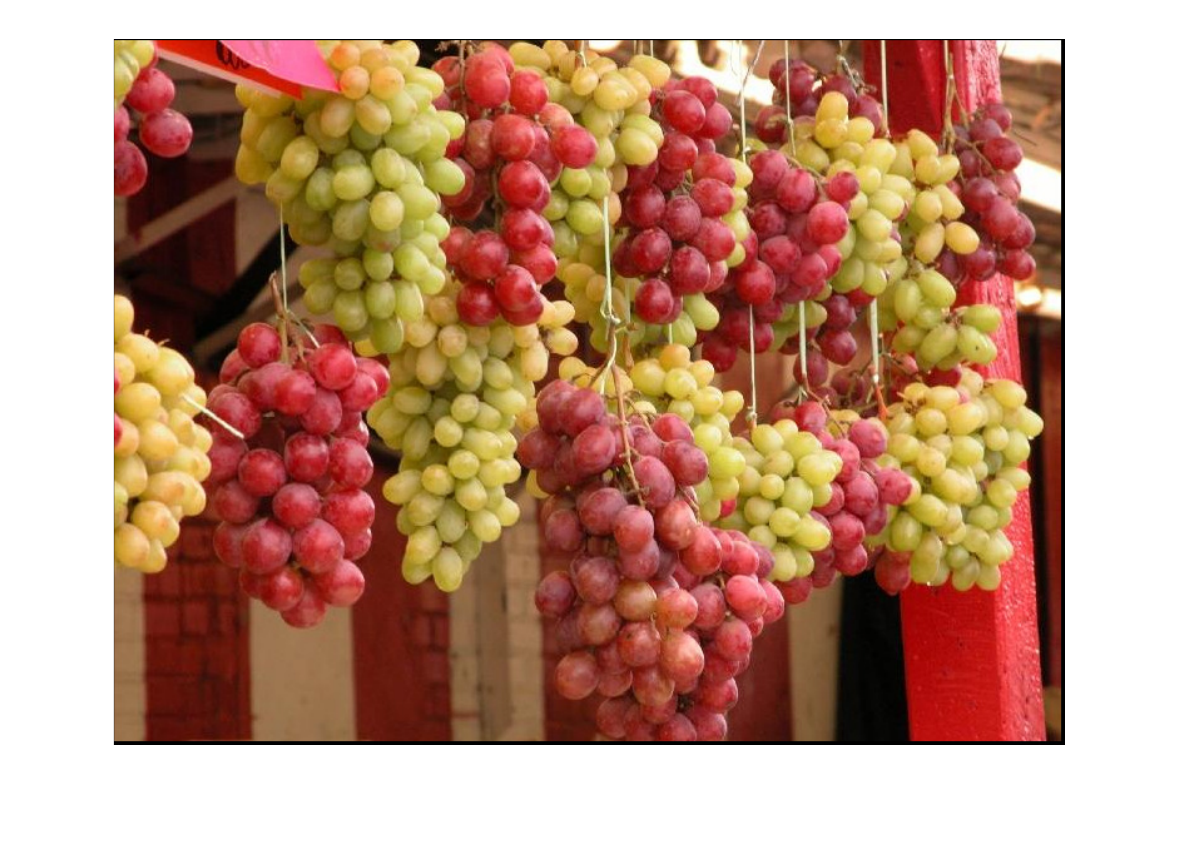}

\includegraphics[width=0.47\columnwidth]{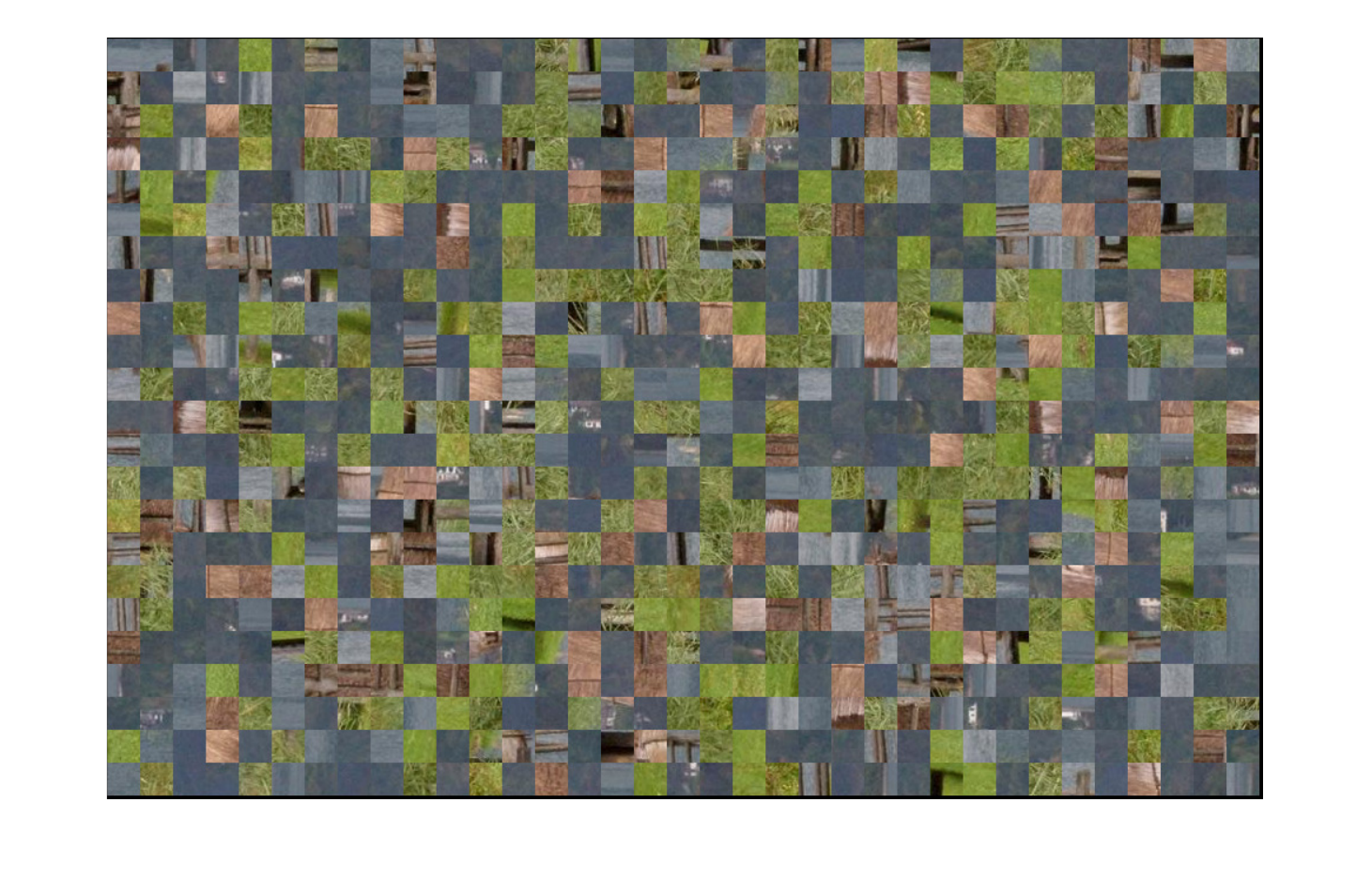}
\includegraphics[width=0.47\columnwidth]{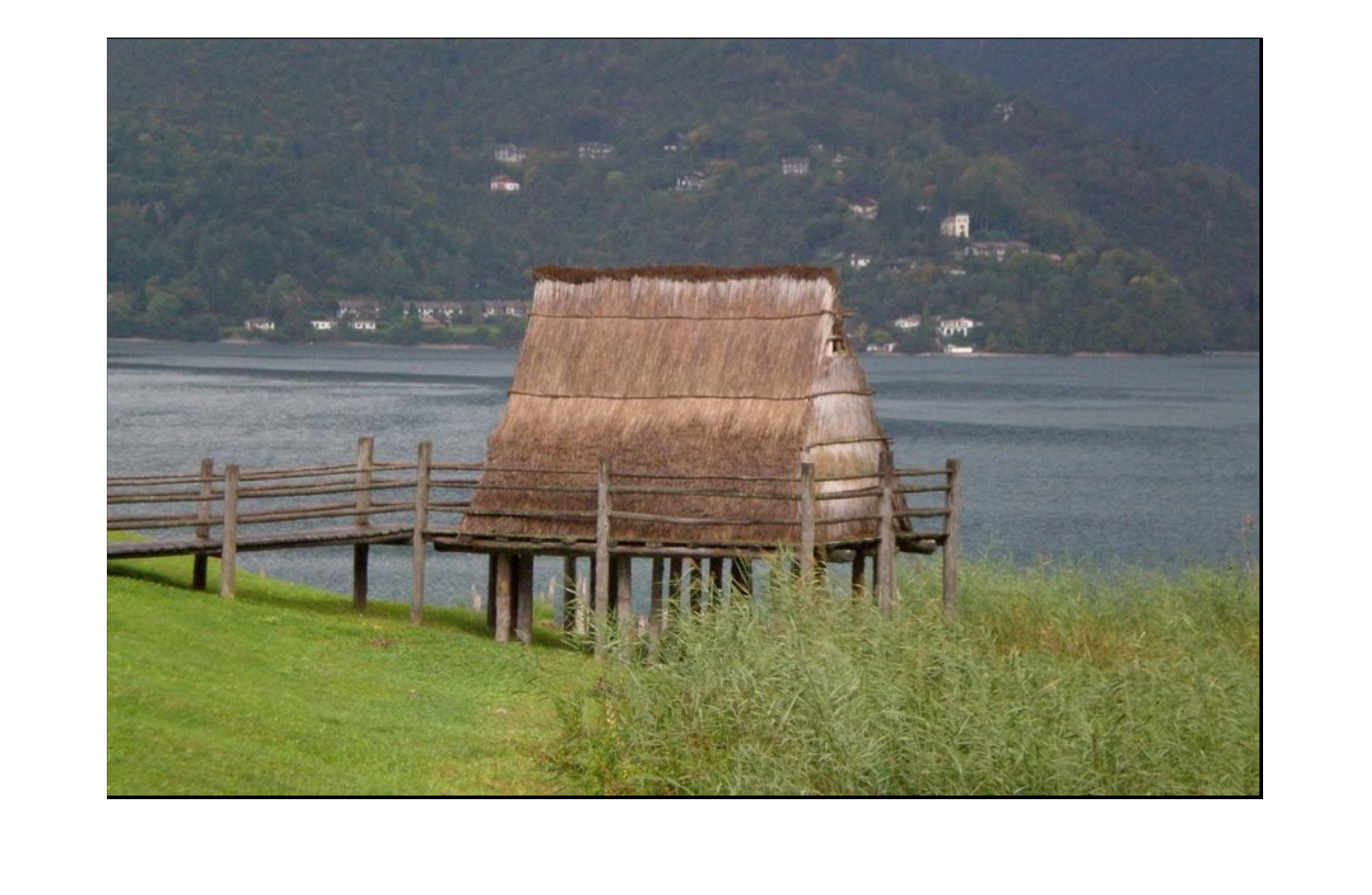}

\includegraphics[width=0.47\columnwidth]{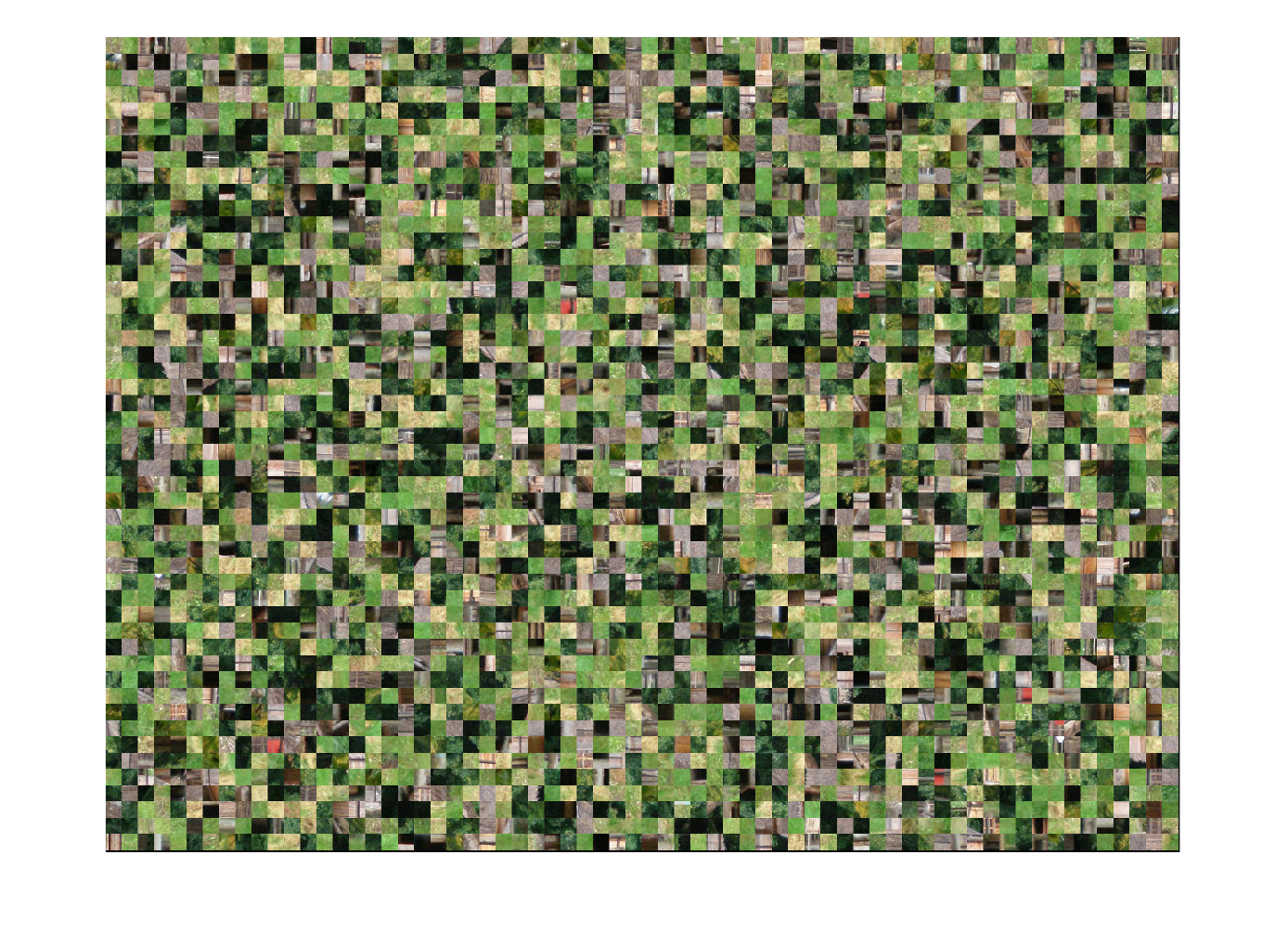}
\includegraphics[width=0.47\columnwidth]{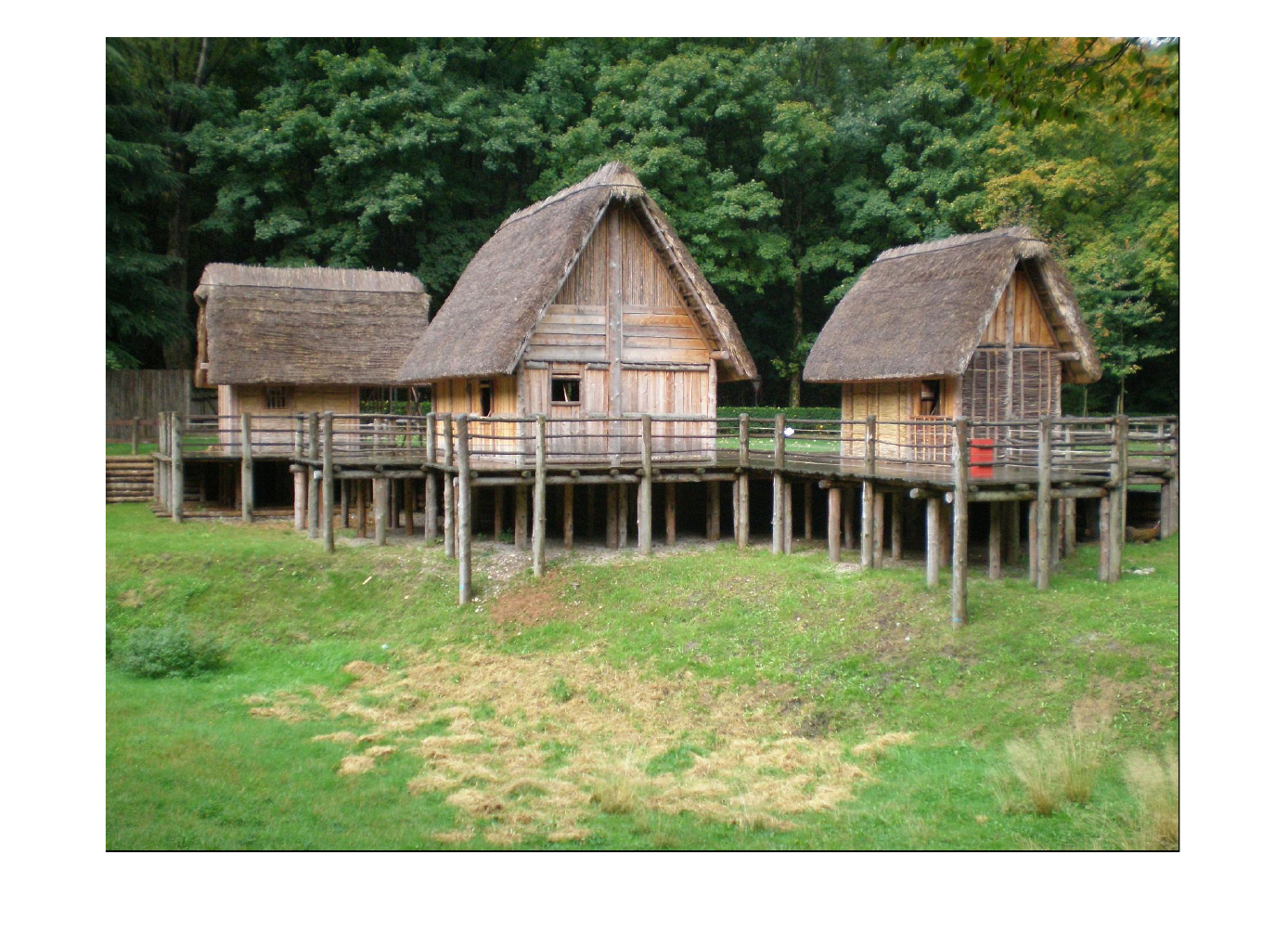}
\caption{Solutions by \Cref{algo:type2_sol} of type 2 puzzles representing the four datasets.
The images in the left column are the inputs for the algorithm and the ones in the right column are the outputs generated by our proposed algorithm.
The puzzle in first row is from the MIT dataset with $432$ patches, the puzzle in the second row is from the McGill dataset with $540$ patches,
the puzzle in the third row is from the Pomeranz dataset with $805$ patches and the puzzle in the fourth row is from the large Pomeranz dataset with $3300$ patches.}
\label{fig:num_res}
\end{figure}

We apply our proposed algorithm to solve square jigsaw puzzles of the following standard image datasets: the MIT dataset from Cho et al.~\cite{cho2010probabilistic}, which contains $20$ images,
each with $432$ patches, and three datasets from Pomeranz et al.~\cite{pomeranz2011fully}, where the first two, which are referred to as McGill and Pomeranz, include $20$ images with
$540$ and $805$ patches, respectively, and the third one has $3$ images with $3300$ patches, which is also referred to as large Pomeranz. For all datasets, the patches are of size $28 \times 28$. \Cref{fig:num_res} demonstrates the application of our proposed algorithm to four images that represent the four datasets. We remark that our algorithm perfectly solves the last 3 puzzles and we later discuss the solution of the first puzzle.
To test the accuracy of our proposed algorithm we use the following four metrics, defined in Gallagher~\cite{gallagher2012jigsaw} and Cho et al.~\cite{cho2010probabilistic}:
the direct comparison, the neighbors comparison, the largest component and the perfect reconstruction. The direct comparison measures the percentage of image patches whose location and orientation are correct.
The neighbors comparison calculates the percentage of pairs of image patches that are matched correctly.
The largest component calculates the percentage of patches in the largest correctly assembled component of the solved puzzle.
Finally, the perfect reconstruction of a puzzle is $1$ if it is solved correctly and $0$ otherwise.

We compared our algorithm with those of Gallagher~\cite{gallagher2012jigsaw} and Yu et al.~\cite{yu2015solving} for type 2 puzzles since they were the only algorithms with available codes (we requested codes from all authors of relevant published algorithms).
%While our algorithm is deterministic,
Note that we use the type 1 and type 2 solvers of Yu et al.~\cite{yu2015solving} in two different ways: We apply the former one as a component of \Cref{algo:type2_sol} and the latter one for comparing performance with \Cref{algo:type2_sol}.
One of the many procedures in our algorithm is random and described in \S\ref{sec:making_conn}. We have also noticed some randomness in the results of  the other two algorithms. Therefore, for each puzzle we run each algorithm 20 times and report the averaged result. To get an idea of the randomness of the three algorithms, we summarize here the averaged standard deviations of the neighbors comparison metric  when applying each algorithm 20 times to each of the 20 puzzles in the MIT dataset and averaging over the 20 puzzles. These averaged standard deviations for  Gallagher~\cite{gallagher2012jigsaw}, Yu et al.~\cite{yu2015solving} and our algorithm, are $6.5$, $1.5$ and $0.17$, respectively.
In this and other experiments, we notice that the randomness of our algorithm is not significant.

\begin{figure}
\centering
\includegraphics[width=0.32\columnwidth]{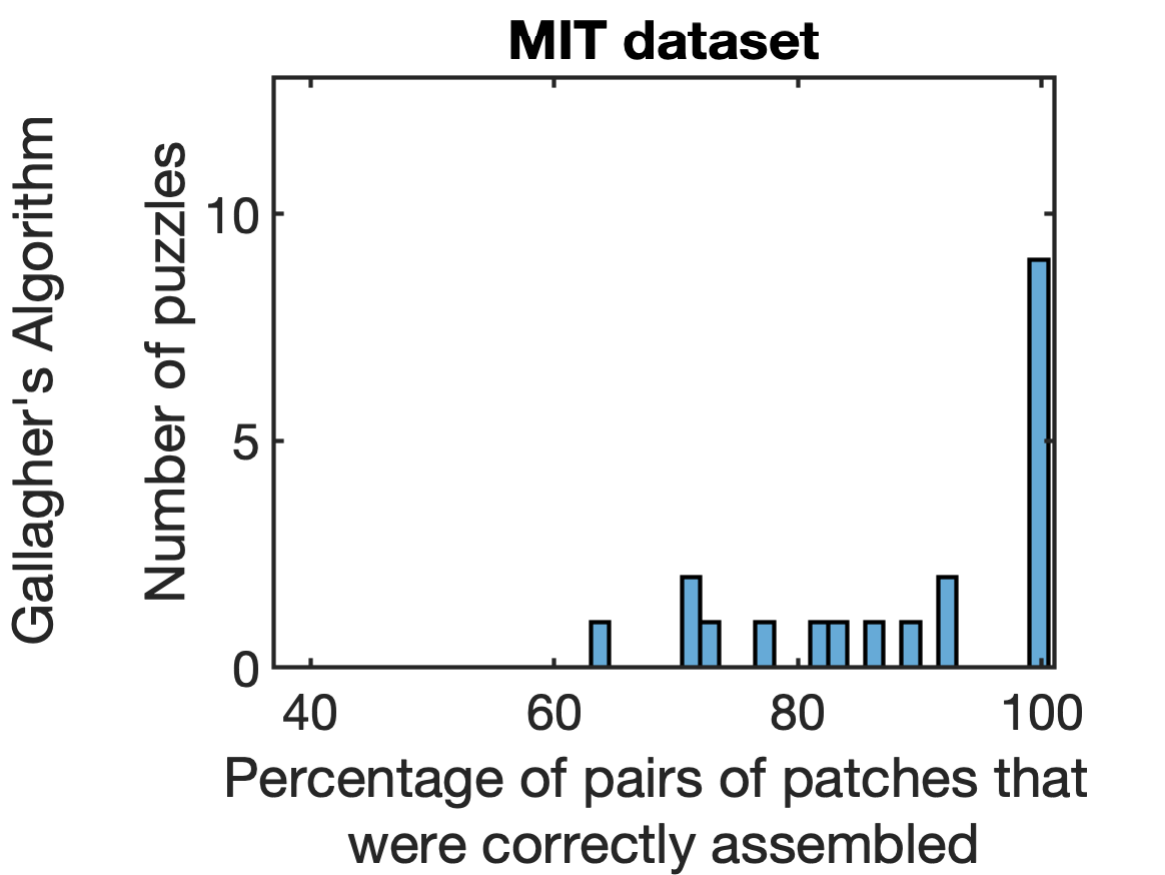}
\includegraphics[width=0.32\columnwidth]{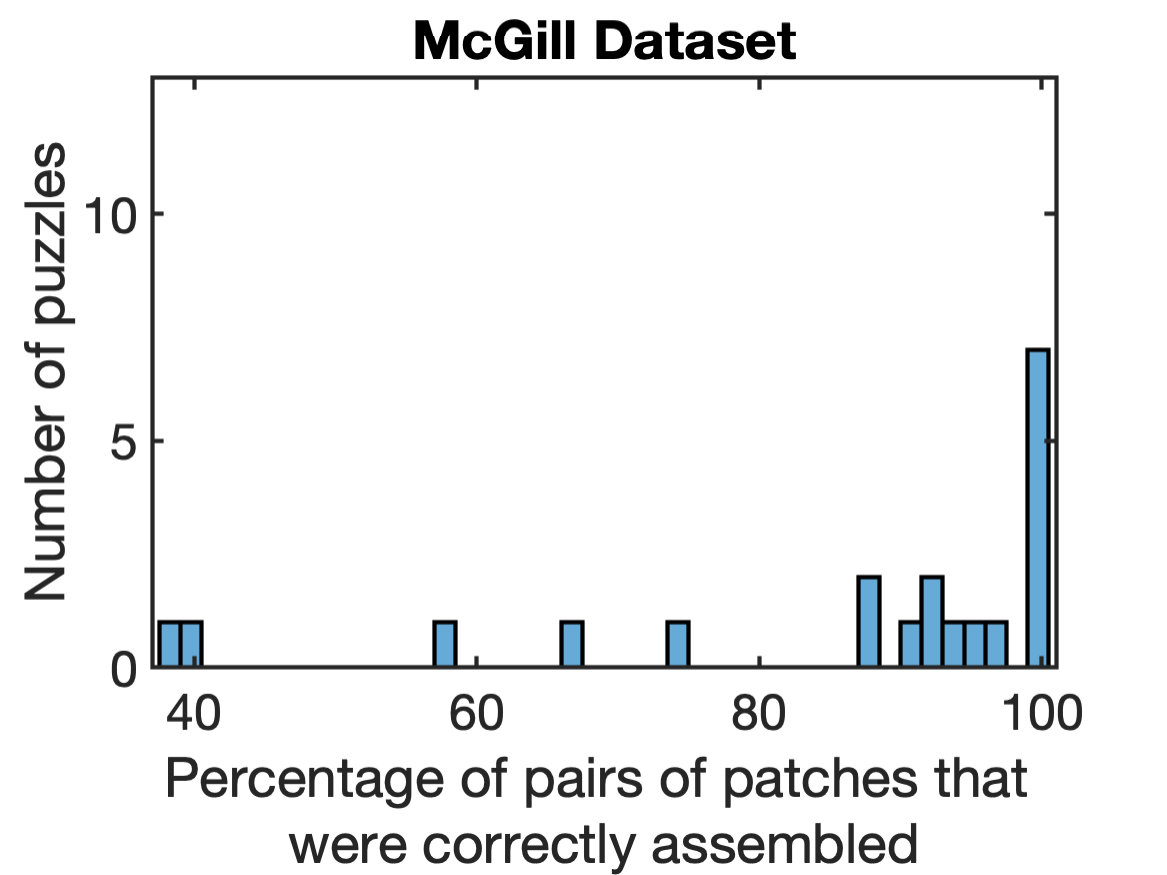}
\includegraphics[width=0.32\columnwidth]{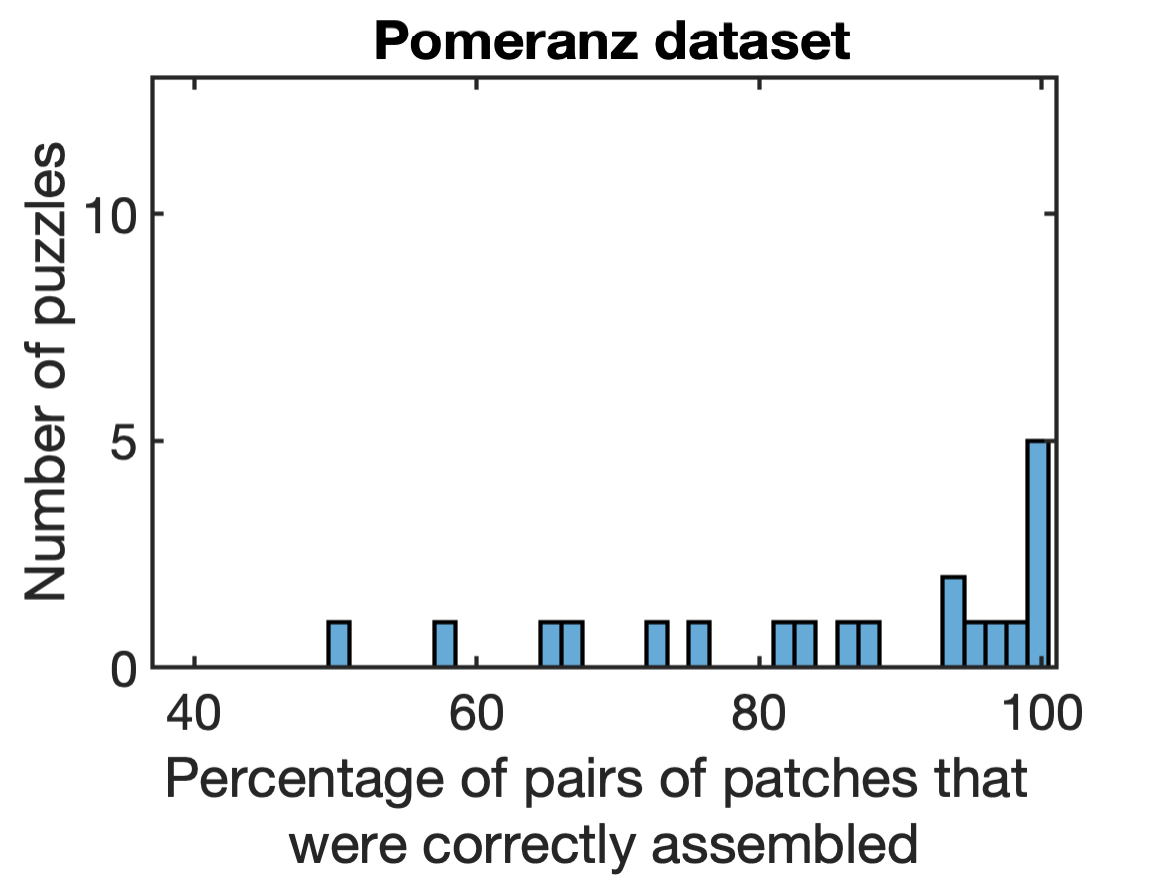}

\includegraphics[width=0.32\columnwidth]{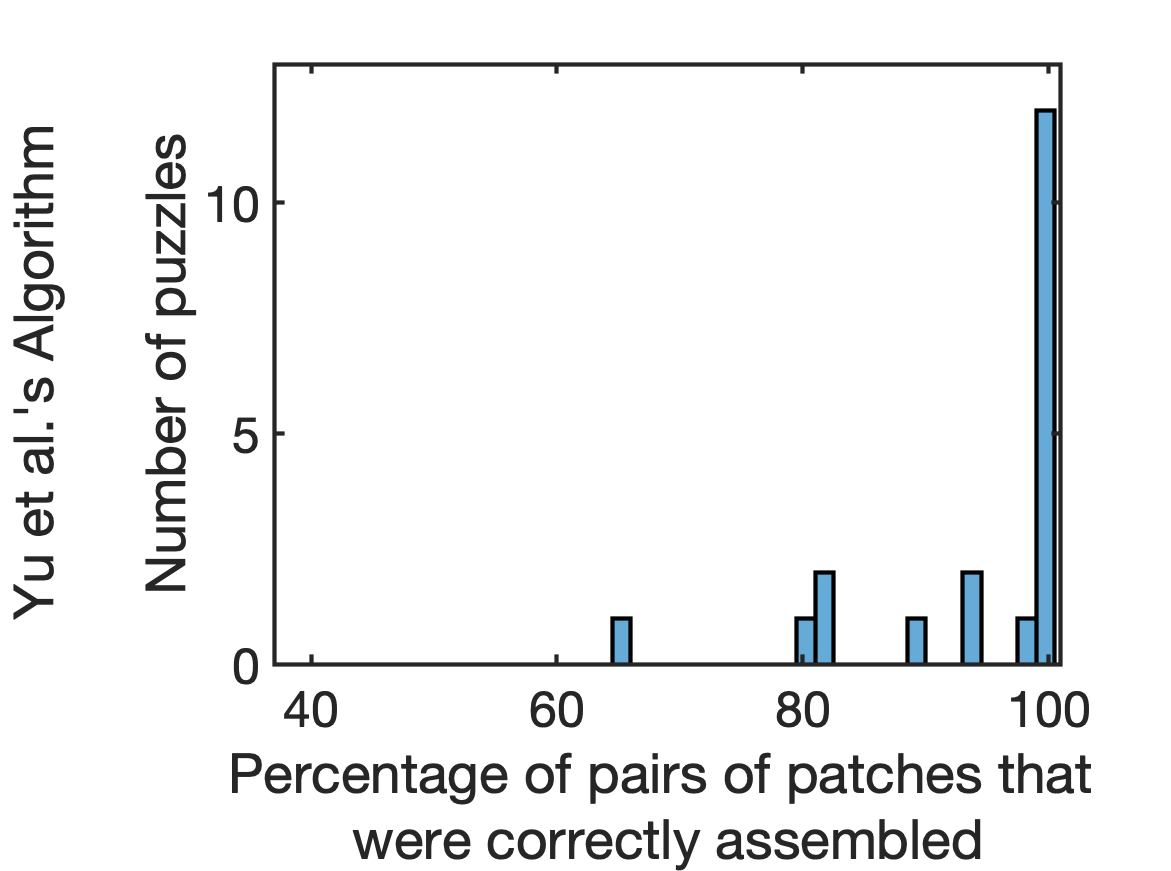}
\includegraphics[width=0.32\columnwidth]{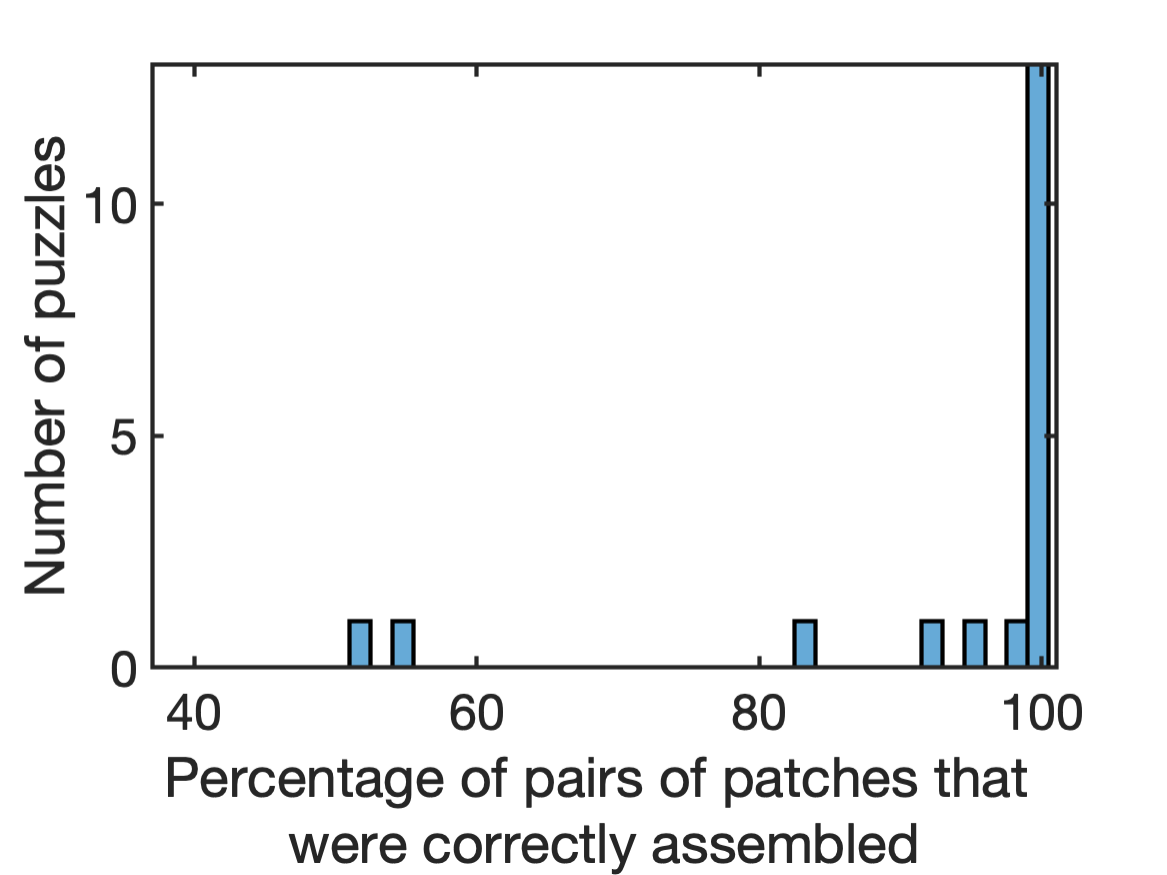}
\includegraphics[width=0.32\columnwidth]{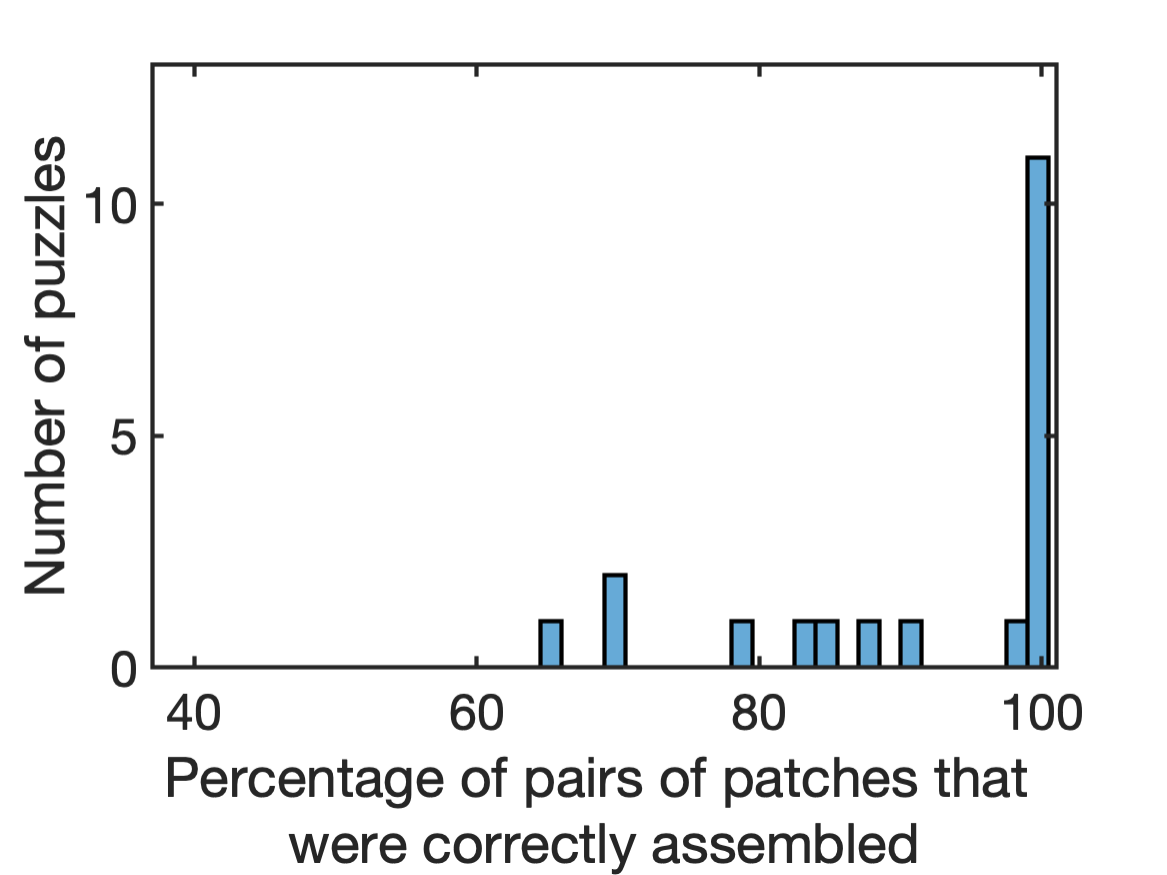}

\includegraphics[width=0.32\columnwidth]{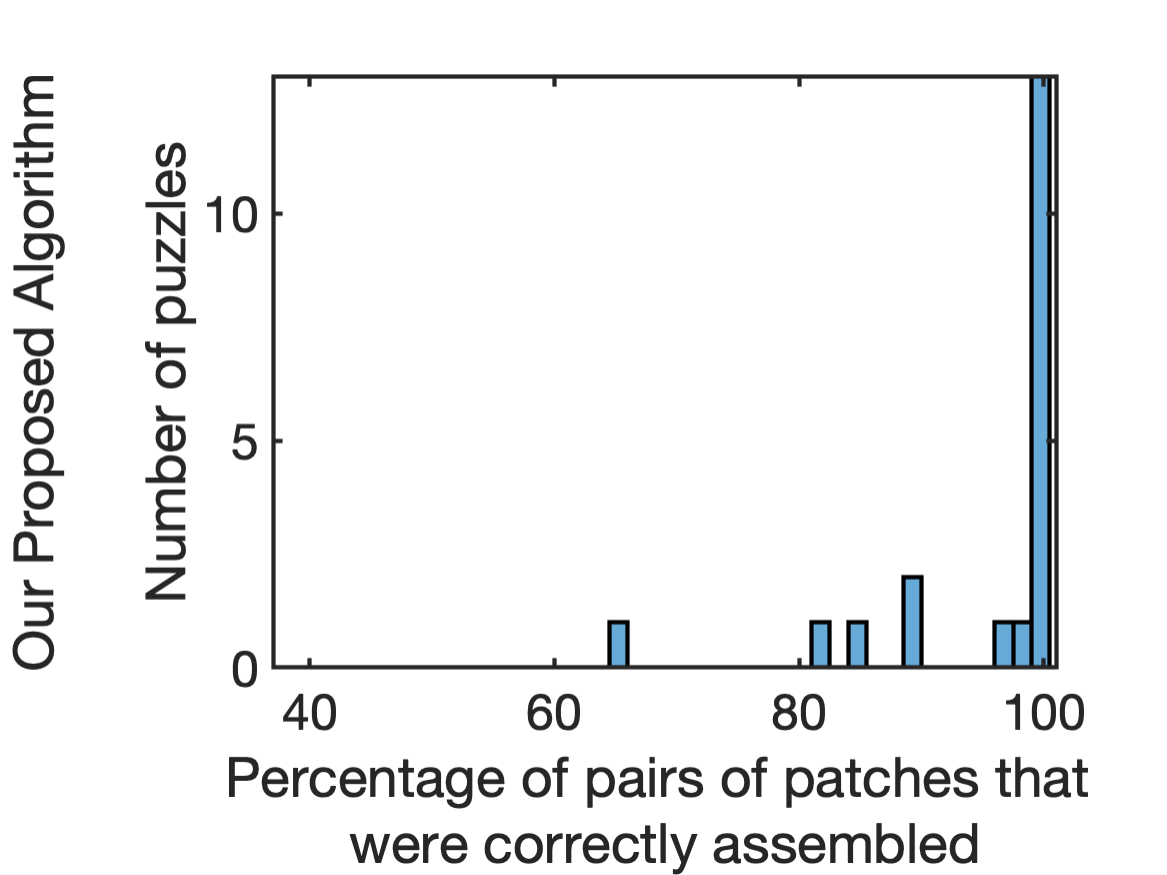}
\includegraphics[width=0.32\columnwidth]{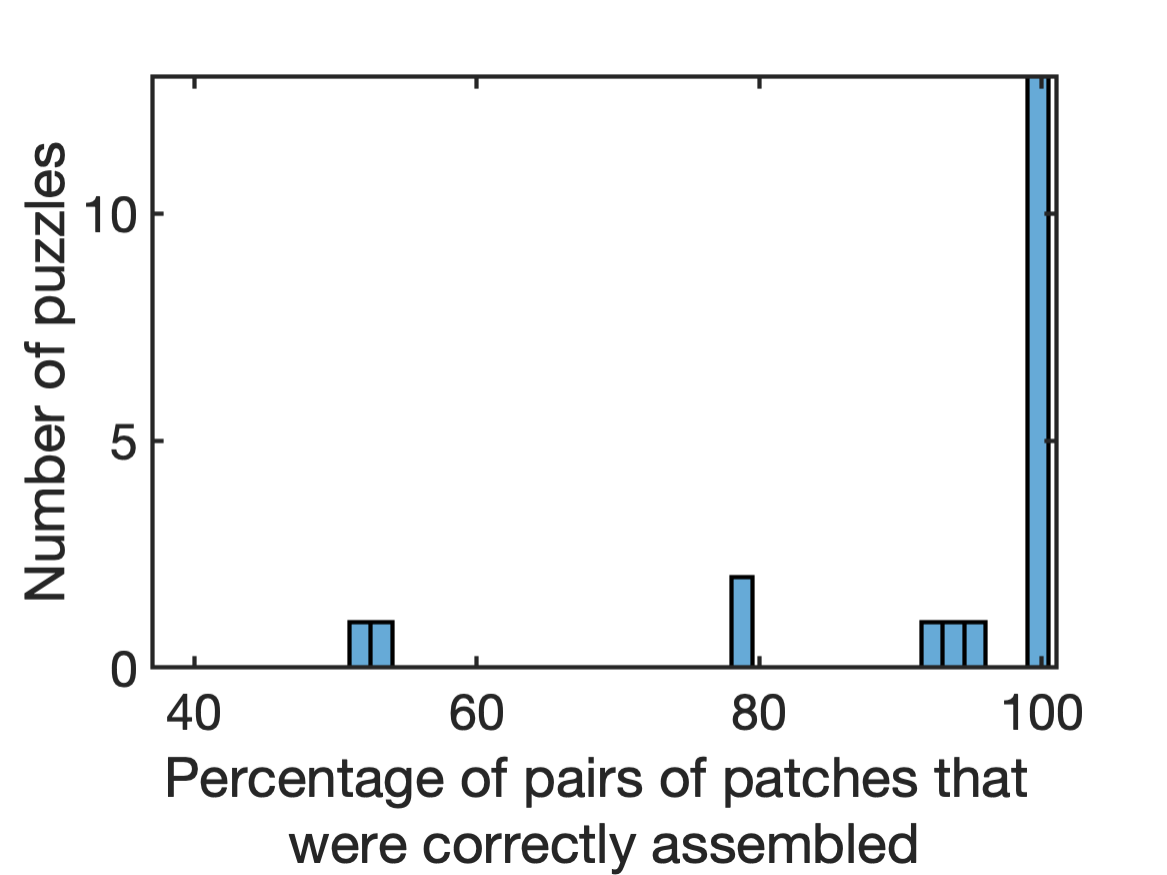}
\includegraphics[width=0.32\columnwidth]{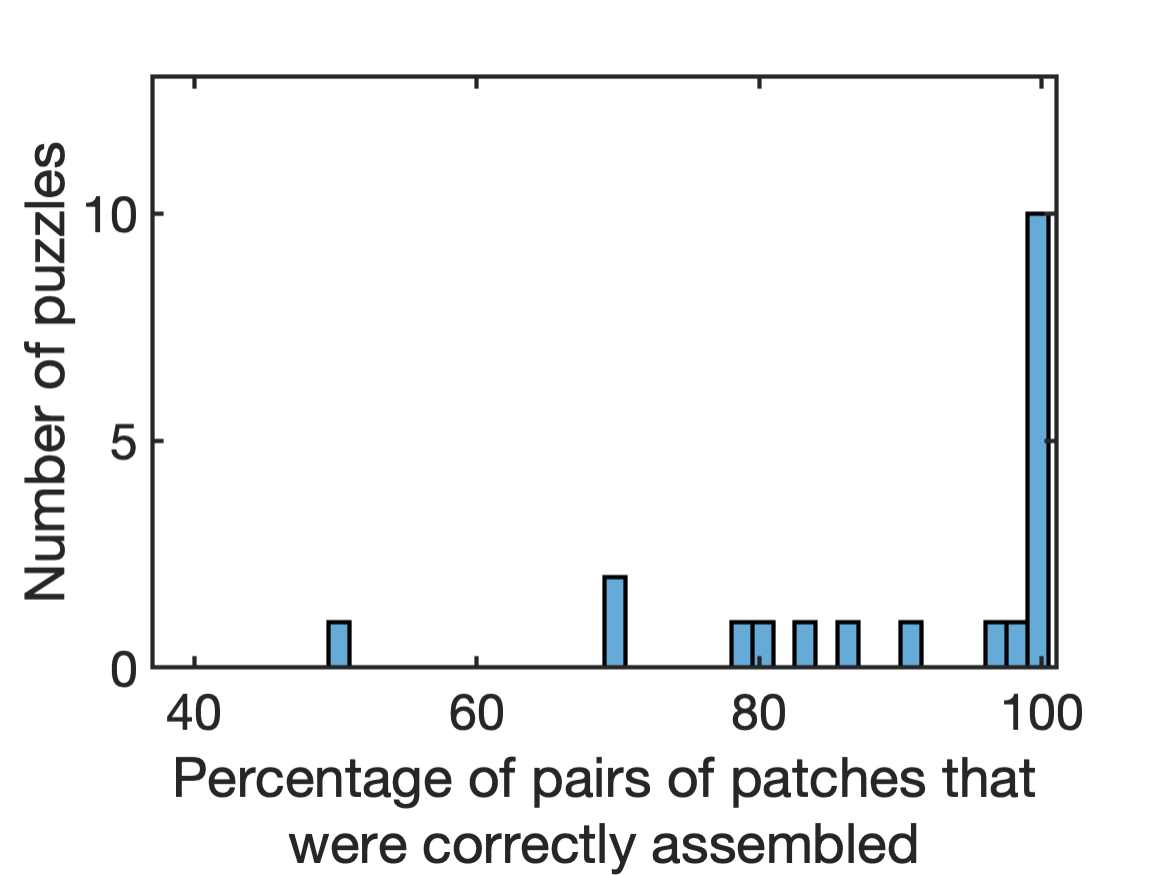}
\caption{Histograms of the neighbors comparison metric (percentage of the recovered pairs of patches).
The three rows correspond to results by the algorithm of \cite{gallagher2012jigsaw}, the algorithm of \cite{yu2015solving} and our proposed algorithm, respectively.
The three columns correspond to the MIT, McGill and Pomeranz datasets, respectively. }
\label{fig:hist_all_dat}
\end{figure}

\Cref{tab:comp_table} compares the four metrics of the three algorithms.
For the first three metrics of percentages, we report the means and standard deviations among the images in each dataset and among the 20 instances per image.
We remark that in the above paragraph, the standard deviations were different as they were computed among 20 instances per image and then averaged among all images in a dataset.
We only report the value of the fourth metric, that is, the number of perfectly
solved images.
One can use this reported metric and the total number of images to compute the mean and standard deviation of perfect reconstruction among the images in a dataset, and we thus
do not find it necessary to include it. We also remark that the standard deviations of this metric among the 20 instances were always zero.
\Cref{fig:hist_all_dat} presents histograms of the neighbors comparison metric for the first three datasets and the three different algorithms. The fourth dataset is excluded from this figure since it only has three images.
Histograms for the other metrics, which are not provided here, indicate similar comparisons of the three algorithms.

\begin{table}[]
\centering
\caption{Comparison of results for type 2 puzzles.
For the first three metrics (direct, neighbor and largest), we report the mean values and standard deviations over all the images in a dataset and over 20 instances of solving each puzzle. For the fourth metric, we report the total number of perfectly solved images (which was identical to all 20 instances per puzzle).}
\label{tab:comp_table}
\begin{tabular}{|l|l|l|l|l|l|l|l|l|}
\hline
\multicolumn{1}{|l|}{Dataset} & Method                                                & \multicolumn{2}{c|}{Direct} & \multicolumn{2}{c|}{Neighbor} & \multicolumn{2}{c|}{Largest} & Perfect \\ \hline
& & mean & std & mean & std & mean & std & \\ \hline
MIT         & Gallagher~\cite{gallagher2012jigsaw} & 84.2 & 19.7 & 89.1 & 12.4  & 87.2 & 14.3 &  9  \\ \cline{2-9}
(20 images,            & Yu et al.~\cite{yu2015solving}       & 95.5 &  13.0 & 95.4 &  8.7 & 95.4 & 13.2 &  13 \\ \cline{2-9}
432 patches)	& Our method                           & 94.8 &  11.3 & 95.2 & 9.2 & 95.4 & 9.1  &  13 \\ \cline{2-9}
\hline
McGill            & Gallagher~\cite{gallagher2012jigsaw} & 77.2 & 35.3 & 85.8 &  19.8 & 84.6 & 21.3 & 7  \\ \cline{2-9}
(20 images,		     & Yu et al.~\cite{yu2015solving}	     & 92.9 & 24.6 & 93.5 & 14.8 & 93.1 & 15.4 & 13        \\ \cline{2-9}
540 patches)	& Our method 							 & 88.3 & 25.6 & 92.2 & 15.2 & 91.4 & 17.2 & 13 \\ \cline{2-9}
\hline
Pomeranz        & Gallagher~\cite{gallagher2012jigsaw} & 77.5 & 27.8 & 85.3 & 15.5 & 79.3 & 22.6 & 5 \\ \cline{2-9}
(20 images                              & Yu et al.~\cite{yu2015solving}	     & 91.8 & 14.2 & 92.7 & 13.0 &  91.7 & 14.2 & 9 \\ \cline{2-9}
805 patches)	& Our method 							 & 86.8 & 21.4 & 90.0 & 14.2 & 89.3 & 15.4 & 9 \\ \cline{2-9}
\hline
Large Pomeranz         & Gallagher~\cite{gallagher2012jigsaw} & 82.9 & 15.6 & 84.2 & 14.2 & 82.8 & 15.7 & 1 \\ \cline{2-9}
(3 images               	  & Yu et al.~\cite{yu2015solving}	     & 89.7 & 12.3 & 90.2 & 11.0 & 89.7 & 12.3 & 1   \\ \cline{2-9}
3300 patches) & Our method 							 & 86.4  & 14.0 & 88.1 & 11.7 & 86.4 & 14.0 & 1 \\ \cline{2-9}
\hline
\end{tabular}
\end{table}

The results presented in \Cref{tab:comp_table} and \Cref{fig:hist_all_dat} indicate that the accuracy of our algorithm is comparable to the state-of-the-art methods when tested on previously suggested datasets.
In general, we noted that when most of the patches have non-zero gradients around their boundaries, our algorithm obtained perfect recovery.
On the other hand, we noted that puzzles with low percentages of recovery by all algorithms have large portions of patches with the same uniform color.
For example, we demonstrate such a puzzle from the MIT dataset, together with its solution by our algorithm, in the first row of \Cref{fig:num_res}.
For this image and all three methods, the neighbors comparison metric was $65\%$, where the errors occurred in the top part of the image of uniform white background. Furthermore, this was the minimal value of this metric among all puzzles in the MIT dataset for all methods.
The solutions obtained by the three algorithms are visually identical to the original one.
On the other hand, the solutions of any of the three methods to the two images with minimal neighbors comparison metric of either the McGill or the Pomeranz datasets (the corresponding two images are the same for all three methods) do not look visually identical to the original images.

\begin{comment}
\begin{table}[]
\caption{Demonstration of the recovery of rotations under different noise levels. We report the mean values of the percentages of correctly recovered rotations over 20 instances for each puzzle.}
\begin{tabular}{|l|l|l|l|l|l|l|l|l|l|l|l|}
\hline
Corruption level                                          & 0.00  & 0.02  &  0.04  & 0.06  & 0.08  & 0.10 & 0.12 & 0.14 & 0.16 & 0.18 & 0.20 \\ \hline
\begin{tabular}[c]{@{}l@{}}MIT (20 images, \\ 432 patches) \end{tabular} & 91.3 & 89.3 & 85.8 & 78.5 & 72.1 & 64.9 & 54.8 & 53.6 & 47.8 & 44.5 & 42.0 \\ \hline
\begin{tabular}[c]{@{}l@{}}McGill (20 images, \\ 540 patches)\end{tabular}  & 88.7 & 85.7 & 82.0 & 76.2 & 70.1 & 67.6 & 59.7 & 49.0 & 47.9 & 45.4 & 42.9 \\ \hline
\end{tabular}
\label{table:noise_analysis}
\end{table}
\end{comment}

We further test the robustness of the three algorithms to corruption. We fix a corruption rate of value 0.02, 0.04, 0.06, 0.08, 0.10, 0.12 or 0.14. We arbitrarily fix a $28 \times 28$ patch and then a side of this patch and change the values of this side (we uniformly sample $28 \times 3$ values from all pixel values of the given image and assign them to this side) with probability that equals the corruption rate. Note that if, for example, the corruption rate is 0.14, then a given edge (which may arise from two different patches) is corrupted with probability $1-0.86^2 \approx 0.26$.

There are several reasons for this model. First, since gradients are sensitive to noise, the MGC metric is sensitive to noise and thus application of standard noise models to the whole image will result in similar degradation of performance by all tested methods. Second, this model of corrupting sides of patches is somewhat similar to adversarially corrupting edges in group synchronization \cite{lerman_shi2019robust, maunu2020provably}, where despite really bad corruption, one has some clean information that can help in solving the synchronization problem. At last, we hope that for images without uniform regions, the conditions of \Cref{theorem:perturbation} may hold under small corruption in this model, but of course we cannot verify this as there is a gap between the clean theory and the applied problem.

\begin{figure}[h]
\begin{center}
\includegraphics[width=0.32\columnwidth]{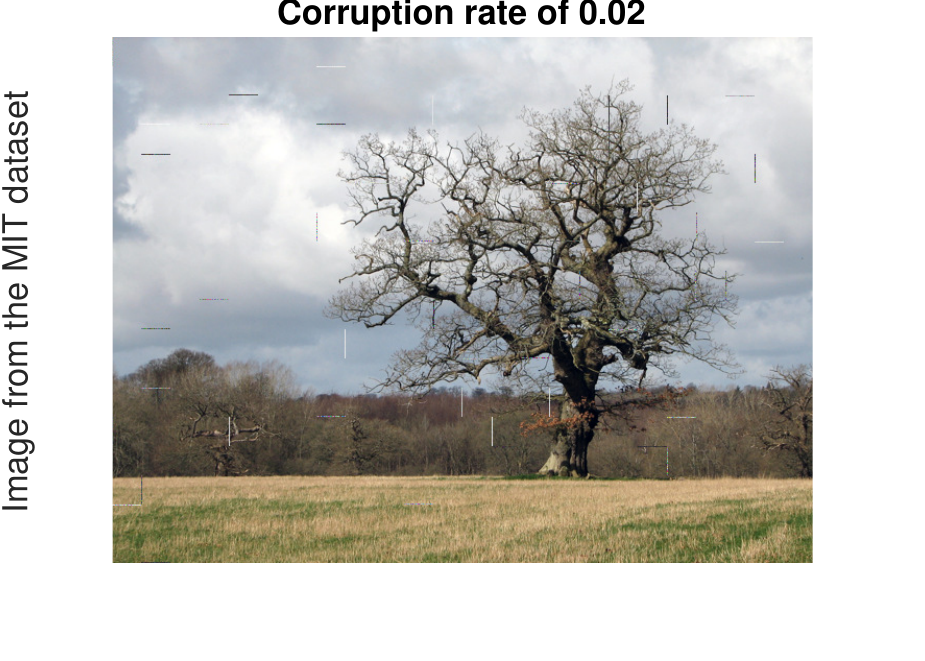}
\includegraphics[width=0.32\columnwidth]{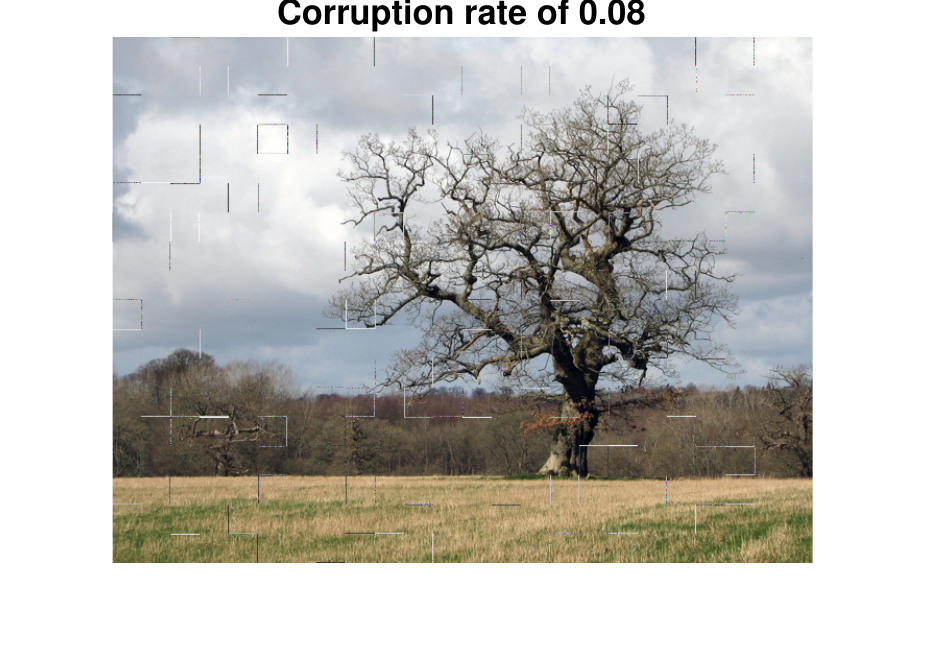}
\includegraphics[width=0.32\columnwidth]{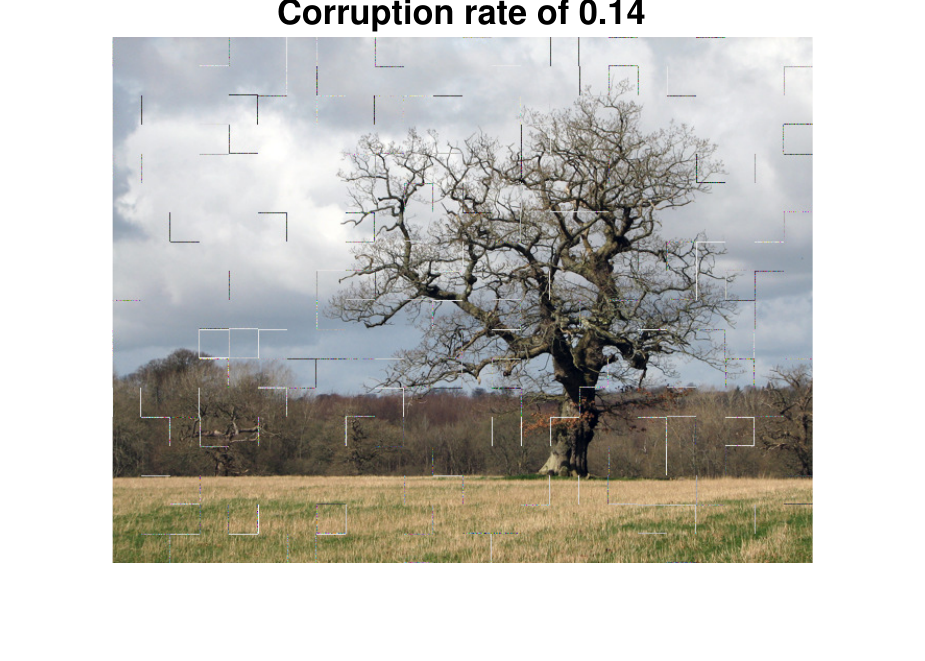}

\includegraphics[width=0.32\columnwidth]{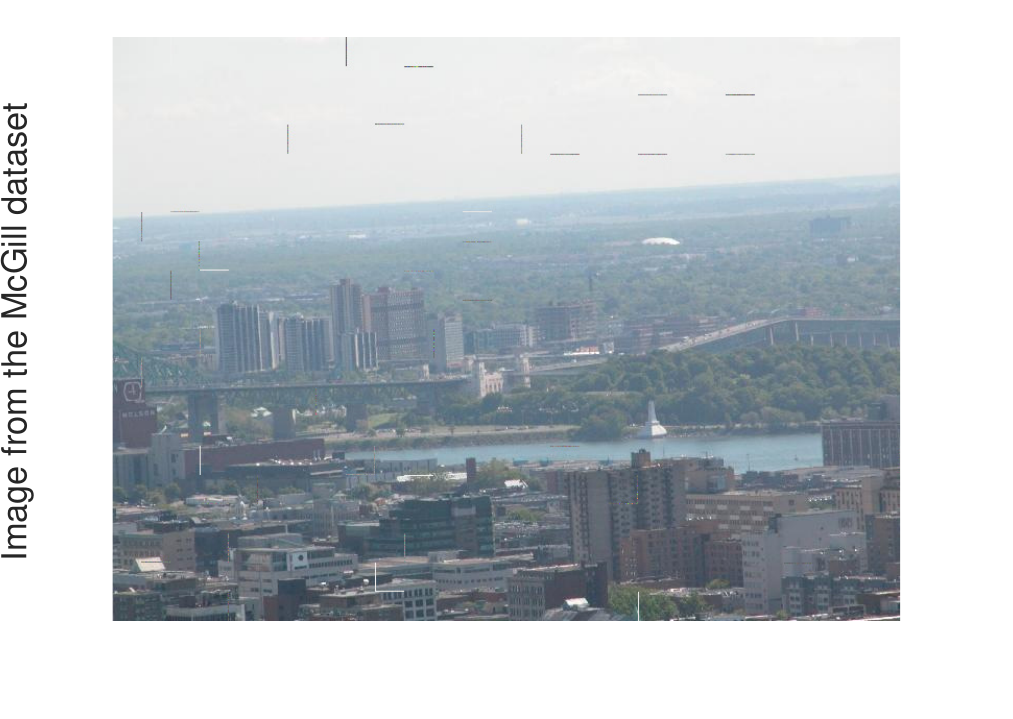}
\includegraphics[width=0.32\columnwidth]{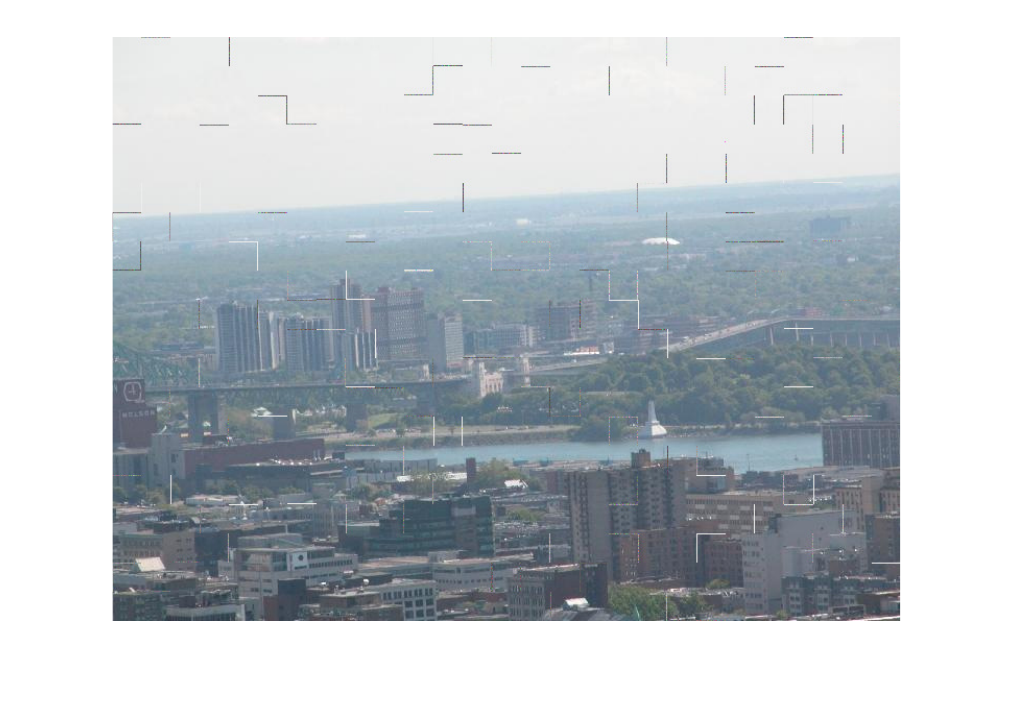}
\includegraphics[width=0.32\columnwidth]{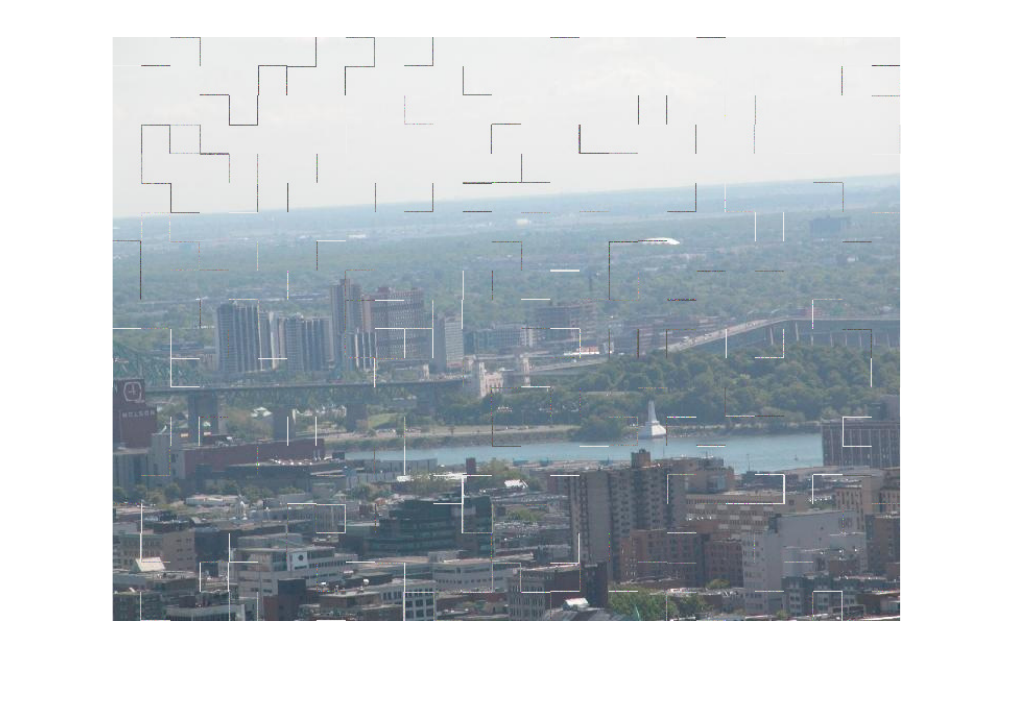}
\end{center}
\caption{Examples of puzzles with corrupted patch sides. The puzzle on the first row is from the MIT dataset and the puzzle on the second row is from the McGill dataset. The three columns correspond to the following corruption rates: $0.02$, $0.08$ and $0.14$.}
\label{fig:corrupted_examples}
\end{figure}

\begin{figure}
\centering     %%% not \center
\includegraphics[width=0.49\columnwidth]{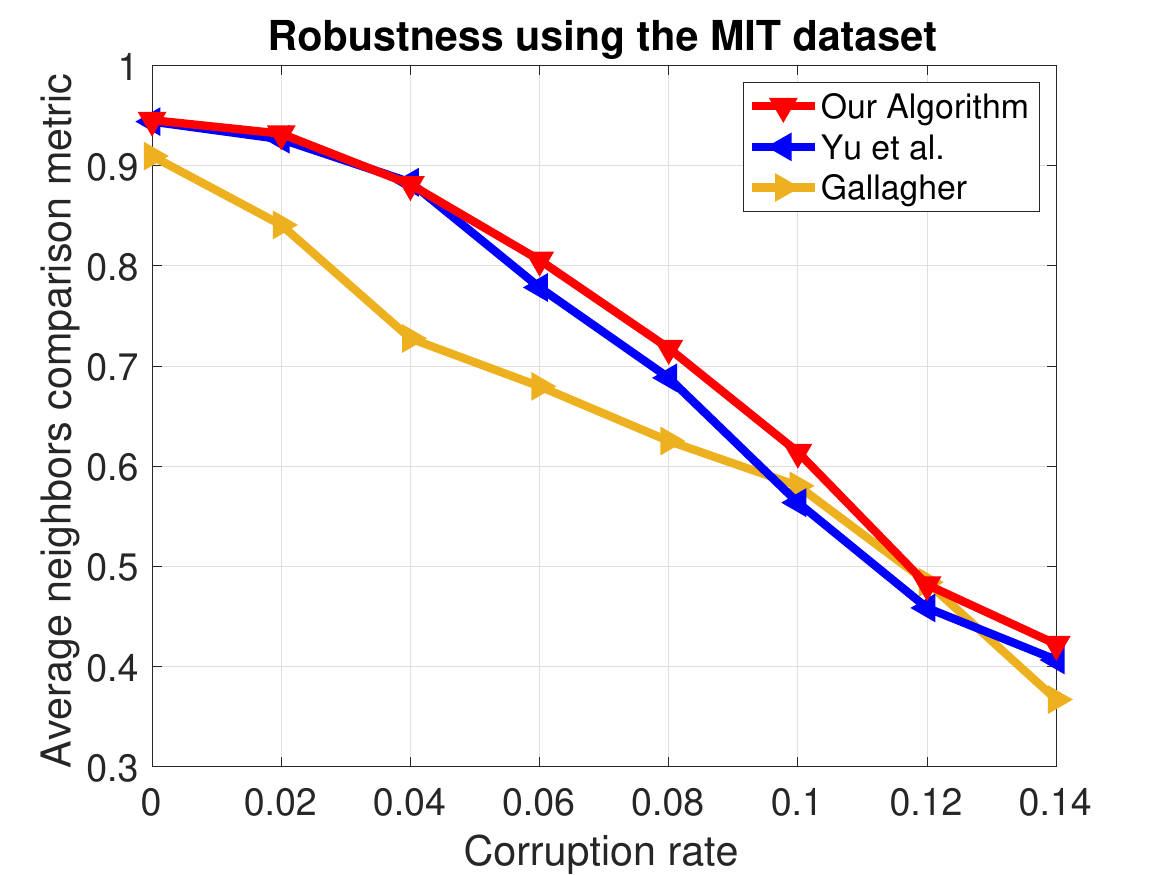}
\includegraphics[width=0.49\columnwidth]{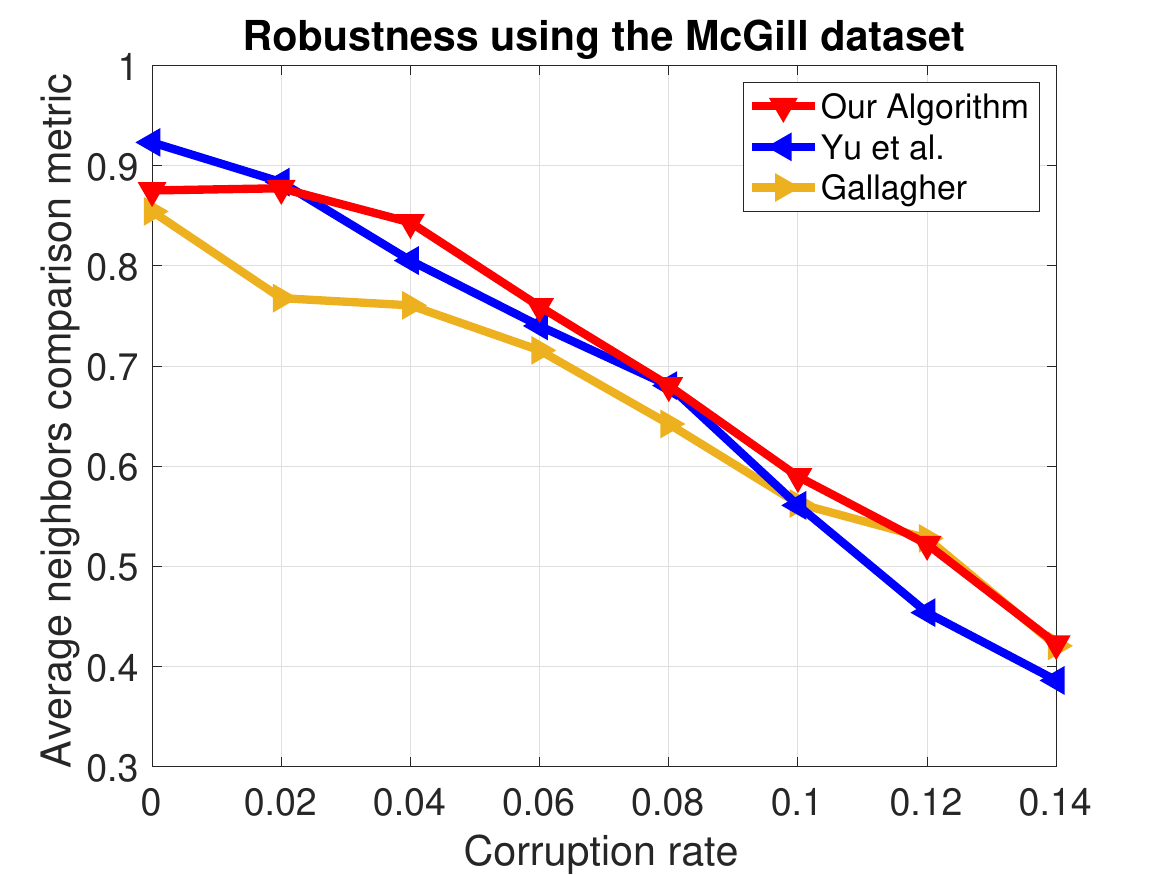}
\caption{Demonstration of the robustness of \Cref{algo:type2_sol} in case of corrupted puzzle sides. We compare our proposed algorithm with the algorithms of Gallagher~\cite{gallagher2012jigsaw} and Yu et al.~\cite{yu2015solving} when the patches are corrupted. The x axis is the corruption rate and the y axis is the average neighbors comparison metric for the puzzles under different noise levels.
%The right column demonstrated the recovery of patch orientations under our proposed algorithm. The x axis is the corruption rate and the y axis is the portion of correctly recovered orientations.
We report the mean values of neighbors comparison metric for 5 instances and 20 puzzles of MIT dataset on the first row and McGill dataset on the second row.}
\label{fig:corruption_test}
\end{figure}

\Cref{fig:corrupted_examples} demonstrates some examples of puzzles with corrupted patches.
%We compute the MGC metric for this puzzle with corrupted edges and run \Cref{algo:rot_sol}.
\Cref{fig:corruption_test} shows graphs of the averaged neighbors comparison metric as a function of the corruption rates computed for puzzles of the MIT dataset and the McGill dataset.
For each dataset, there are three graphs for the three algorithms: \Cref{algo:type2_sol}, Gallagher~\cite{gallagher2012jigsaw} and Yu et al.~\cite{yu2015solving}.
We note that our algorithm obtains the highest averaged accuracy for all nonzero corruption rates. Nevertheless, Yu et al.~\cite{yu2015solving} is still somewhat comparable to our algorithm, however, Gallagher~\cite{gallagher2012jigsaw} is clearly less accurate than both methods.

%On the right column of \Cref{fig:corruption_test} we demonstrate the recovery error of only orientations by SP1 of our proposed algorithm. We remark that the decay of recovery is slower as the corruption rate increases.

\begin{table}[]
\centering
\caption{Comparison of running times for type 2 puzzles for the four datasets.}
\label{tab:running_time}
\begin{tabular}{|l|l|l|}
\hline
Dataset                                                                                                                & Method                                                     & Time (Seconds) \\ \hline
\multirow{4}{*}{\begin{tabular}[c]{@{}l@{}}MIT  \\ (20 images, \\ 432 patches, $28 \times 28$) \end{tabular}}      & MGC metric calculation  &   32.5           \\ \cline{2-3}
                                                                                                                       & Gallagher~\cite{gallagher2012jigsaw}    &        36.4      \\ \cline{2-3}
                                                                                                                       & Yu et al.~\cite{yu2015solving}              &       60.2       \\ \cline{2-3}
                                                                                                                       & Our method with 1 iteration                                  &       38.1       \\ \cline{2-3}
                                                                                                                       & Our method with 5 iterations                                &      57.6        \\ \hline
\multirow{4}{*}{\begin{tabular}[c]{@{}l@{}}McGill \\ (20 images, \\ 540 patches, $28 \times 28$) \end{tabular}}   & MGC metric calculation  &    51.3          \\ \cline{2-3}
                                                                                                                       & Gallagher~\cite{gallagher2012jigsaw}  &    58.5     	\\ \cline{2-3}
                                                                                                                       & Yu et al.~\cite{yu2015solving}     &      100.1        \\ \cline{2-3}
                                                                                                                       & Our method with 1 iteration                                     &       58.5       \\ \cline{2-3}
                                                                                                                       & Our method with 5 iterations                                    &       87.5       \\ \hline
\multirow{4}{*}{\begin{tabular}[c]{@{}l@{}}Pomeranz \\ (20 images, \\ 805 patches, $28 \times 28$) \end{tabular}} &  MGC metric calculation  &   112.3     \\ \cline{2-3}
                                                                                                                       & Gallagher~\cite{gallagher2012jigsaw}  &         135.5   \\ \cline{2-3}
                                                                                                                       & Yu et al.~\cite{yu2015solving}      &        234      \\ \cline{2-3}
                                                                                                                       & Our method with 1 iteration                                     &      128        \\ \cline{2-3}
                                                                                                                       & Our method with 5 iterations                                    &     178         \\ \hline
\multirow{4}{*}{\begin{tabular}[c]{@{}l@{}}Large Pomeranz \\ (3 images, \\ 3300 patches, $28 \times 28$) \end{tabular}} &  MGC metric calculation  &   1908.5       \\ \cline{2-3}
                                                                                                                       & Gallagher~\cite{gallagher2012jigsaw}       &     3288       \\ \cline{2-3}
                                                                                                                       & Yu et al.~\cite{yu2015solving}       &        6120      \\ \cline{2-3}
                                                                                                                       & Our method with 1 iteration                                    &    2214          \\ \cline{2-3}
                                                                                                                       & Our method with 5 iterations                                    &   2857           \\ \hline
\end{tabular}
\end{table}

Next, we compare the running times of Gallagher~\cite{gallagher2012jigsaw}, Yu et al.~\cite{yu2015solving} and our algorithm, where we use a Macbook pro with a 2.3 GHz Intel Core i5 processor and a 8 GB 2133 MHz LPDDR3 memory.
\Cref{tab:running_time} reports results for the four different datasets. It includes results for both the faster implementation of our algorithm with a single iteration (and no update) and our default implementation with 5 iterations described in \Cref{algo:type2_sol}. It also includes the time of calculating the MGC metric, which is shared by all algorithms and is rather slow.
We can see from \Cref{tab:running_time} that Gallagher~\cite{gallagher2012jigsaw} is the fastest algorithm for the smaller puzzles and our algorithms is the fastest for the largest puzzles.  Yu et al.~\cite{yu2015solving} is the slowest algorithm.

Comparing the ratio of the time of each algorithm over the time of computing the MGC metric, one can notice the following: For our single iteration implementation, these ratios are very similar across different puzzle sizes . Thus, the complexity of this implementation seems to match order $O(n^2)$ on these datasets (as the complexity of computing the MGC metric is of order $O(n^2)$). For our full implementation according to \Cref{algo:type2_sol}, these ratios slightly decrease. That is, relatively, less updates are needed for larger puzzles. These ratios slightly increase for Gallagher~\cite{gallagher2012jigsaw} and they significantly increase for Yu et al.~\cite{yu2015solving}. In view of this observation and the discussion in \Cref{sec:time_complex}, it is possible that for typical puzzles the order of complexity of Yu et al.~\cite{yu2015solving} is higher than $O(n^2)$.

We remark that for most images, our algorithm obtains competitive accuracy with either one or two iterations.
However, there are a couple of images for which more iterations are needed to achieve competitive accuracy.

Finally, we would like to mention that most state-of-the-art algorithms, in particular \cite{gallagher2012jigsaw, son2014solving, yu2015solving}, use a greedy step to make final corrections.
We believe that by using that final step of corrections we could further improve our results; however, we would like to avoid greedy procedures.

All codes necessary to duplicate these results are available in \url{https://github.com/vahanhuroyan/PuzzleDemoGCL}.

\section{Conclusion}
\label{sec:discussion}

This paper introduces a novel and constructive mathematical approach for solving square jigsaw puzzles.
It first suggests a procedure for recovering the unknown orientations of type 2 puzzle patches and guarantees its robustness to measurement errors assuming a special clean setting.
Furthermore, it also suggests a principled strategy for updating the full puzzle solution based on the latter strategy for solving orientations.
Some components of the proposed algorithm, in particular, the strategy for recovering orientations, are relatively fast. Nevertheless, the main bottleneck in the computational complexity of our algorithm, that is, calculating the MGC metric, is shared by all existing algorithms.
Numerical experiments on datasets of square jigsaw puzzles indicate that the accuracy of our algorithm is comparable to that of state-of-the-art methods.
Furthermore, on average, our algorithm seems to outperform the existing methods in the presences of corruption. It is also more computationally efficient for large puzzles.

We expect some possible extensions of the proposed algorithm. First of all, we believe that the ideas pursued in this work could be extended to puzzles that come from more complicated manifolds, such as the two-dimensional sphere or a three-dimensional cube jigsaw puzzle, or
puzzles with more complicated shapes of patches, such as tangrams.
The GCL algorithm should be the same; however, instead of considering the group $\ZZ_4$, one needs to consider the corresponding rotation group. Two challenges though are defining a good metric between puzzle pieces and constructing the connection graph.
By doing this, one will extend the applicability of this work to various real-world applications, such as three-dimensional image reconstruction from two-dimensional images.

In terms of theory, it is interesting to analyze our proposed GCL algorithm with more
complicated perturbations.
Additional theoretical questions arise from different ideas discussed in the supplemental material that we cannot make practical.
For example, we are  interested to find out if one can effectively utilize the vector diffusion distances or a modification of them.
Moreover, we would like to know if one can better estimate the locations of the patches by using a quadratic assignment problem formulation.

\section{Acknowledgement}

We would like to thank Rui Yu for kindly sending us his code of the algorithm presented in \cite{yu2015solving} and to Andrew Gallagher for posting the code used in \cite{gallagher2012jigsaw}.
We are very thankful for the anonymous reviewers for the very careful reading of the manuscript and their valuable comments. We are also thankful to Dr.~Brendt Wohlberg for his professional handling of the manuscript.

\bibliographystyle{siamplain}
\bibliography{references}

\begin{thebibliography}{10}

\bibitem{altman1989solving}
{\sc T.~Altman}, {\em Solving the {JIGSAW} puzzle problem in linear time},
  Applied Artificial Intelligence, 3 (1989), pp.~453--462,
  \url{http://dx.doi.org/10.1080/08839518908949937}.

\bibitem{andalo2012solving}
{\sc F.~A. Andal{\'o}, G.~Taubin, and S.~Goldenstein}, {\em Solving image
  puzzles with a simple quadratic programming formulation}, in Graphics,
  Patterns and Images (SIBGRAPI), 2012 25th SIBGRAPI Conference on, IEEE, 2012,
  pp.~63--70.

\bibitem{bandeira2013cheeger}
{\sc A.~S. Bandeira, A.~Singer, and D.~A. Spielman}, {\em A cheeger inequality
  for the graph connection {L}aplacian}, SIAM Journal on Matrix Analysis and
  Applications, 34 (2013), pp.~1611--1630.

\bibitem{bordenave2016shotgun}
{\sc C.~Bordenave, U.~Feige, and E.~Mossel}, {\em Shotgun assembly of random
  jigsaw puzzles}, CoRR, abs/1605.03086 (2016),
  \url{http://arxiv.org/abs/1605.03086}.

\bibitem{brown2008system}
{\sc B.~J. Brown, C.~Toler-Franklin, D.~Nehab, M.~Burns, D.~P. Dobkin,
  A.~Vlachopoulos, C.~Doumas, S.~Rusinkiewicz, and T.~Weyrich}, {\em A system
  for high-volume acquisition and matching of fresco fragments: reassembling
  {T}heran wall paintings}, {ACM} Trans. Graph., 27 (2008), pp.~84:1--84:9.

\bibitem{chen2018new}
{\sc L.~Chen, D.~Cao, and Y.~Liu}, {\em A new intelligent jigsaw puzzle
  algorithm base on mixed similarity and symbol matrix}, International Journal
  of Pattern Recognition and Artificial Intelligence, 32 (2018), p.~1859001.

\bibitem{cho2010probabilistic}
{\sc T.~S. Cho, S.~Avidan, and W.~T. Freeman}, {\em A probabilistic image
  jigsaw puzzle solver}, in Computer Vision and Pattern Recognition (CVPR),
  2010 IEEE Conference on, IEEE, 2010, pp.~183--190.

\bibitem{cormen2009introduction}
{\sc T.~H. Cormen, C.~E. Leiserson, R.~L. Rivest, and C.~Stein}, {\em
  Introduction to Algorithms, 3rd Edition}, {MIT} Press, 2009,
  \url{http://mitpress.mit.edu/books/introduction-algorithms}.

\bibitem{davis1970rotation}
{\sc C.~Davis and W.~M. Kahan}, {\em The rotation of eigenvectors by a
  perturbation. iii}, SIAM Journal on Numerical Analysis, 7 (1970), pp.~1--46.

\bibitem{deever2012semi}
{\sc A.~Deever and A.~Gallagher}, {\em Semi-automatic assembly of real
  cross-cut shredded documents}, in Image Processing (ICIP), 2012 19th IEEE
  International Conference on, IEEE, 2012, pp.~233--236.

\bibitem{demaine07jigsaw}
{\sc E.~D. Demaine and M.~L. Demaine}, {\em Jigsaw puzzles, edge matching, and
  polyomino packing: Connections and complexity}, Graphs and Combinatorics, 23
  (2007), pp.~195--208.

\bibitem{el2016graph}
{\sc N.~El~Karoui and H.-T. Wu}, {\em Graph connection {L}aplacian methods can
  be made robust to noise}, The Annals of Statistics, 44 (2016), pp.~346--372.

\bibitem{fan2017linfinity}
{\sc J.~Fan, W.~Wang, and Y.~Zhong}, {\em An {$l_{\infty}$} eigenvector
  perturbation bound and its application}, J. Mach. Learn. Res., 18 (2017),
  pp.~207:1--207:42, \url{http://jmlr.org/papers/v18/16-140.html}.

\bibitem{fredman1987fibonacci}
{\sc M.~L. Fredman and R.~E. Tarjan}, {\em Fibonacci heaps and their uses in
  improved network optimization algorithms}, J. {ACM}, 34 (1987), pp.~596--615,
  \url{https://doi.org/10.1145/28869.28874},
  \url{https://doi.org/10.1145/28869.28874}.

\bibitem{freeman1964apictorial}
{\sc H.~Freeman and L.~Garder}, {\em Apictorial jigsaw puzzles: The computer
  solution of a problem in pattern recognition}, {IEEE} Transactions on
  Electronic Computers, 13 (1964), pp.~118--127.

\bibitem{gallagher2012jigsaw}
{\sc A.~C. Gallagher}, {\em Jigsaw puzzles with pieces of unknown orientation},
  in 2012 {IEEE} Conference on Computer Vision and Pattern Recognition,
  Providence, RI, USA, June 16-21, 2012, 2012, pp.~382--389,
  \url{http://dx.doi.org/10.1109/CVPR.2012.6247699}.

\bibitem{goldberg2005global}
{\sc D.~Goldberg, C.~Malon, and M.~Bern}, {\em A global approach to automatic
  solution of jigsaw puzzles}, Computational Geometry, 28 (2004), pp.~165 --
  174.

\bibitem{grim2016automatic}
{\sc A.~Grim, T.~O’Connor, P.~J. Olver, C.~Shakiban, R.~Slechta, and
  R.~Thompson}, {\em Automatic reassembly of three-dimensional jigsaw puzzles},
  International Journal of Image and Graphics, 16 (2016), p.~1650009.

\bibitem{hoff2014automatic}
{\sc D.~J. Hoff and P.~J. Olver}, {\em Automatic solution of jigsaw puzzles},
  Journal of Mathematical Imaging and Vision, 49 (2014), pp.~234--250,
  \url{http://dx.doi.org/10.1007/s10851-013-0454-3}.

\bibitem{jaccard1901etude}
{\sc P.~Jaccard}, {\em \'{E}tude comparative de la distribution florale dans
  une portion des alpes et des jura}, Bulletin del la Soci\'{e}t\'{e} Vaudoise
  des Sciences Naturelles, 37 (1901), pp.~547--579.

\bibitem{jin2014jigsaw}
{\sc S.-Y. Jin, S.~Lee, N.~A. Azis, and H.-J. Choi}, {\em Jigsaw puzzle image
  retrieval via pairwise compatibility measurement}, in Big Data and Smart
  Computing (BIGCOMP), 2014 International Conference on, IEEE, 2014,
  pp.~123--127.

\bibitem{justino2006reconstructing}
{\sc E.~Justino, L.~S. Oliveira, and C.~Freitas}, {\em Reconstructing shredded
  documents through feature matching}, Forensic science international, 160
  (2006), pp.~140--147.

\bibitem{koller2006computer}
{\sc D.~Koller and M.~Levoy}, {\em Computer-aided reconstruction and new
  matches in the forma urbis romae}, Bullettino Della Commissione Archeologica
  Comunale di Roma,  (2006), pp.~103–--125.

\bibitem{kovalsky2015global}
{\sc S.~Z. Kovalsky, D.~Glasner, and R.~Basri}, {\em A global approach for
  solving edge-matching puzzles}, {SIAM} J. Imaging Sciences, 8 (2015),
  pp.~916--938, \url{https://doi.org/10.1137/140987869}.

\bibitem{lerman_shi2019robust}
{\sc G.~Lerman and Y.~Shi}, {\em Robust group synchronization via cycle-edge
  message passing}, 2019, \url{https://arxiv.org/abs/1912.11347}.

\bibitem{liu2011automated}
{\sc H.~Liu, S.~Cao, and S.~Yan}, {\em Automated assembly of shredded pieces
  from multiple photos}, IEEE Transactions on Multimedia, 13 (2011),
  pp.~1154--1162.

\bibitem{makridis2006new}
{\sc M.~Makridis and N.~Papamarkos}, {\em A new technique for solving a jigsaw
  puzzle}, in Image Processing, 2006 IEEE International Conference on, IEEE,
  2006, pp.~2001--2004.

\bibitem{marande2007mitochondrial}
{\sc W.~Marande and G.~Burger}, {\em Mitochondrial dna as a genomic jigsaw
  puzzle}, Science, 318 (2007), pp.~415--415.

\bibitem{marques2009reconstructing}
{\sc M.~A. Marques and C.~O. Freitas}, {\em Reconstructing strip-shredded
  documents using color as feature matching}, in Proceedings of the 2009 ACM
  symposium on Applied Computing, ACM, 2009, pp.~893--894.

\bibitem{martinsson2016shotgun}
{\sc A.~Martinsson}, {\em Shotgun edge assembly of random jigsaw puzzles},
  CoRR, abs/1605.07151 (2016), \url{http://arxiv.org/abs/1605.07151}.

\bibitem{maunu2020provably}
{\sc T.~Maunu and G.~Lerman}, {\em A provably robust multiple rotation
  averaging scheme for {SO(2)}}, 2020, \url{https://arxiv.org/abs/2002.05299}.

\bibitem{mondal2013robust}
{\sc D.~Mondal, Y.~Wang, and S.~Durocher}, {\em Robust solvers for square
  jigsaw puzzles}, in 2013 International Conference on Computer and Robot
  Vision, IEEE, 2013, pp.~249--256.

\bibitem{mossel2018shotgun}
{\sc E.~Mossel and N.~Ross}, {\em Shotgun assembly of labeled graphs}, IEEE
  Transactions on Network Science and Engineering,  (2018), pp.~1--1,
  \url{https://doi.org/10.1109/TNSE.2017.2776913}.

\bibitem{nielsen2008solving}
{\sc T.~R. Nielsen, P.~Drewsen, and K.~Hansen}, {\em Solving jigsaw puzzles
  using image features}, Pattern Recognition Letters, 29 (2008),
  pp.~1924--1933.

\bibitem{oxholm2013flexible}
{\sc G.~Oxholm and K.~Nishino}, {\em A flexible approach to reassembling thin
  artifacts of unknown geometry}, Journal of cultural heritage, 14 (2013),
  pp.~51--61.

\bibitem{paikin2015solving}
{\sc G.~Paikin and A.~Tal}, {\em Solving multiple square jigsaw puzzles with
  missing pieces}, in Proceedings of the IEEE Conference on Computer Vision and
  Pattern Recognition, 2015, pp.~4832--4839.

\bibitem{pintus2016survey}
{\sc R.~Pintus, K.~Pal, Y.~Yang, T.~Weyrich, E.~Gobbetti, and H.~Rushmeier},
  {\em A survey of geometric analysis in cultural heritage}, Computer Graphics
  Forum, 35 (2015), pp.~4--31.

\bibitem{pomeranz2011fully}
{\sc D.~Pomeranz, M.~Shemesh, and O.~Ben{-}Shahar}, {\em A fully automated
  greedy square jigsaw puzzle solver}, in The 24th {IEEE} Conference on
  Computer Vision and Pattern Recognition, {CVPR} 2011, Colorado Springs, CO,
  USA, 20-25 June 2011, 2011, pp.~9--16,
  \url{http://dx.doi.org/10.1109/CVPR.2011.5995331}.

\bibitem{sholomon2014generalized}
{\sc D.~Sholomon, O.~E. David, and N.~S. Netanyahu}, {\em A generalized genetic
  algorithm-based solver for very large jigsaw puzzles of complex types}, in
  Proceedings of the Twenty-Eighth {AAAI} Conference on Artificial
  Intelligence, July 27 -31, 2014, Qu{\'{e}}bec City, Qu{\'{e}}bec, Canada.,
  2014, pp.~2839--2845,
  \url{http://www.aaai.org/ocs/index.php/AAAI/AAAI14/paper/view/8650}.

\bibitem{sholomon2014genetic}
{\sc D.~Sholomon, O.~E. David, and N.~S. Netanyahu}, {\em Genetic
  algorithm-based solver for very large multiple jigsaw puzzles of unknown
  dimensions and piece orientation}, in Proceedings of the 2014 Annual
  Conference on Genetic and Evolutionary Computation, GECCO '14, New York, NY,
  USA, 2014, ACM, pp.~1191--1198.

\bibitem{sholomon2016automatic}
{\sc D.~Sholomon, O.~E. David, and N.~S. Netanyahu}, {\em An automatic solver
  for very large jigsaw puzzles using genetic algorithms}, Genetic Programming
  and Evolvable Machines, 17 (2016), pp.~291--313,
  \url{http://dx.doi.org/10.1007/s10710-015-9258-0}.

\bibitem{sholomon2016dnn}
{\sc D.~Sholomon, O.~E. David, and N.~S. Netanyahu}, {\em Dnn-buddies: A deep
  neural network-based estimation metric for the jigsaw puzzle problem}, in
  Artificial Neural Networks and Machine Learning -- ICANN 2016, A.~E. Villa,
  P.~Masulli, and A.~J. Pons~Rivero, eds., 2016, pp.~170--178.

\bibitem{singer2011angular}
{\sc A.~Singer}, {\em Angular synchronization by eigenvectors and semidefinite
  programming}, Applied and computational harmonic analysis, 30 (2011),
  pp.~20--36.

\bibitem{singer2012vector}
{\sc A.~Singer and H.-T. Wu}, {\em Vector diffusion maps and the connection
  {L}aplacian}, Communications on pure and applied mathematics, 65 (2012),
  pp.~1067--1144.

\bibitem{sizikova2017wall}
{\sc E.~Sizikova and T.~Funkhouser}, {\em Wall painting reconstruction using a
  genetic algorithm}, Journal on Computing and Cultural Heritage (JOCCH), 11
  (2017), p.~3.

\bibitem{son2014solving}
{\sc K.~Son, J.~Hays, and D.~B. Cooper}, {\em Solving square jigsaw puzzles
  with loop constraints}, in European Conference on Computer Vision, Springer,
  2014, pp.~32--46.

\bibitem{son2016solving}
{\sc K.~Son, D.~Moreno, J.~Hays, and D.~B. Cooper}, {\em Solving small-piece
  jigsaw puzzles by growing consensus}, in Proceedings of the IEEE Conference
  on Computer Vision and Pattern Recognition, 2016, pp.~1193--1201.

\bibitem{toler2010multi}
{\sc C.~Toler-Franklin, B.~Brown, T.~Weyrich, T.~Funkhouser, and
  S.~Rusinkiewicz}, {\em Multi-feature matching of fresco fragments}, ACM
  Trans. on Graphics (Proc. SIGGRAPH Asia), 29 (2010), pp.~185:1--185:11.

\bibitem{vaidya1989speeding}
{\sc P.~M. Vaidya}, {\em Speeding-up linear programming using fast matrix
  multiplication (extended abstract)}, in 30th Annual Symposium on Foundations
  of Computer Science, Research Triangle Park, North Carolina, USA, 30 October
  - 1 November 1989, {IEEE} Computer Society, 1989, pp.~332--337,
  \url{https://doi.org/10.1109/SFCS.1989.63499},
  \url{https://doi.org/10.1109/SFCS.1989.63499}.

\bibitem{yang2011particle}
{\sc X.~Yang, N.~Adluru, and L.~J. Latecki}, {\em Particle filter with state
  permutations for solving image jigsaw puzzles}, in Computer Vision and
  Pattern Recognition (CVPR), 2011 IEEE Conference on, IEEE, 2011,
  pp.~2873--2880.

\bibitem{yao2003shape}
{\sc F.-H. Yao and G.-F. Shao}, {\em A shape and image merging technique to
  solve jigsaw puzzles}, Pattern Recognition Letters, 24 (2003),
  pp.~1819--1835.

\bibitem{yu2015solving}
{\sc R.~Yu, C.~Russell, and L.~Agapito}, {\em Solving jigsaw puzzles with
  linear programming}, in Proceedings of the British Machine Vision Conference
  2016, {BMVC} 2016, York, UK, September 19-22, 2016, 2016.

\bibitem{yu2015useful}
{\sc Y.~Yu, T.~Wang, and R.~J. Samworth}, {\em A useful variant of the
  {D}avis--{K}ahan theorem for statisticians}, Biometrika, 102 (2015),
  pp.~315--323.

\bibitem{zhao2007puzzle}
{\sc Z.~Yu-Xiang, S.~Mu-Chun, C.~Zhong-Lie, and L.~Jonathan}, {\em A puzzle
  solver and its application in speech descrambling}, in Proceedings of the
  2007 WSEAS International Conference on Computer Engineering and Applications,
  Gold Coast, Australia, January 17-19, 2007, 2007, pp.~171--176.

\bibitem{zhang2006complex}
{\sc S.~Zhang and Y.~Huang}, {\em Complex quadratic optimization and
  semidefinite programming}, SIAM Journal on Optimization, 16 (2006),
  pp.~871--890, \url{https://doi.org/10.1137/04061341X}.

\end{thebibliography}

\newpage

\clearpage
\pagenumbering{arabic}% resets `page` counter to 1
\renewcommand*{\thepage}{A\arabic{page}}

\section*{Supplemental Material}
~
\appendix

\section{A General Mathematical Formulation}
\label{subsec:math_general}

Our general mathematical formulation assumes a $d$-dimensional compact manifold $M$ embedded in $\RR^q$ via the inclusion map $\iota$.
For simplicity, we refer to the embedded manifold by $M$ instead of $\iota(M)$. For this embedded $M$, we further consider a sufficiently smooth function $f: M\to \RR^k$, where $k \geq 1$. The required smoothness of $f$ depends on the application domain. For the application we consider, which has a discrete setting with discontinuities of $f$, the assumption $f\in L^2(M,\RR^k)$ seems natural. Note that $M$ serves as the ``physical space'' and $f$ as an ``image'' defined on this space, where $k>1$ can correspond to a multi-spectral image.

We will first discuss the notion of patches partitioning the embedded $M$ as well as image patches.
Generally, a patch is a subset of the embedded $M$. %We only worry about the topological shape of the patch and not on the actual embedding of it within $\RR^q$.
Since our mathematical setting is continuous, we assume that patches are open sets. We later explain how this assumption does not matter to the discrete setting of this paper.
An image patch on the embedded $M$ is a pair of a patch and the restriction of $f$ on it. For simplicity, we denote a patch by $P$, even though $\iota(P)$ is more precise.
%Similarly, we denote an image patch by $(P, f|_P)$, even though $(\iota(P), f|_{\iota(P)})$ is more precise.

We partition $M$ into open patches $\{P_i\}_{i=1}^n$ so that
$M=\cup_{i=1}^n\bar{P}_i$,
where for $1 \leq i \leq n$, $P_i\subset M$ and $\bar{P}_i$ is the closure of $P_i$, and also for $1 \leq i\neq j \leq n$, $P_i \cap P_j = \emptyset.$
When defining the corresponding image patches we allow local rigid transformations, such as rotations and translations. We make the problem formulation even more general by considering local diffeomorphic transformations.
For each $1 \leq i \leq n$, consider a transform $D_i$ on $\RR^q$, so that $D_i(P_i)\subset \RR^q$ is diffeomorphic to $P_i$. For $\vx \in \RR^q$, define $(D_i \circ f|_{P_i})(\vx):=f(D_i^{-1}(\vx))$ when $D_i^{-1}(\vx)\in P_i$ and $\vzero$ otherwise.

There are three jigsaw puzzle problems we can formulate:
\begin{enumerate}
\item[P0:] Given a set of image patches $\cQ := \{(D_i(P_i),D_i\circ f|_{P_i})\}_{i=1}^n$ and $M$, recover $f.$
\item[P1:] Given a set of image patches $\cQ := \{(D_i(P_i),D_i\circ f|_{P_i})\}_{i=1}^n,$ recover $f$ and $M.$
\item[P2:] Given a set of patches $\cP:=\{D_i(P_i)\}_{i=1}^n$, recover $M.$
\end{enumerate}
In general these are ill-defined and challenging problems, since more conditions may be needed. For example, if $f$ is a constant function on a sufficiently large region of $M$ and the shapes of the puzzle patches are not sufficient to uniquely determine neighboring patches, then there is no information available for reconstructing $f$. Similarly, estimating the unknown local diffeomorphic functions is a challenging problem, and it makes sense to further restrict them.
On the other hand, there are simplified, well-defined versions of these problems.

A very special case of P0 is the square jigsaw puzzle problem.
One may similarly consider more complicated shapes of patches, such as polygonal shapes, which are common in tangram puzzles, or shapes with curvy sides, which are common in commercial jigsaw puzzles. Mathematical ideas for solving these two  kinds of puzzles appear in \cite{kovalsky2015global} and \cite{hoff2014automatic}, respectively.
In the more general case, one needs to consider arbitrary shifts and rotations in $\RR^2$. We remark that such consideration in the square jigsaw puzzle is equivalent to the restriction of rotations to $\ZZ_4$ and shifts to permutations of patches. We also remark that there are cases of more complicated shapes that are easier to solve.
For example, if the shapes of the patches lead to unique determination of the neighboring patches, then exact reconstruction is easier.
On the other hand, there are clearly very difficult cases of complicated shapes with many possibilities of aligning them together. In general, one may also consider various 3D puzzles or more complicated problems. Note that most of the ideas discussed in this paper can be well suited for puzzles with non-square patches and a higher-dimensional non-flat manifold.

Examples of Problems P1 and P2 with different physical spaces of different dimensions appear in \cite{goldberg2005global, hoff2014automatic, grim2016automatic, yao2003shape, makridis2006new, nielsen2008solving, oxholm2013flexible}. Note that in these works, the patches may have different shapes.
In general, these problems may be ill-defined or not unique. For example, if one reformulates the square jigsaw puzzle into a P1 problem and asks to find the rectangle $M$, then the solution is unique up to a proper rigid transformation. Furthermore, P2 is ill-defined for the setting of the square jigsaw puzzle as it has many possible solutions. In general, a solution of P2 requires stronger assumptions, for example, on the shape of puzzle patches or on the manifold that may need to be closed.

\subsection{Another Formulation for Solving $\ZZ_4$ Synchronization}
%\subsection{An Analogous Algorithm by The Synchronization Problem}
\label{sec:sync}

We describe here a least-squares formulation for $\ZZ_4$ synchronization, review two common solutions for it and discuss the similarities and differences of one such solution with the method above.
We note that the solution is a vector of $n$ elements in $\ZZ_4$, that is, a vector in $\ZZ_4^n$, which we represent as a block vector in $\RR^{2n \times 2}$ (following \eqref{eq:represent_rotations}).
Using the $2n \times 2n$ matrix $\mS$ defined in \eqref{eq:vdmS}, the least-squares formulation for $\ZZ_4$ synchronization asks to solve the optimization problem
\begin{equation}
\label{eq:sync_1}
\argmin_{\vu \in \ZZ_4^n} \left \Vert \vu \vu^T - \mS \right\Vert_F^2,
\end{equation}
Since for $\vu \in \ZZ_4^n$, $\Vert \vu \vu^T - \mS \Vert_F^2 = 4 n^2 + \Vert \mS \Vert_F^2 - 2 \tr(\vu \vu^T \mS)$, \eqref{eq:sync_1} is equivalent with
\begin{equation}
\label{eq:sync_2}
\argmax_{\vu \in \ZZ_4^{n}} \tr\left(\vu \vu^T \mS \right).
\end{equation}

This problem is NP-hard~\cite{zhang2006complex}. Nevertheless, approximate solutions were proposed, in particular, semidefinite programming and spectral relaxation~\cite{singer2011angular}.
The semidefinite programming method suggests to remove the rank $2$ constraint on the PSD (positive semi-definite) matrix $\vu \vu^T$ in \eqref{eq:sync_2} and consequently solve
\begin{equation}
\label{eq:sync_3}
\argmax_{\mH \succeq \vzero, \mH(i, i) = \mI_2} \tr(\mH \mS).
\end{equation}
An approximation to the solution of \eqref{eq:sync_2} is obtained by projecting onto $\ZZ_4$ each $2 \times 2$ block of the ${2n \times 2}$ block matrix obtained by concatenating the top $2$ eigenvectors of \eqref{eq:sync_3}.

The spectral relaxation method suggests to relax the set $\ZZ_4^n$ into $\RR^{2n \times 2}$ and solve the following eigenvalue/eigenvector problem
\begin{equation}
\label{eq:sync_relaxed}
\argmax_{\boldsymbol{u}\in\RR^{2n \times 2}:\,\|\boldsymbol{u}\| = 2n} \tr(\boldsymbol{u}^T \mS \boldsymbol{u}).
\end{equation}
Note that the solution of \eqref{eq:sync_relaxed} is the ${2n \times 2}$ block matrix obtained by concatenating the top $2$ eigenvectors of $\mS$.
Projecting each $2 \times 2$ block of this matrix onto $\ZZ_4$ yields an approximation to the solution of \eqref{eq:sync_2}.
The spectral relaxation is faster and better suited for higher-volume data and we thus apply it in our work.
We remark though that the SDP relaxation is often more accurate than the spectral relaxation for $\SO(2)$.
For the special case of $\ZZ_4$, spectral relaxation might be a sufficiently good approximation. Indeed, since $\ZZ_4$ contains only four, well-separated elements, the projection of the relaxed solution onto them can recover them when the errors are sufficiently small.
We later propose in \S\ref{sec:GCL} a spectral relaxation of \eqref{eq:sync_1} with $\mS$ replaced by $\mC$.
An advantage of using $\mC$ instead of $\mS$ is that it gives rise to a natural diffusion distance, which is discussed later in \S\ref{sec:VDD_est}.

\section{Optional Steps for The Proposed Solution of Type 2 Puzzles}
\label{sec:appen_2}

We describe here optional steps to improve the algorithm. Implementation of these ideas did not obtain the desired improvement, but we believe that they might be interesting and useful for future explorations.
In  \S\ref{sec:VDD_est} we review the vector diffusion map and distance and explain how to use them for updating the MGC metric after applying \Cref{algo:aff_gr}. We also note that these distances are only informative for sufficiently far away vertices and not for nearby vertices.
In \S\ref{sec:qap-supp} we discuss the problem of recovering the location of patches after updating the MGC metric in \S\ref{sec:VDD_est}. We propose a mathematical idea for solving the problem by applying a quadratic assignment formulation with respect to the affinity function $\mW$ defined in \S\ref{sec:conn_graph_const_Type2}.  However, the solver of the combinatorial optimization problem is not sufficiently fast and accurate.
Lastly, \S\ref{sec:appen_4} suggests using other top eigenvectors of the  GCW matrix.
%We thus recommend applying some other existing algorithms for finding locations of patches.

\subsection{Updating the Metric between Puzzle Pieces by Vector Diffusion Distances}
\label{sec:VDD_est}

The MGC metric defined in \S\ref{sec:mgc} is usually not a perfect metric, but it provides some information whether two patches are neighbors or not.  However, if two patches are not neighbors,
the MGC metric between them does not provide any information about their distance in the image. Such information can be helpful since the estimated information on neighboring patches can be wrong.
For this purpose, we suggest updating the MGC metric by considering the diffusion process associated with the random walk determined by $\mC$. The diffusion vector framework for doing this was suggested in \cite{singer2012vector}.
This part is performed after the rotations of the patches were estimated according to \Cref{algo:aff_gr}.

\begin{figure}
\centering     %%% not \center
%\subfigure{\label{fig:vdd_example}
\includegraphics[width=0.5\columnwidth]{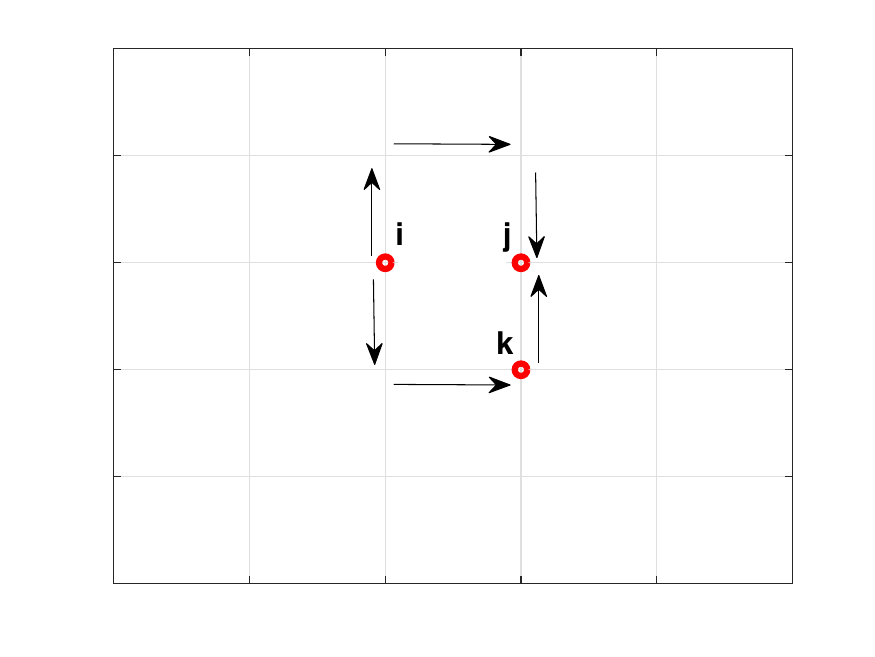}
\caption{An example where VDD fails to reflect the distance between nearby patches.
The graph is a grid with one missing edge between vertices $i$ and $j$ (all other neighboring edges are connected by an edge).
Due to the structure of the grid, the shortest path between vertices $i$ and $j$ is of length $3$, whereas the shortest path
between vertices $i$ and $k$ is $2$.
Thus the use of VDD leads to the wrong conclusion that vertex $i$ is closer to vertex $k$ than to vertex $j$.
}
\label{fig:vdd_example}
\end{figure}

For $t>0$, the {\it vector diffusion map} (VDM)\cite{singer2012vector} in our setting is a function $V_{t,n}: \cQ \to \RR^{2n \times 2n}$ defined by
\begin{equation*}
V_{t,n} : P_i \mapsto \left( (\mu_{\mC, l} \mu_{\mC, r})^{t} \langle v_{\mC, l}[i], v_{\mC, r}[i] \rangle\right)_{l,r=1}^{2n}\in \mathbb{R}^{(2n)^2},
\end{equation*}
where $\mu_{\mC, l}$ and $v_{\mC, l}$ are the $l$-th eigenvalue and eigenvector, respectively, of $\mC$, and $v_{\mC, l}[i]$ is a two-dimensional vector containing the $(2(i-1)+1)$-th and $(2i)$-th entries of $v_{\mC, l}$.
%With this map, the Hilbert-Schmidt norm of the $(i,j)$-th block of $\widetilde{\mC} = \mD^{-1/2}\mS\mD^{-1/2}$ satisfies
%\begin{equation*}
%\|\widetilde{\mC}^{2t}(i,j)\|^2_{HS} = \langle V_{t,n}(P_i), V_{t,n}(P_j) \rangle,
%\end{equation*}
%where $\langle \cdot, \cdot \rangle$ denotes an inner product in  $\RR^{2n \times 2n}$.
The  {\it vector diffusion distance} (VDD)  at time $t>0$ \cite{singer2012vector}  between two patches indexed by $i$ and $j$ is
\begin{equation}
\label{eq:vdd_calc}
d_{\mC, t, n}(i,j) := \| V_{t, n}(P_i)- V_{t, n}(P_j)\|_{\mathbb{R}^{(2n)^2}}.
\end{equation}
This distance converts the local information into global information and provides an estimate of the distance between patches in the original grid.
Based on this distance one can infer whether two patches are close to each other in the original image or far away.
As demonstrated in \Cref{fig:vdd_example}, this distance is not sufficiently accurate to infer nearness when patches have comparable distances. Specifically, a problem arises when two neighboring patches, represented by $i$ and $j$ in  \Cref{fig:vdd_example}, were not estimated to be neighbors by any metric, and therefore the only paths connecting them are through their neighbors. In this case, the diffusion distance between them, in particular, between $i$ and $j$ in \Cref{fig:vdd_example}, is larger than that between one of them and its diagonal neighbor, for example, between $i$ and $k$ in \Cref{fig:vdd_example}. Therefore, this distance does not completely reflect the true underlying geometry. As a result, in general it cannot be used to infer whether two patches are neighbors or diagonal neighbors. We remark that this is due to the discrepancy between the metric we design and the true underlying metric.

Nevertheless, it is still possible to use the VDD for improving the MGC metric in the following way. We first compute the VDD between all patches. For each image patch $P_i \in \cQ$,
we sort the VDD distances of all other patches to $P_i$ and record the patches in the 0.1-quantile of largest distances.
The MGC metric between these patches and $P_i$ is then increased by the factor $\alpha =2$. This ensures that patches that are not likely neighbors
of $P_i$ are penalized by a larger distance and thus have a smaller chance of becoming neighbors in the final solution.
Formally, the modified algorithm for type 2 puzzles is \Cref{algo:type2_sol},
where right after running \Cref{algo:rot_sol}, one needs to update the MGC metric according to the above procedure.

Our numerical tests did not indicate any significant improvement when using this procedure. In order to reduce the computational time of the algorithm, we do not apply it in practice
and mention it as an optional step for the whole algorithm.

\newpage

\section{Possible Estimation of Locations by Quadratic Assignment}
\label{sec:qap-supp}

As we have already mentioned, the main contribution of this work is to introduce a new approach for the recovery of the unknown orientations of patches in type 2 puzzles by using GCL.
Nevertheless, one can try to take advantage of the affinity function $\mW_{\est}$, whose construction is described in \S\ref{sec:conn_graph_const_Type2},
and the fact that for square jigsaw puzzles the affinity function for the unshuffled puzzle, which we denote by $\mW_{\orig}$, is known. In view of \S\ref{sec:theoretical_justification}, the knowledge of $\mW_{\orig}$ is equivalent to the knowledge of $E_{\true}$ and one may write $\mW_{\true} = \mP^{T} \mW_{\orig} \mP$, where $\mP$ is an unknown permutation. Therefore, one may try to match $\mW_{\est}$ with $\mW_{\orig}$. This gives rise to the problem of finding a permutation matrix $\mP$
such that $\mW_{\est}$ and $\mP^{T} \mW_{\orig} \mP$ match. The desired permutation can be expressed as the solution of the following optimization problem:
\begin{equation}
\label{eq:qap}
\argmin_{\mP \in \Perm(n)} \Vert \mW_{\est} - \mP^T \mW_{\orig} \mP \Vert_2^2.
\end{equation}
Note that \eqref{eq:qap} is the Quadratic Assignment Problem (QAP) for the matrices $\mW_{\orig}$ and $\mW_{\est}$.
However, existing solvers are slow as the number of patches increases and thus we are not sure how to make this procedure practical for large puzzles.
A similar idea has been proposed by Andalo et al.~\cite{andalo2012solving} for solving type 1 puzzles. They suggest solving a QAP with different weight matrices by using constrained gradient descent. It is unclear to us
if their procedure is applicable to the QAP problem in \eqref{eq:qap}.

\section{Using Other Top Eigenvectors of the GCW Matrix}
%\label{sec:final_sol}
\label{sec:appen_4}

As we have discussed in \S\ref{sec:theoretical_justification}, if the constructed connection graph is good enough, the top 2 eigenvectors of the GCW matrix can recover the orientations of puzzle patches.
However, when it is impossible to construct an accurate affinity graph (e.g., \Cref{fig:bad_nb}), one might consider the top few eigenvectors,
as they might also contain some useful information about the orientations of patches.
For some puzzles and poorly-estimated connection graphs, the orientations recovered by the top $3$-rd and $4$-th eigenvectors are more accurate than the ones recovered by the top $2$ eigenvectors.
We thus suggest two candidate solutions, one where in the initial iteration (before applying the updates described in \S\ref{sec:w_upd}) we use the top $2$ eigenvectors and another one where in the initial iteration we use the top $3$-rd and $4$-th eigenvectors.
Indeed, in practice, it seems that other top eigenvectors contain some relevant local information.  However, there is no theoretical guarantee for the recovery of the desired orientations by other top eigenvectors.

We remark that this procedure of using the top $3$-rd and $4$-th eigenvectors is not needed at the later updates of \S\ref{sec:w_upd} since the connection graphs are then nicely approximated.
Our experiments indicate that even for the initial stage, the use of this procedure is beneficial only for few images. We thus leave this step as optional. This procedure is summarized in \Cref{algo:type2_sol_opt}.

\begin{algorithm}
\caption{Variation on the solution of type 2 puzzles}
\label{algo:type2_sol_opt}
\begin{algorithmic}
\State{\textbf{Input:} Puzzle Patches: $\{P_i\}_{i=1}^n \subset \RR^{s \times s \times 3}$}
\begin{itemize}
\State Apply \Cref{algo:type2_sol} with $\{P_i\}_{i=1}^n$ to obtain the solution $\{R_{i, 1}\}_{i = 1}^n$ and $\sigma_1$
\State Apply \Cref{algo:type2_sol} with $\{P_i\}_{i=1}^n$, but for the initial iteration (which applies \Cref{algo:rot_sol} in step 2 of \Cref{algo:type2_sol}) use the top $3$-$4$ eigenvectors instead of the top 2 (step 4 of \Cref{algo:rot_sol}) to obtain the orientations $\{ R_{i, 2} \}_{i=1}^n$ and $\sigma_2$
\State Compute the $\Err$ values for $\{ \{R_{i, 1}\}_{i = 1}^n, \sigma_1 \}$ and $\{ \{R_{i, 2}\}_{i = 1}^n, \sigma_2 \}$ according to \eqref{eq:check_sol} and let $\{ \{R_i\}_{i = 1}^n, \sigma \}$ be the one that produces the smaller $\Err$ value
\end{itemize}
\State{\textbf{Return:} $\{R_i\}_{i = 1}^n$ and $\sigma$}
\end{algorithmic}
\end{algorithm}

\section{Summary of the Parameters of \Cref{algo:type2_sol}}
\label{sec:parameter_discussion}
There are eight numerical parameters that were chosen according to some numerical experiments, where three of them are of the update step.
%We remark that with this number of parameters, it is very expensive to find the best values by cross validation. Moreover the statistical properties of natural images are rather diverse, which makes it hard to do cross validation of even few of these parameters.

The first parameter was chosen to be 0.01 in the matrix $\mW_{\init}$ defined in \eqref{eq:w_init_def}. This parameter and the two other obvious values of this matrix, 0 and 1, quantify the neighborness of patches according to the MGC metric. %The value of this parameter needs to be between 0 and 1 and also sufficiently small due to the evidence that the corresponding patches may not be neighbors.

The second parameter was chosen to be 0.2 in \eqref{eq:w_nb}, where its complimentary value was $0.8$. The latter value is a weight for edges that pass the Jaccard test, %and needs to be sufficiently high. On the other hand,
whereas the former one is a weight for edges that could possibly be correct but did not pass the Jaccard test.

The third parameter was set to 0.005 in \eqref{eq:w_nb_005}. This is a weight for an edge connecting two disconnected components. We chose it to be less than 0.01 (the first parameter), as we ordered the weights according to evidence for connection found by the algorithm.

The fourth and fifth parameters were set to $1/3$ and $2/3$ in \eqref{eq:diag_nb_dir_nb_update}. The weights assigned in the latter equation reflect performance according to the diagonal neighbor test. When passing this test, the weight is 1. In the cases of partial passing and no passing of this test, the previously assigned weights are decreased by the factors $2/3$ and $1/3$, respectively.

At last, for the update step, there are three additional parameters. Two of them are set as 0.3 and 0.6 in \eqref{eq:update_nb_aff_final}. These two weights are used in the iterative update of the affinity matrix. The last parameter is the number of iterations of updates, which was set to be 5. Empirically, we noticed no difference when increasing this number.

\section{Demonstration of the Effect of Iterations}
\label{sec:appen_5}

We demonstrate the performance of our algorithm with different numbers of iterations (recall that the iterations are specified in \Cref{algo:type2_sol}).
\Cref{tab:update_5_iter_res} reports the neighbors comparison metric of a variant of our algorithm with fixed numbers of iterations between 1 and 5, while using all four datasets (MIT, McGill, Pomeranz, large Pomeranz).
%These results thus compliments the ones of
As described in \Cref{algo:type2_sol}, our original algorithm iterates the same procedure 5 times, but then chooses either 1, 2, 3, 4, or 5 iterations according to the smallest $\Err$ value, which is defined in \eqref{eq:check_sol}. Therefore, the results of our proposed solution reported in \Cref{tab:comp_table} are slightly better than the ones reported in \Cref{tab:update_5_iter_res}  with 5 iterations.
We note from \Cref{tab:update_5_iter_res} that the average performance with two iterations is better than the one with a single iteration.
In some higher number of iterations, the performance may deteriorate, but there seems to be an improvement with either 3, 4 or 5 iterations over 2 iterations.

\begin{table}[]
\centering
  \caption{\label{tab:update_5_iter_res} Demonstration of performance of our algorithm with fixed numbers of iterations between 1 and 5.
Performance is measured by the neighbors comparison metric.
It is averaged over all images of the dataset and over 20 instances per image; the corresponding standard deviation is also reported.}
\begin{tabular}{|l|l|l|l|l|l|l|l|l|l|l|}
\hline
 Dataset   & \multicolumn{2}{c|}{Iteration 1} & \multicolumn{2}{c|}{Iteration 2} & \multicolumn{2}{c|}{Iteration 3} & \multicolumn{2}{c|}{Iteration 4} & \multicolumn{2}{c|}{Iteration 5} \\ \hline
& mean & std & mean & std & mean & std & mean & std & mean & std \\ \hline
MIT & 90.8  &  14.7   & 94.4 & 10.0     &94.5  &  9.9   & 93.4 &  10.1   & 94.4 & 10.0    \\ \hline
McGill & 84.9  &  23.7  &  88.7 & 18.6   &  89.5  &  17.8  &  91.0  & 16.5  & 91.7  &  15.1  \\ \hline
Pomeranz  & 87.4  &  16.9  & 88.9 &  15.4  & 87.9 &  18.4  & 90.0 & 14.2   & 89.9 & 14.1   \\ \hline
Large Pomeranz &  81.7 &  12.2  & 87.2  & 12.4  & 84.7  & 13.8  &   87.5 & 12.4 &  87.8  &  12.4 \\ \hline
\end{tabular}
\end{table}

\begin{figure}[ht]
  \centering
  \begin{subfigure}[b]{0.32\linewidth}
    \centering\includegraphics[width=1.0\columnwidth]{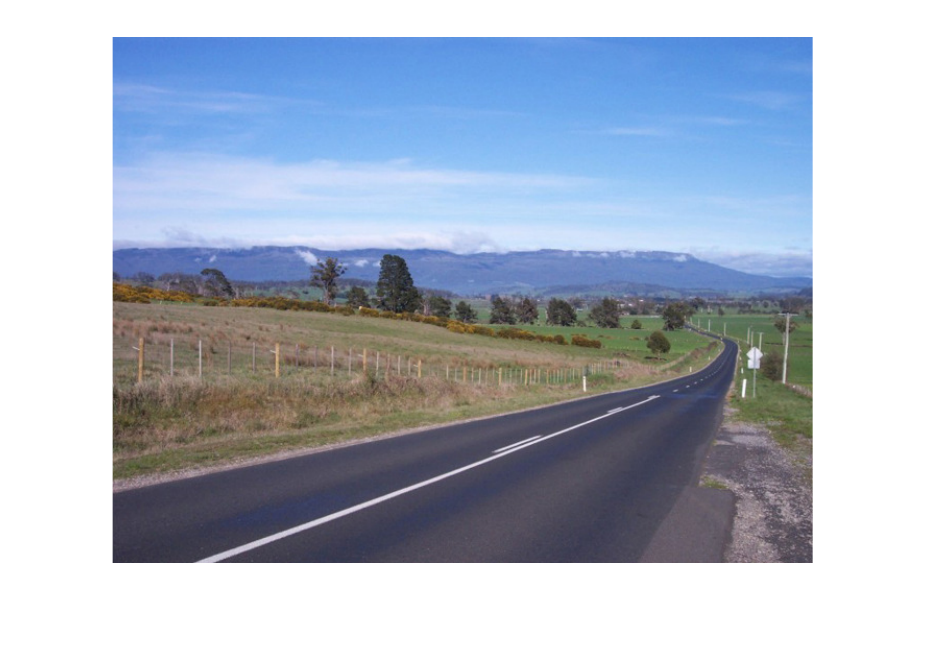}
    \caption{Original Image \label{fig:fig_orig} \\ \text{ }}
  \end{subfigure}%
  \begin{subfigure}[b]{0.32\linewidth}
    \centering\includegraphics[width=1.0\columnwidth]{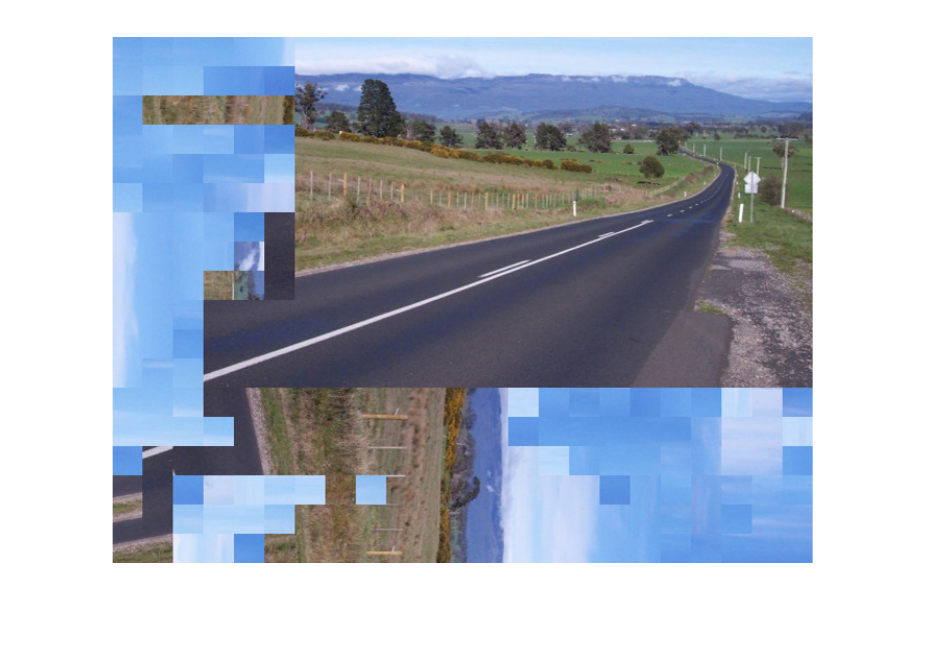}
    \caption{Solution after 1 iteration \\ ~  \text{ } ~ \text{ } (accuracy: 50.1\%)  \label{fig:sol_it_1}}
  \end{subfigure}%
  \begin{subfigure}[b]{0.32\linewidth}
    \centering\includegraphics[width=1.0\columnwidth]{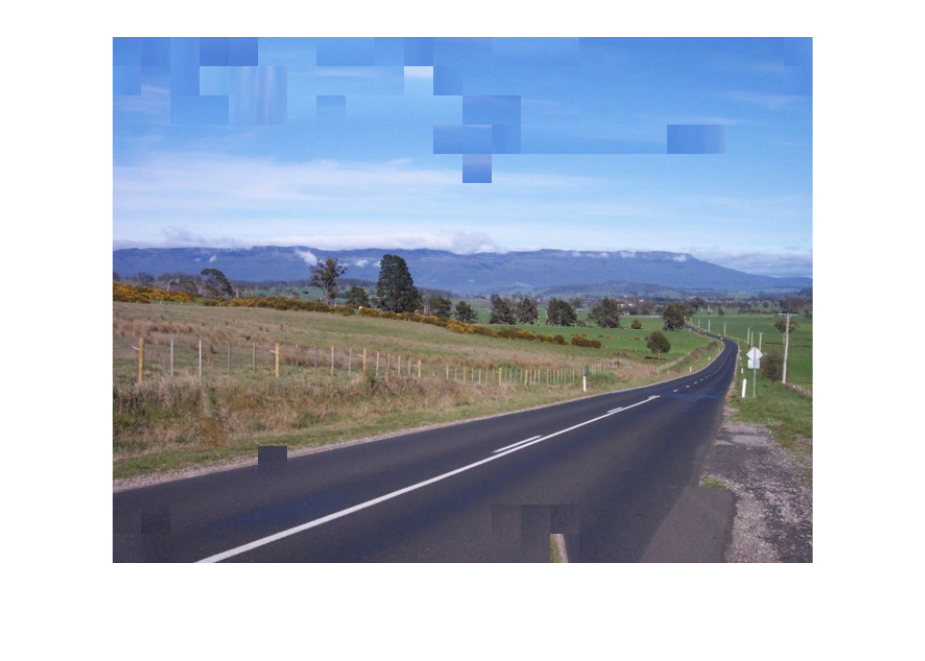}
    \caption{Solution after 2 iterations \\ ~ \text{ } ~ \text{ }  (accuracy: 81.8\%) \label{fig:sol_it_2}}
  \end{subfigure}%

   \begin{subfigure}[b]{0.32\linewidth}
    \centering\includegraphics[width=1.0\columnwidth]{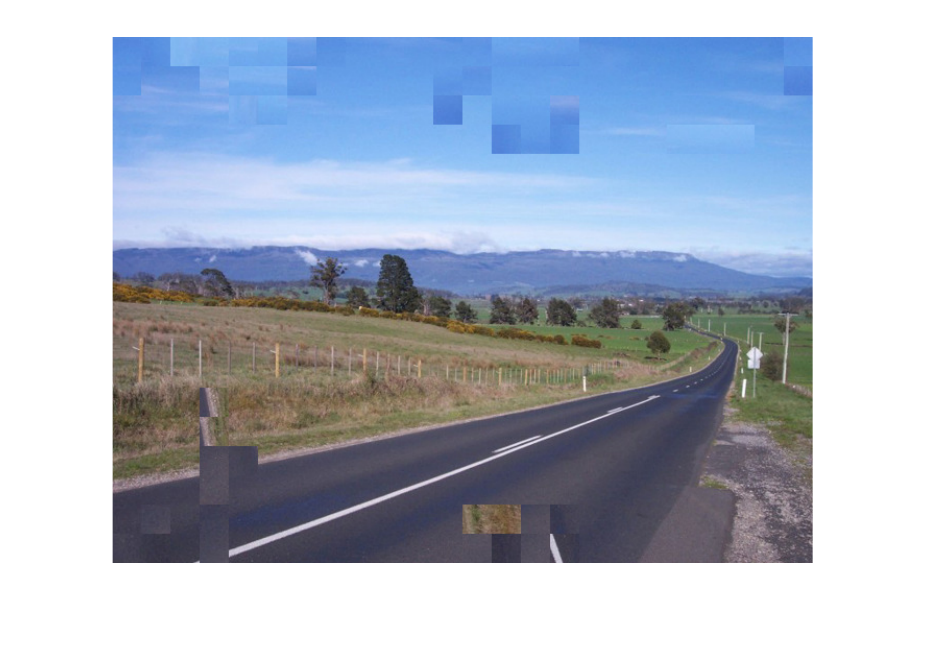}
    \caption{Solution after 3 iterations \\ ~ \text{ } ~ \text{ } (accuracy: 83.2\%) \label{fig:sol_it_3}}
  \end{subfigure}%
  \begin{subfigure}[b]{0.32\linewidth}
    \centering\includegraphics[width=1.0\columnwidth]{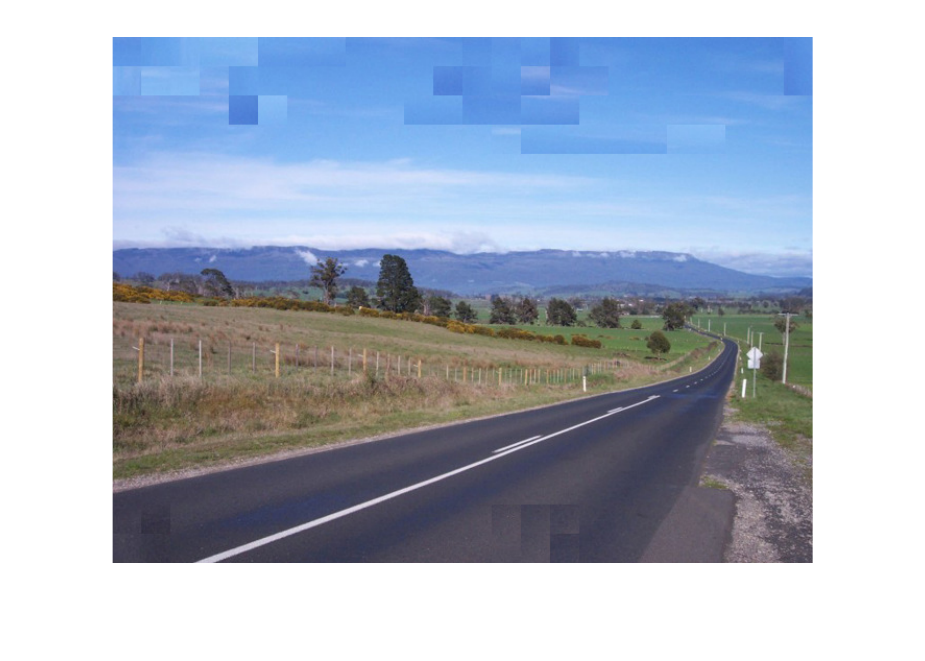}
    \caption{Solution after 4 iterations \\ ~ \text{ } ~ \text{ } (accuracy: 84.9\%)\label{fig:sol_it_4}}
  \end{subfigure}%
  \begin{subfigure}[b]{0.32\linewidth}
    \centering\includegraphics[width=1.0\columnwidth]{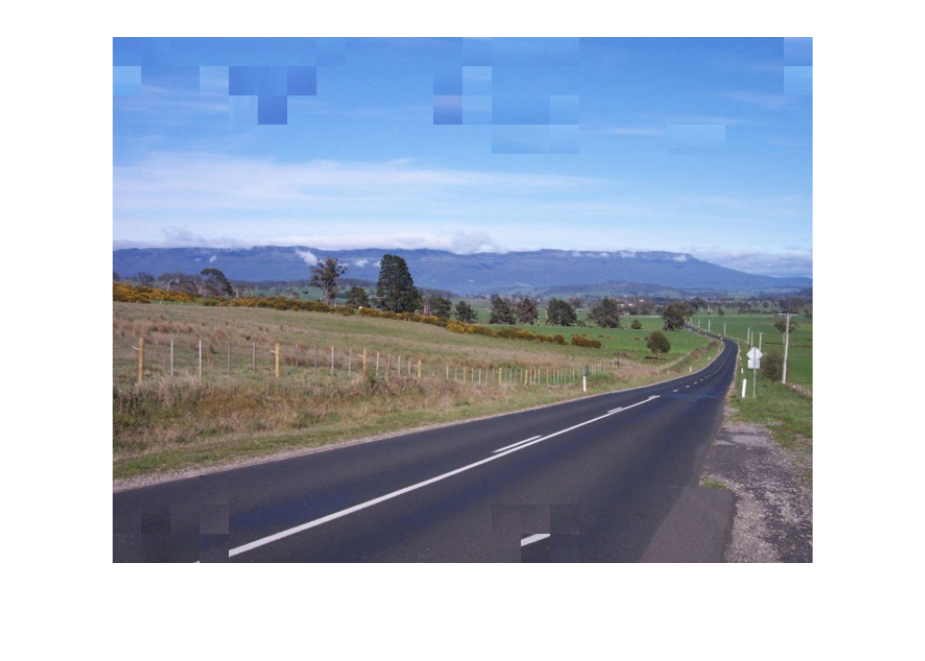}
    \caption{Solution after 5 iterations \\ ~ \text{ } ~ \text{ } (accuracy: 86.4\%) \label{fig:sol_it_5}}
  \end{subfigure}%
  \caption{Example of applying our algorithm with different numbers of iterations.  The original image is shown on top left and has 432 patches of size $28 \times 28$. The next images show solutions by our algorithm with 1-5 iterations. Their captions report accuracy in terms of the neighbors comparison metric. \label{fig:update_5_iter}}
\end{figure}

\Cref{fig:update_5_iter} demonstrates the step by step solution of a certain puzzle after 5 iterations of \Cref{algo:type2_sol}. The original puzzle is in \Cref{fig:fig_orig} and the corresponding solutions for iterations 1-5 are in \Cref{fig:sol_it_1}-\ref{fig:sol_it_5}, respectively. Note that the biggest improvement happened when applying two iterations (vs.~a single iteration).

\end{document}